\documentclass[journal]{IEEEtran}  

\usepackage{graphicx} 
\usepackage{amsmath} 
\usepackage{amssymb}  
\usepackage{algorithm}
\usepackage[T1]{fontenc}
\usepackage{algpseudocode}
\usepackage{xcolor}
\usepackage{nicefrac}
\usepackage{booktabs}  
\usepackage[utf8]{inputenc}
\usepackage{amsmath}
\usepackage{amssymb}
\usepackage{graphicx}
\usepackage{cancel}
\usepackage{dsfont}
\usepackage{algorithm}
\usepackage{algpseudocode}
\usepackage{tikz}
\usepackage{bbm}
\usepackage{dsfont}
\usepackage{graphicx}
\usepackage{dsfont}
\usepackage{graphicx}
\usepackage{subcaption}
\usepackage{multirow}
\usepackage{placeins}

\usepackage{enumerate}
\usepackage{xspace}
\usepackage{xcolor,colortbl}
\usepackage{tabularx}
\usepackage[labelfont=bf]{caption}

\usepackage{multicol}
\usepackage{multirow}
\usepackage{wrapfig}

\usepackage[font=scriptsize,labelfont=bf]{caption}

\newtheorem{thm}{Theorem}
\newtheorem{lem}{Lemma}

\newtheorem{cor}{Corollary}
\newtheorem{defn}{Definition}

\newcommand{\qed}{$\blacksquare$}

\newcommand{\bydef}{\triangleq}
\newcommand{\probd}{\mathbb{P}}
\newcommand{\prob}{\mathrm{P}}

\definecolor{ultraviolet}{RGB}{95, 75, 139}
\definecolor{deepteal}{RGB}{0, 51, 51}
\definecolor{hibiscus}{RGB}{176,48,96}
\definecolor{emeraldgreen}{RGB}{0,155,119}
\definecolor{nicegreen}{rgb}{0.1,0.5,0.1}
\definecolor{nicebrown}{RGB}{135,48,0}
\definecolor{violet}{RGB}{238,130,238}
\definecolor{inkscapePurple}{RGB}{128,0,128}
\definecolor{inkscapeOlive}{RGB}{128,128,0}
\definecolor{indigo}{rgb}{75,0,130}
\definecolor{turquoise}{RGB}{0,116,157}

\title{\LARGE \bf
Anytime Probabilistically Constrained Provably Convergent Online Belief Space Planning
}

\author{Andrey Zhitnikov$^{1}$ and Vadim Indelman$^{2,3}$ \\
\thanks{This work was  supported by the Israel Science Foundation (ISF).}
$^1$Technion Autonomous Systems Program (TASP) \\
$^2$Department of Aerospace Engineering\\
$^3$Department of Data and Decision Science\\
Technion - Israel Institute of Technology, Haifa 32000, Israel\\
\small{\texttt{andreyz@campus.technion.ac.il, vadim.indelman@technion.ac.il}}}

\begin{document}

\maketitle

\begin{abstract}
	Taking into account future risk is essential for an autonomously operating robot to find online not only the best but also a safe action to execute.     
	In this paper, we build upon the recently introduced formulation of probabilistic belief-dependent constraints.  We present an anytime approach employing the Monte Carlo Tree Search  (MCTS) method in continuous domains. Unlike previous approaches, our method assures safety anytime with respect to the currently expanded search tree without relying on the convergence of the search.   We prove convergence in probability with an exponential rate of a version of our algorithms and study proposed techniques via extensive simulations.  Even with a tiny number of tree queries, the best action found by our approach is much safer than the baseline.  Moreover, our approach constantly finds better than the baseline action in terms of objective. This is because we revise the values and statistics maintained in the search tree and remove from them the contribution of the pruned actions.   
\end{abstract}
\IEEEpeerreviewmaketitle


\section{Introduction and Related Work}
\label{sec:Intro}
\IEEEPARstart{C}{asting} decision-making under uncertainty as a Partially Observable Markov Decision Process (POMDP) is considered State-Of-The-Art (SOTA). Under partial observability the decision-making agent does not have complete information about the state of the problem, so it can only make its decisions based on its ``belief'' about the state. In a continuous domains in terms of POMDP state, the belief, in a particular time index, is the Probability Density Function (PDF) of the state given all concurrent information in terms of performed actions and received observations in an alternating manner, plus the prior belief. 
A POMDP is known to be undecidable \cite{Madani03AI} in finite time.

Introducing various constraint formulations into POMDP is essential for, e.g., ensuring safety \cite{Santana16aaai}, \cite{Zhitnikov22arxiv} and efficient Autonomous Exploration \cite{Zhitnikov24TRO}.  Yet, the existing online approaches in anytime setting have problems and therefore fall short of providing reliable and safe optimal autonomy.   This crucial gap we aim to fill in this paper. 

Similar to almost any online POMDP
solver today such as MCTS, our method constructs a belief tree and uses the tree to represent the POMDP policy. We prune dangerous actions from the belief tree and revise the values and statistics that an MCTS tree maintains. Anytime, our search tree contains only the safe actions in accord to our definition of safe action, which will appear shortly. 
Our work lies in continuous domain in terms of actions and the observations. In such a setting, there are approaches to tackle averaged cumulative constraint using anytime MCTS methods \cite{Jamgochian23ICAPS}, \cite{Jamgochian23arxiv}.  We now linger on the explanation of what the averaged constraint is.  

Under partial observability, namely in the POMDP setting, there are naturally two stages to consider in order to introduce a constraint.   The first stage arises from the belief itself. Usually, at this stage, the state-dependent payoff operator 
is averaged with respect to the corresponding belief to obtain a belief-dependent one.  It is then summed up to achieve a cumulative payoff. We use the term payoff to differentiate between reward operator and emphasize that a belief-dependent payoff constraint operator shall be as large as possible as opposed to the cost operator. The second stage arises from the distribution of possible future observations episodes. At this stage, commonly, the cumulative payoff is again averaged but with respect to future observations episodes and then thresholded, thereby forming an averaged cumulative constraint.  
Such a formulation is sufficient for ensuring safety in limited cases as we will further see in Section \ref{sec:ExConstr}. 
This is because it permits deviations of the individual values within the summation. 

Let us now describe the  MCTS methods mentioned above  to tackle averaged cumulative constraint. The seminal paper in this direction is \cite{Lee18nips}. It leans on the rearrangement of the constrained objective using the occupancy measure described in \cite{Altman99book}. Such a reformulation is appealing since it transforms the problem into linear programming bringing convexity to the table and enjoying from strong duality. The authors of \cite{Jamgochian23ICAPS} extend the approach from \cite{Lee18nips} to continuous spaces. Still, both papers \cite{Lee18nips} and \cite{Jamgochian23ICAPS} assure constraint satisfiability only at the limit of the convergence of the iterative procedure, namely in infinite time. Since these are iterative methods, to assure anytime constraint satisfiability we need to project the obtained occupancy measure at each iteration to the space defined by the constraint. If dual methods are involved \cite{Beck17first} such a projection does not make much sense, e.g., the projection might lead to a step direction vector on the boundary of all the constraints, making it zero vector. Employing the primal methods in continuous spaces also appears to be problematic since the summations in \cite{Lee18nips} are transformed into integrals. The paper \cite{Jamgochian23arxiv} provides some sort of anytime satisfiability by introducing high-level action primitives (options). Still, \cite{Jamgochian23arxiv} suffers from limitations, e.g. it requires crafting low-level policies, meaning knowing how the robot shall behave a priori. In addition, the options shall be locally feasible.  
Additionally, for efficiency reasons, the duality based approaches perform a single tree query of the MCTS, instead of running MCTS until convergence in the maximization of the Lagrangian dual objective function phase (See section $8.5.2$ in \cite{Beck17first}) of dual ascend.

In all three papers \cite{Lee18nips}, \cite{Jamgochian23ICAPS}, \cite{Jamgochian23arxiv} the averaged cumulative constraint is enforced solely from the root of the belief tree. This is suboptimal since within a planning session it is not taken into account that the constraint will be enforced at the future planning sessions. In other words, the contemplation of a robot about the future differs from its actual future behavior.   
This aspect has been fixed by \cite{Ho2023arxiv}. As we will further see in Section~\ref{sec:Approach}, our approach naturally handles this problem. Moreover,  \cite{Ho2023arxiv} assures fulfillment (admission) of the recursive averaged cumulative constraint anytime with respect to search tree constructed partially with the reward bounds and partially with rewards themselves.  Yet, the algorithm presented in \cite{Ho2023arxiv} requires that the value function is bounded on the way down the tree to assure the exploration. This is commonly achieved by assuming that the state-dependent reward is trivially bounded from above and below. This does not hold for general belief-dependent reward functions. Moreover, the exploration outlined in that paper is valid for discrete spaces only. All in all, the extension of that work to continuous spaces and belief-dependent rewards requires clarification. 

\paragraph{Support for general belief dependent rewards and payoff/cost operators and MCTS convergence}
We now clarify whether or not the mentioned above solvers support belief-dependent cost/payoff operators and rewards. It was suggested in \cite{Zhitnikov22arxiv},\cite{Zhitnikov24TRO} that general belief-dependent payoff/cost operators are extremely important. As mentioned in \cite{Zhitnikov22arxiv} Value-at-Risk (VaR) and Conditional VaR (CVaR) over the distance to the safe space allow for control of the depth the robot can plunge into the obstacle. To rephrase that, these operators measure how bad the disaster (collision) will be. See Appendix~\ref{sec:VaRCVaR}, for details. The Information Gain discussed in \cite{Zhitnikov24TRO} is relevant for exploration. The paper \cite{Zhitnikov24TRO} discussed the general belief-dependent averaged constraint of the form \eqref{eq:AveragedConstraintRoot} 
in a high dimensional setting and in the context of Information Gain.
The iterative schemes in  \cite{Lee18nips}, \cite{Jamgochian23ICAPS} lean on the convergence of MCTS. It has been shown in \cite{Munos2014book} that even in discrete spaces and with bounded rewards it can take a very long time for MCTS to converge. In the case of unbounded reward or the cost-augmented objective of \cite{Lee18nips}, \cite{Jamgochian23ICAPS}, the MCTS may converge slowly. 
If such an augmented reward has a large variance, it will be needed a huge amount of tree queries for action-value estimate (to be defined shortly) at each belief node of the belief tree to converge. The large variance can be the result of an unrestrained variability of the rewards or a large Lagrange multiplier.        

There are several constraint formulations for POMDP. Below we discuss the most prominent techniques one by one.  
\paragraph{Shielding POMDPs} 
There is a growing body of literature on shielding POMDPs. The shield is a technique to disable the actions that can be executed by the agent and violate the shield definition. There are several shield definitions.  Online methods \cite{Ajdarow23AAAI}, \cite{Mazzi23ai} in this category utilize Partially Observable Monte-Carlo Planning (POMCP) algorithm \cite{Silver10nips}. These works have the same problems we are solving in this paper: one way or another, the actions violating the shield definition participate in the planning procedure,  yielding a suboptimal result. The work  \cite{Mazzi23ai} enforces the shied outside the POMCP planning. As we further show,  not considering safety in the future times, namely within the planning session, can lead to a suboptimal  planning result.   

\paragraph{Chance Constrained (CC) Online Planning}
A recent work \cite{Moss24Arxiv} tackles online planning with chance constraints in an anytime setting. 
This paper suggests using a Neural Network (NN) to approximate CC enforced, with an adaptive threshold, from each belief considered in the planning session. This work trains NN offline. Therefore the error stemming from the discrepancy of simulated and real data is unknown. Moreover, it is not clear how complex the NN shall be to achieve zero loss in training to ensure no error in CC approximation, so even if no discrepancy discussed before exists, the NN inference may be slow.  In this method, dangerous actions do not participate in the planning session.   

\paragraph{Safe control Under Partial Observability}
There are a variety of robust control approaches natively tailored for continuous state/action/observation spaces \cite{Chou22WAFR},\cite{Dean21CORL}. However, these methods are usually limited to very specific rewards/objectives and tasks, such as reaching a goal state or to be as close as possible to a  nominal trajectory. Moreover, in both papers the system dynamics are control-affine. Without this assumption, it is not clear how to enforce the constraint through a derivative of the barrier function.    
\subsection{Contributions}
Below  we list down our contributions in the same order as they appear in the manuscript.  
\begin{itemize}
	\item By constraining directly the problem space and not the dual space we present an anytime MCTS based algorithm for safe online decision making with safety governed by a Probabilistic Constraint (PC). Our approach enjoys anytime safety guarantees with respect to the belief-tree expanded so far and works in continuous state, action  and observation spaces. 
	When stopped anytime, the action returned can be considered as the best safe action under the safe future policy (tree policy) expanded so far. Our search tree {\bf solely} consists of  safe actions. We prove convergence in probability with an exponential rate of our approach. 
	\item Another contribution on our end is constraining the beliefs with incorporated outcome uncertainty stemming from an action performed by the robot and without incorporating the received observation.  This is alongside the constraint over the posterior belief with included last observation. To the best of our knowledge, no previous works do that.  
	\item We also spot a problem happening in duality based approaches arising from averaging unsafe actions in MCTS phase. Therefore, an additional contribution of ours is an analysis of this phenomenon.   
	\item We simulate our finding on  several continuous POMDP problems.  
\end{itemize}

\subsection{Notation} 
We use the $\square$ as a placeholder for various quantities. The values in $\square$ can be replaced by one of the respective options. We also extensively use the indicator function notation, which is $\mathbf{1}_A(\square)$. This function equals to one if and only if $\square {\in} A$. 
By lowercase letters we denote the random variables of their realizations depending on context. By the bold font we denote vectors of operators in time of different lengths.  
We denote estimated values by $\hat{\square}$. 
\subsection{Paper Roadmap}
This paper proceeds  with the  following structure. Section \ref{sec:Background} presents  relevant background. Section~\ref{sec:PF} then formulates the problem. Section \ref{sec:Approach} presents our approach. Section~\ref{sec:Baseline} discusses our baseline. Section \ref{sec:SimResults} gives experimental validation of the proposed methodology. Finally, Section~\ref{sec:Concl} concludes the paper.   

\section{Background}
\label{sec:Background}
This section gives the background required for presenting our approach. Specifically, we discuss belief-dependent POMDP, its reformulation to Belief-MDP (BMDP), and the MCTS.
\begin{figure}[t]
	\centering 
	\includegraphics[width=0.5\textwidth]{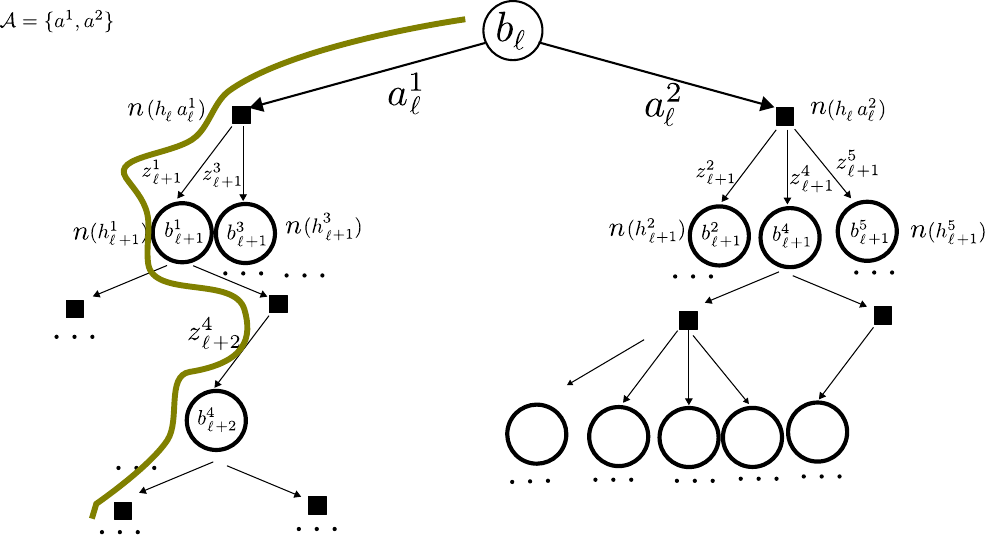}
	\caption{Here we plot the asymmetric search tree approximating stochastic future policy. For simplicity the action space here is $\mathcal{A}{=}\{a^1, a^2\}$. We behold that many actions emanating from each belief node and each action has weight defined by relevant visitation count as in \eqref{eq:QEstApproxMCTSNORoll}. Thus, the MCTS approximates stochastic future policy. Note that here the observations and beliefs has global index (superscript) while actions have local index according to the action number in the space $\mathcal{A}$.}
	\label{fig:AssymetricPolicyTreeLace}
\end{figure}
\subsection{Belief-dependent POMDP}
The POMDP is a tuple $\langle\mathcal{X}, \mathcal{A}, \mathcal{Z}, \mathrm{T}, \mathrm{O}, \rho, \gamma, b_0\rangle$ where $\mathcal{X}, \mathcal{A}, \mathcal{Z}$ represent continuous state, action, and observation spaces with $x { \in } \mathcal{X}$, $a { \in } \mathcal{A}$, $z { \in  } \mathcal{Z}$ the 
individual state, action, and observation, respectively. 
$\mathrm{T}(x' , a, x) {\bydef}  \probd_{\mathrm{T}}(x' | x, a)$ is a stochastic transition model from the past state $x$ to the subsequent $x'$ through action
$a$, $\mathrm{O}(z , x)  {\bydef}  \probd_{\mathrm{O}}(z|x)$ is the stochastic observation model. $\rho{:} \mathcal{B}{\times} \mathcal{A} {\times} \mathcal{Z} {\times} \mathcal{B} {\mapsto} \mathbb{R}$ is a  belief-dependent reward incurred as a result of taking  an action $a$ from the belief $b$, receiving and observation $z'$ and updating the belief to $b'$. By $\mathcal{B}$ we denote the space of all possible beliefs. $\gamma {\in} (0, 1]$ is the discount factor, $b_0$ is the prior belief.  
Purely for clarity of the exposition we further assume that the reward depends solely on a pair of consecutive-in-time beliefs and an action in between.  In addition we suppose $\gamma{=}1$.  To remove unnecessary clutter we assume that planning starts from $b_0$. Extension to the arbitrary planning time is straightforward. 

Let $h_{\ell}$ be a history. The history is the set that comprises the prior belief $b_{0}$, the actions $a_{0:\ell-1}$ and the observations $z_{1:\ell}$ that would be obtained by the agent up to time instance $\ell$ such that $h_{\ell} {\bydef} \{{\color{nicegreen}{b_0}}, a_{0:\ell-1}, z_{1:\ell} \}$. We emphasize by the {\color{nicegreen} green} color that ${\color{nicegreen}{b_0}}$ is given, but the actions $a_{0:\ell-1}$ and observations $z_{1:\ell}$  can vary. In addition due to the assumption that the planning session starts from the prior belief $b_0$ we can have only the future history simulated in planning in this work.  For completeness we  define $h_0 {\bydef} \{b_0\}$
The posterior belief $b_\ell$ is given by
\begin{align}
b_\ell(x_\ell) {\bydef}  \probd(x_\ell |{\color{nicegreen}{b_0}}, a_{0:\ell-1}, z_{1:\ell}){=} \probd(x_\ell |h_{\ell}){=}\probd(x_\ell |b_{\ell}).   \label{eq:Belief}
\end{align}
The belief is a function of history such that we sometimes write $b(h)$ instead of $b(x)$ and use the corresponding $h$ notation to point to the belief $b(h)$.  
The actions within the history are coming from the execution policy. A deterministic policy $\pi$ is a sequence of functions $\pi {=} \pi_{0{:}\ell{-}1}$ for $\ell {\in} [1 {\dots} L{-}1]$, where the momentary
function $\pi_i {:} \mathcal{B} {\mapsto} \mathcal{A} \ \forall i$. In each time index, the policy maps belief to action. For better readability sometimes we will omit the time index for policy or denote $\pi_{0:\ell-1}$
as $\pi_{0+}$ and $\pi_{1:\ell-1}$ as $\pi_{1+}$ . The policy can also be stochastic. In this case, it is a distribution of taking an action $a_{\ell}$ from a belief $\pi_{\ell}(a_{\ell},b_{\ell}) {=} \pi_{\ell}(a_{\ell},h_{\ell}){=} \probd^{\pi}_{\ell}(a_{\ell}|b_{\ell}(h_{\ell})) {=} \probd^{\pi}_{\ell}(a_{\ell}|h_{\ell})$\footnote{Here, the capability of history being switched with the belief has to be inspected for a particular belief update. In MCTS, as we will shortly see,  the stochastic policy is history-dependent and can vary even if the belief is the same at different history nodes. In this paper, the belief update is a particle filter. Therefore, the probability of obtaining the same belief at different histories is zero.}. Here the action space $\mathcal{A}$ is the space of outcomes and the mapping is $\pi_{i} {:} \mathcal{B} {\times} \mathcal{A} {\mapsto} \mathbb{R}$. We have that $\pi_{0:L-1}{=}\{ \probd^{\pi}_i \}_{i=0}^{L-1}$. Yet, in $h_{\ell}$ we have a specific realization of actions of such a policy in previous time instances. 
\begin{figure*}[t]
	\centering
	\begin{minipage}[t]{0.20\textwidth}
		\centering 
		\includegraphics[width=0.65\textwidth]{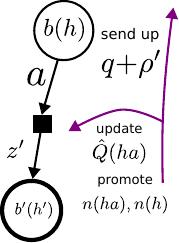}
		\subcaption{}
		\label{fig:MCTSup}
	\end{minipage}
	\hfill
	\begin{minipage}[t]{0.79\textwidth}
		\includegraphics[width=\textwidth]{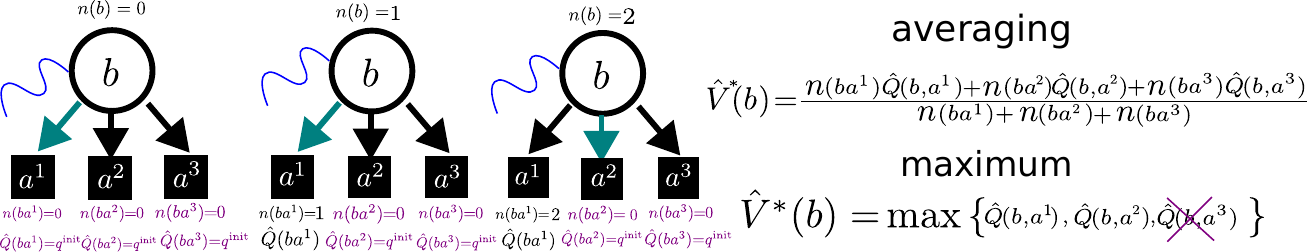}
		\subcaption{ }      
		\label{fig:Exploration}	 
	\end{minipage}
	\caption{\textbf{(a)} Visualization of the MCTS operations when ascending up the search tree. We update $\hat{Q}(ha)$, visitation counts $n(ha)$ and $n(h)$, send up the lace $q$ of the cumulative reward; \textbf{(b)} Illustration of the MCTS operation when descending down the tree. First, upon reaching a leaf node, the current action space is unfolded to belief-action nodes. MCTS selects each action infinitely often. At the way up the belief tree the classical MCTS takes the average of the actions tried so far (after relevant updates on the way up) to update the estimator of \eqref{eq:Qfunc}. In this illustration, $a^3$ still did not tried and therefore do not participate. On the way up $n(ha^3)$ stays zero.  }
	\label{fig:MCTS}
\end{figure*}
When the agent performs an action $a$ and receives an observation $z'$, it shall update its belief from $b$ to $b'$. Let us denote the update operator by $\psi$ such that $b'{=}\psi(b, a, z')$. In our context, it will be a Particle Filter (PF) since we focus on the setting of nonparametric beliefs. However, this is not an inherent limitation of our approach. Any belief update method would be suitable. We define a propagated belief $b'^{-}$ as the belief $b$ after the robot performed an action $a$ and before it received and observation, namely  
\begin{align}
b^-_{\ell}(x_{\ell}) {\bydef}  \probd(x_{\ell} |h_{\ell-1}, a_{\ell-1}){=} \probd(x_{\ell} |h^-_{\ell}) {=} \probd(x_{\ell} |b^-_{\ell}). \label{eq:PropagatedBelief}
\end{align} 
We define $h^-_{\ell}{\triangleq}h_{\ell} {\setminus} \{z_{\ell}\} {=} \{b_0, a_{0:\ell-1}, z_{1:\ell-1}\} $. The unconstrained, online decision making objective is the action-value function specified as  
\begin{align}
\!\!\!\!Q^{\pi}(b_0, a_0; \boldsymbol{\rho}_1) {\bydef}  \mathbb{E}^{\mathrm{T},\mathrm{O}}_{z_{1}}\big[\rho_{1}(b_0, a_0, b_{1}) {+} V^{\pi}(b_{1};\boldsymbol{\rho}_2)\big| b_0, a_0\big].\!\!\!\!\!\!\! \label{eq:Qfunc}
\end{align}
Here the we added the subscript to the reward $\rho_{\square+1}(b_{\square}, b_{\square+1})$ to emphasize that it is a random variable and it is allowed not to specify dependency on consecutive-in-time beliefs and the action in between. The $V^{\pi}(b_{\square};\boldsymbol{\rho}_{\square+1})$ is the value function under the stochastic policy $\pi$ and $\boldsymbol{\rho}_{\ell}$ is a vector of belief-dependent operators of appropriate length.   
The value function materializes as  
\begin{align}
V^{\pi}(b_{0};\boldsymbol{\rho}_{1}) {\bydef}\textstyle \mathbb{E}^{\mathrm{T},\mathrm{O}}\big[\sum_{\ell=0}^{L-1}\rho_{\ell+1}(b_{\ell}, a_{\ell}, b_{\ell+1})\big| b_0, \pi \big]. \label{eq:Value}
\end{align}
Let us present the following lemma to better understand the structure of \eqref{eq:Value} under a  stochastic policy.   
\begin{lem}[Representation of the Value Function] \label{thm:RepresentValueStochPolicy}
	The value function under a stochastic execution policy complies to the following form
	\begin{equation}
	\label{eq:ValueStochPolicy}
	\begin{gathered}
	\mathbb{E}^{\mathrm{T},\mathrm{O}}\big[\textstyle\sum_{\ell=0}^{L-1}\rho_{\ell+1}(b_{\ell}, a_{\ell}, b_{\ell+1})\big| b_0, \pi \big] = \\
	\textstyle\sum_{\ell=0}^{L-1} \mathbb{E}^{\mathrm{T},\mathrm{O}}\big[\rho_{\ell+1}(b_{\ell}, a_{\ell}, b_{\ell+1})\big| b_0, \pi \big]=\\
	\textstyle\sum_{\ell=0}^{L-1} \underset{a_0}{\mathbb{E}}\Big[\underset{b_1}{\mathbb{E}} \Big[  \underset{a_1}{\mathbb{E}} \Big[\underset{b_2}{\mathbb{E}}\Big[  \dots \\
	\underset{a_{\ell}}{\mathbb{E}}\big[\mathbb{E}\big[\rho_{\ell+1}| b_{\ell}, a_{\ell}\big]\big| b_{\ell},\pi_{\ell}\Big] {\dots} \Big| b_1, a_1\Big]\Big|b_1, \pi_1\Big] \! \Big| b_0, a_0\Big]\!\Big| b_0, \pi_0\Big]. \!\!\!\!\!\!\!\!
	\end{gathered}
	\end{equation}
\end{lem}
\noindent  We laid out the detailed proof in Appendix~\ref{proof:RepresentValueStochPolicy}.
In online decision making, the future belief tree policy $\pi_{1+}$ is approximated as part of the decision process. We denote the best future policy as $\pi^*_{(k+1)+}$. 
The best deterministic policy for the present time is given by $\pi_0(b_0) {=} \arg\max_{a_0 \in \mathcal{A}}  Q^{\pi^*_{1+}}(b_0, a_0;\boldsymbol{\rho}_1)$. The best stochastic policy is the solution of $\max_{\pi_\ell} \mathbb{E}_{a_\ell \sim \probd^{\pi}_{\ell}(a_{\ell}|b_{\ell}) }[Q^{\pi^{*}_{(\ell+1)+}}(b_\ell, a_{\ell};\boldsymbol{\rho}_{\ell+1})]$.  
The interlink between \eqref{eq:Value} and \eqref{eq:Qfunc} is  $V^{\pi}(b_{\ell};\boldsymbol{\rho}_{\ell+1}){=} Q^{\pi_{(\ell+1)+}}(b_\ell, \pi_\ell(b_\ell);\boldsymbol{\rho}_{\ell+1})$ in case of deterministic policies and 
$V^{\pi}(b_\ell;\boldsymbol{\rho}_{\ell+1}) = \mathbb{E}_{a_{\ell} \sim \probd^{\pi}_{\ell}(a_\ell|b_{\ell}) }[Q^{\pi}(b_\ell, a_{\ell};\boldsymbol{\rho}_{\ell+1})]$ in case of the stochastic policies.  
\subsection{Belief State MDP}
\label{seq:BMDP}
To employ solvers crafted for fully observable Markov Decision Processes (MDP) we can cast POMDP as a Belief-MDP (BMDP).  The BMDP is a following tuple  $\langle\mathcal{B}, \mathcal{A}, \mathrm{T}_b, \rho, \gamma, b_0\rangle$, where $\mathcal{B}$ is the space of all possible beliefs defined by \eqref{eq:Belief}. The belief state transition model follows
\begin{align}
\!\!\!\!\!\mathrm{T}_{b}(b, a, b') {\triangleq} \probd_{\mathrm{T}_b}(b'|b, a) {=} \textstyle \!\!\!\!\!\int\limits_{z' \in \mathcal{Z}}\! \underbrace{\probd(b'|b,a, z')}_{\delta(b'-\psi(b,a,z'))} \probd(z'|b,a)\mathrm{d}z'.\!\!\!\!\!
\end{align}    
%
The next section describes SOTA approach to solve unconstrained continuous POMDP online, namely MCTS. There we deal with estimators of the \eqref{eq:Qfunc} and \eqref{eq:Value}.  We denote estimated values by $\hat{\square}$. 

Further, we shorten the notation and mark $\hat{V}^{\pi^*}(b;\boldsymbol{\rho}')$ by $\hat{V}^{\ast}(h)$ and $\hat{Q}^{\pi^*}(ba; \boldsymbol{\rho}')$ by $\hat{Q}(ha)$. We will use the dependence on history $h$ and the corresponding belief $b(h)$ interchangeably since the history $h$ defines the location in the belief tree as opposed to the belief which possibly can be identical for more than single history. It will be clarified in the next section. 
In the next section we will see why in time zero we have deterministic policy and in future time the policy is stochastic. 


\subsection{Monte Carlo Tree Search}
\label{sec:mcts}

MCTS constructs the search tree comprised by belief nodes (transparent circles) and belief-action nodes (black squares), by iteratively descending down the tree and ascending back to the root (See Fig.~\ref{fig:AssymetricPolicyTreeLace} and \ref{fig:MCTS}).  On the way down the tree, the exploration mechanics selects an action. The Double Progressive Widening (DPW) manages the sampling of new actions and observations.  On the way back to the root MCTS updates action value estimates at each belief action node (Fig.~\ref{fig:MCTSup}) and relevant visitation counts.  In the case of belief-dependent rewards, beliefs represented by particles and continuous setting of states, actions, and observations, MCTS is applied on the level of Belief-MDP (BMDP) and called Particle Filter Tree with DPW (PFT-DPW) \cite{Sunberg18icaps}. DPW solves the problem of shallow trees in a continuous setting. This problem arises because in this setting it is impossible to sample the same action and observation twice. The DPW technique enables gradually expanding new actions and observations as the tree search progresses. 
With a slight abuse of notation, we sometimes switch the dependence of various quantities on belief and dependence on the corresponding history. This is because same belief can correspond to different histories. Therefore to properly mark the position at the search tree we shall use history $h$ instead of belief $b(h)$.    
The exploration score is defined as
\begin{align}
\mathrm{sc}(h, a) \bydef \underbrace{\hat{Q}(h, a)}_{\substack{ {\text{belief action node } ba} \\ {\text{indexed by } ha }}}{+} k\sqrt{\nicefrac{f(n(h))}{n(ha)}}  \label{eq:Expl}
\end{align}
governs the selection of the actions down the tree, where $n(h)$ is the visitation count of the belief nodes, $n(ha)$ is the visitation count of belief-action nodes and $k$ is the exploration constant (Fig.~\ref{fig:Exploration}). The notation $ha$ is the history $h$ with action $a$ appended to the end, alias to $h^-$ with action $a$ explicitly seen.  The function $f$ is $\log$ in the case of Upper Confidence Bound (UCB) \cite{Kocsis06ecml} exploration and power in the case of Polynomial Upper Confidence Tree (PUCT) \cite{Auger13Sp}.  
The MCTS  can be run with rollout and without. In the case of rollout configuration from each new belief node, the rollout is initiated to provide an initial 
$\hat{V}^*$ of the newly added belief node. This is not mandatory since if no rollout is initiated the MCTS will continue to descend down the tree until the deepest level with the first action from the action space $\mathcal{A}$ (first sampled action in case of continuous action space). 
{\bf Not in every tree query} the MCTS will expand a new node. In some queries, only visitation counts are promoted (lace already present in the tree incorporated to pertinent $\hat{Q}$). In continuous spaces it happens because of DPW. 
DPW as well as increasing the visitation counts without adding a new lace introduces observations distribution shift. This is out of the scope of this paper.   
The  $\hat{Q}(b(h), a)$ estimates are assembled from the laces (yellow curve in Fig.~\ref{fig:AssymetricPolicyTreeLace}). Another name for lace is decision epoch or episode or script. 
Imagine that at the depth $\ell$ of the belief tree, each belief has a {\bf global} index  $i_{\ell}$ per depth $\ell$, say index runs from left to right over all the belief nodes at level $\ell$. Let us define the set of global indices of posterior beliefs which are children of $b^{i_{\ell}}_{\ell}(h^{i_{\ell}}_{\ell})$ and action $a_{\ell}$ by $C(h^{i_{\ell}}_{\ell}a_{\ell})$. We also define the set of actions emanating from  $b^{i_{\ell}}_{\ell}$ by $C(h^{i_{\ell}}_{\ell})$. Only in time zero we make these sets and visitation counts depend on belief instead of history. In the next equation, we omit the subscript denoting time instance of histories, beliefs, and actions. Suppose MCTS is configured to run without rollout. In this case $\hat{Q}(h^{i_\ell}_{\ell}, a_{\ell})$ reads 
\begin{equation}
\label{eq:QEstApproxMCTSNORoll}
\begin{gathered}
\!\!\!\!\hat{Q}(h^{i_\ell}, a) {=} \overbrace{ \textstyle\sum_{i_{\ell+1} \in C(h^{i_\ell} a)} \!\! \frac{n(h^{\prime, i_{\ell+1}}) }{n(h^{i_\ell} a)}\Big(\! \rho_{\ell+1}(b^{i_{\ell}}, a, b^{\prime, i_{\ell+1}})}^{\text{single immediate action}}{+}   \\
\overbrace{\!\!\!\!\!\!\!\!\!\! \underbrace{\textstyle \sum_{\color{red}{a' \in C(h^{\prime, i_{\ell+1}}})}\!\! \frac{n(h^{\prime, i_{\ell+1}}a')}{n(h^{\prime, i_{\ell+1}})}  \hat{Q}(h^{\prime, i_{\ell+1}}, a')}_{\hat{V}^{{\color{magenta}{\pi^\ast}}}(h^{\prime, i_{\ell+1}})}}^{\substack{ \text{different actions due to eq.~\eqref{eq:Expl}} \\ \text{approximating the best exploratory future tree policy $\pi^*$}}}\!\!\!\!\!\Big).
\end{gathered}
\end{equation}
The future policy highlighted by {\color{magenta}{magenta}} color is tree query dependent (See Fig.~\ref{fig:AssymetricPolicyTreeLace}).
In the same manner, the sets  $C(h^{i_{\ell}}a)$ and $C(h^{\prime, i_{\ell+1}})$ implicitly depend on the tree query number. One of our crucial insights in this paper is the summation over the actions in \eqref{eq:QEstApproxMCTSNORoll} marked by the {\color{red}{red}} color. This average can also be perceived as a stochastic policy. In {\bf finite} time  this summation can include unsafe actions in an unconstrained MCTS approach.  

\section{Problem Formulation and Rationale} 
\label{sec:PF}
We now proceed to our theoretical problem formulation. To reduce clutter we assume that the planning time index is zero. This is not an inherent limitation of our approach, every further relation can be easily modified to accommodate general planning time index. We endow the BMDP described in Section~\ref{seq:BMDP} with belief-dependent operator $\phi$ and obtain $$\langle\mathcal{B}, \mathcal{A}, \mathrm{T}_b, \underbrace{\rho}_{\substack{{\text{belief}}\\{\text{dependent}} \\ {\text{reward}}}}, \underbrace{{\color{nicegreen}{\phi}}}_{\substack{{\text{belief}}\\{\text{dependent}} \\ {\text{payoff}}}}, \gamma, b_0\rangle.$$
\subsection{Problem Formulation}
Our aim is to tackle the problem presented in \cite{Zhitnikov22arxiv} and \cite{Zhitnikov24TRO} narrowed to the multiplicative form of the inner constraint considering  a stochastic future  policy.   In  \cite{Zhitnikov22arxiv} and \cite{Zhitnikov24TRO} we presented our Probabilistic Constraint (PC) defined as such $\prob(c{=}1|b_0,  a_0, \pi){=}1$ where $c$ is a Bernoulli random variable.  In this work $c$ maps to one the event $\bigcap_{\ell=0}^{L} A^{\delta}_{\ell}$ such that the problem we want to solve is  
\begin{align}
&a^*_{0} \in \arg\max_{a_0 \in \mathcal{A}}  Q^{\pi^*}(b_0, a_0;\boldsymbol{\rho}_1) \text{ \ \ subject to} \label{eq:ConstrObj} \\
&\underbrace{\textstyle\prob\big(  \bigcap_{\ell=0}^{L} A^{\delta}_{\ell} \big|  b_0,  a_0, \pi^{\ast}_{1:L-1}  \big) {=}  1}_{\substack{{\text{outer}} \\ {\text{constraint}}}} \label{eq:OuterConstr} 
\end{align}
In this paper, we define the following sets as said 
$	A^{\delta}_{0} {\bydef} \left\{b_{0}{:}\phi(b_{0}) {\geq} \delta \right\} $
and for $\ell {\in} [1{:}L]$ the relevant set appears as  
\begin{equation}
\label{eq:Aset}	
\begin{gathered}
		A^{\delta}_{\ell} {\bydef} 
		\left\{{\color{nicegreen}{b^-_{\ell}}}, b_{\ell}{:} {\color{nicegreen}{b^-_{\ell} {\in} \mathcal{B}^-_{\ell}}}, b_{\ell} {\in} \mathcal{B}_{\ell}, {\color{nicegreen}{\phi(b^-_{\ell}) {\geq} \delta}}, \phi(b_{\ell}) {\geq} \delta \right\}.  
\end{gathered}
\end{equation}
One example of an operator $\phi$ is the probability to be safe given belief, specified as: 
\begin{align}
	&\phi(b_{\ell}) {=} \prob\big(\{x_{\ell} {\in} \mathcal{X}^{\mathrm{safe}}_{\ell}\}\big|b_{\ell}\big) {=} \mathbb{E}_{x_{\ell} \sim b_{\ell}}[\mathbf{1}_{\{x_{\ell} \in \mathcal{X}^{\mathrm{safe}}_{\ell}\}}] \label{eq:ProbSafeGivenBelief}\\
	&\phi(b^-_{\ell}) {=} \prob\big(\{x_{\ell} {\in} \mathcal{X}^{\mathrm{safe}}_{\ell}\}\big|b^-_{\ell}\big) {=} \prob\big(\{x_{\ell} {\in} \mathcal{X}^{\mathrm{safe}}_{\ell}\}\big|h^-_{\ell}\big). \label{eq:ProbSafeGivenPropagatedBelief} 
\end{align}
Here, $\mathcal{X}^{\mathrm{safe}}$ is the safe space, e.g.~the space where a robot can move without inflicting damage on itself. 
Therefore, we can think about the event  $\bigcap_{\ell=0}^{L} A^{\delta}_{\ell}$ as the {\bf Safe Belief Space}.

The  $\mathcal{B}^-_{\ell}$ and $\mathcal{B}_{\ell}$ in \eqref{eq:Aset}  are the reachable spaces in time $\ell$ of propagated beliefs $b^-_{\ell}$ and posteriors $b_{\ell}$ respectively. The reachable space in time $\ell$ is the space of all the beliefs in time $\ell$ that can be reached from a  belief given in planning session, using the  stochastic execution policy $\pi$ and changing the actions and the observations in \eqref{eq:Belief} and \eqref{eq:PropagatedBelief} accordingly. In our case, the belief given in planning session is $b_0$.  
By the green color in \eqref{eq:Aset} we highlight that we constrain the propagated beliefs in addition to the posteriors.
 
The probability of the event $ \bigcap_{\ell=0}^{L} A^{\delta}_{\ell}$ equals to the probability of the  event $\big(\mathbf{1}_{A^{\delta}_{0}}(b_0) \prod_{\ell=1}^{L} \!\! \mathbf{1}_{A^{\delta}_{\ell}}(b^-_{\ell}, b_{\ell})\big){=}1$.  
In this work, although we use Particle Filter (PF) as the belief update $\psi$ we do not take into account the stochasticity of the belief update operator as opposed to \cite{Zhitnikov22ai},\cite{Lim23jair} and treat $\psi$ operator as deterministic.  Since it would significantly complicate the paper,  we leave this aspect to the future work.   

One can extract the propagated belief from the belief update $\psi$, namely $\psi(b, a, z') {\bydef}\psi^{\text{post}}(\psi^{\text{prop}}(b,a), z')$.  Therefore,  to make the exposition clearer, from now on the indicator $\mathbf{1}_{A^{\delta}_{\ell}}(b_{\ell})$ depends solely on the posterior $b_{\ell}$ and not both the posterior $b_{\ell}$ and the propagated belief $b^-_{\ell}$. 
Note that in algorithms, for the sake of clarity, we make  the indicators dependent on both beliefs, propagated and posterior.    

The $\pi^*_{1:L-1}$ is the best future exploratory stochastic policy approximated by our probabilistically-constrained MCTS as we will further see. The approximation of the best future tree policy improves over time as proved by \cite{Auger13Sp} for an unconstrained problem. In our problem, instead of the best future stochastic tree policy, we have the best future stochastic probabilistically-constrained policy. This is because our PC is automatically enforced in future times due to its recursive nature, as we will see in Section~\ref{sec:Approach}.   From the discussion above and indicator properties, \eqref{eq:OuterConstr}  equals to 
\begin{equation}
	\label{eq:PCInner}
	\begin{gathered}  
		\prob\big(\underbrace{\textstyle\big(\mathbf{1}_{A^{\delta}_{0}}(b_{0})\! \prod_{\ell=1}^{L} \!\! \mathbf{1}_{A^{\delta}_{\ell}}(b_{\ell})\big){=}1}_{\substack{{\text{inner}} \\ {\text{constraint}}}} |  b_0, a_0, \pi  \big)=\\
		\textstyle\mathbb{E}^{\mathrm{T},\mathrm{O}}[\mathbf{1}_{A^{\delta}_{0}}(b_{0})\! \prod_{\ell=1}^{L} \!\! \mathbf{1}_{A^{\delta}_{\ell}}(b_{\ell}) |  b_0, a_0,  \pi ].
	\end{gathered}
\end{equation}
The outer condition \eqref{eq:OuterConstr} coupled with inner condition outlined by \eqref{eq:PCInner} says that with probability one (almost surely) future propagated and posterior beliefs $b^-$ and $b$, $L$ steps ahead,  will satisfy $\phi(b^-) {\geq} \delta$ and $\phi(b) {\geq} \delta$ correspondingly.  
	
Constraining the propagated belief \eqref{eq:ProbSafeGivenPropagatedBelief} means constraining on average (theoretical expectation) the posterior as discussed in the next section. 
\subsection{Implications of Constraining Propagated Belief}
\label{sec:ImplConstrProp}
In this section we shed light on the question what does it mean to constrain the propagated beliefs alongside with posterior beliefs. To cancel the constraining of the propagated beliefs one must redefine the set $A^{\delta}_{\ell}$ for every $\ell$  as follows
$$A^{\delta}_{\ell} {\bydef} 
\left\{\bcancel{b^-_{\ell}}, b_{\ell}{:} \bcancel{b^-_{\ell} {\in} \mathcal{B}^-_{\ell}}, b_{\ell} {\in} \mathcal{B}_{\ell}, \bcancel{\phi(b^-_{\ell}) {\geq} \delta}, \phi(b_{\ell}) {\geq} \delta \right\}.$$
 Further in the paper all the developments are valid for both versions of the set $A^{\delta}_{\ell}$.  The probability to be safe given a propagated belief equals to 
\begin{equation}
\label{eq:ConstrPropagated}
\begin{gathered}
\mathrm{P}\big(\{x_{\ell} {\in} \mathcal{X}^{\mathrm{safe}}_{\ell}\}\big|b^-_{\ell}\big){=}\mathrm{P}\big(\{x_{\ell} {\in} \mathcal{X}^{\mathrm{safe}}_{\ell}\}\big|h^-_{\ell}\big) {=}  \\
\textstyle \int_{{\color{nicegreen}{z_{\ell} \in \mathcal{Z}}}}\mathrm{P}\big(\{x_{\ell} {\in} \mathcal{X}^{\mathrm{safe}}_{\ell}\}\big|h^-_{\ell}, z_{\ell}\big)\probd(z_{\ell}|h^-_{\ell}) \mathrm{d}z_{\ell}=\\
\mathbb{E}_{z_{\ell}}[\mathrm{P}\big(\{x_{\ell} {\in} \mathcal{X}^{\mathrm{safe}}_{\ell}\}\big|h^-_{\ell}, z_{\ell}\big)|h^-_{\ell}]{=}\\
\mathbb{E}_{z_{\ell}}[\mathrm{P}\big(\{x_{\ell} {\in} \mathcal{X}^{\mathrm{safe}}_{\ell}\}\big|b_{\ell}\big)|b^-_{\ell}].
\end{gathered}
\end{equation}  
The theoretical expectation in \eqref{eq:ConstrPropagated} is out of the reach. Yet we evaluate it using the propagated belief $b^{-}(h^-)$. Defining the set $A^{\delta}_{\ell}$ as \eqref{eq:Aset}, with the propagated beliefs, allows to account for all the possible posterior beliefs in \eqref{eq:ConstrPropagated}.  Additionally, we know that $\forall \epsilon {>}0$
\begin{equation}
	\label{eq:ConvergenceInP}
	\begin{gathered}
		\textstyle\lim\limits_{|C(h^{-}_{\ell})|\to \infty}\prob\big(|\mathrm{P}\big(\{x_{\ell} {\in} \mathcal{X}^{\mathrm{safe}}_{\ell}\}\big|h^-_{\ell}\big) -\\
		\textstyle \frac{1}{|C(h^-_{\ell})|}\sum_{z^l_{\ell} {\in} C(h^-_{\ell})} \mathrm{P}\big(\{x_{\ell} {\in} \mathcal{X}^{\mathrm{safe}}_{\ell}\}\big|h^-_{\ell},z^{l}_{\ell}\big) |{>}\epsilon\big| h^{-}_{\ell}\big) {=}0.
	\end{gathered}
\end{equation}
With a slight abuse of notation, $C(h^-_{\ell})$ is now a list of the enumerated observations that are children of $h^-_{\ell}$.
Equation \eqref{eq:ConvergenceInP} means that for any arbitrary small error $\epsilon$, the difference between  \eqref{eq:ConstrPropagated}  and its approximation by the children of $h^-_{\ell}$ tends to zero as the number of children of $h^-_{\ell}$ grows.  
\begin{thm}[Necessary  condition for entire observation space $\mathcal{Z}$ of children of $h^-_{\ell}$ to be safe]
\label{thm:Cond}	
Fix $\delta {\in} [0,1]$ and assume that
\begin{align}
	\mathrm{P}\big(\{x_{\ell} {\in} \mathcal{X}^{\mathrm{safe}}_{\ell}\}\big|h^-_{\ell}\big) {\geq} \delta.  \label{eq:Necessary}
\end{align}
Eq.~\eqref{eq:Necessary} is a necessary condition for the entire observation space $\mathcal{Z}$ of children of $h^-_{\ell}$ to be safe. To rephrase that 
\begin{align}
	\mathrm{P}\big(\{x_{\ell} {\in} \mathcal{X}^{\mathrm{safe}}_{\ell}\}\big|h^-_{\ell}\big) {<} \delta
\end{align}
implies that $\exists b_{\ell}(h_{\ell})$ a child of $h^-_{\ell}$ which is not safe, namely, $\mathrm{P}\big(\{x_{\ell} {\in} \mathcal{X}^{\mathrm{safe}}_{\ell}\}\big|h^-_{\ell}, z_{\ell}\big) {<} \delta$.
\end{thm}
See  Appendix~\ref{proof:Cond} for  a detailed proof. 
We still need to check the children posteriors $\{z^l_{\ell}\}_{l=1}^{|C(h^-_{\ell})|}$. This is because the condition \eqref{eq:Necessary} is only necessary and not sufficient.  In other words, if for all the children $\forall z_{\ell} {\in} \mathcal{Z}$ of $h^-_{\ell}$,  it holds that $\mathrm{P}\big(\{x_{\ell} {\in} \mathcal{X}^{\mathrm{safe}}_{\ell}\}\big|h^-_{\ell}, z_{\ell}\big) 
{\geq} \delta$ it has to be that  $\mathrm{P}\big(\{x_{\ell} {\in} \mathcal{X}^{\mathrm{safe}}_{\ell}\}\big|h^-_{\ell}\big) 
{\geq} \delta$.  Since the condition is not sufficient we cannot say that $\mathrm{P}\big(\{x_{\ell} {\in} \mathcal{X}^{\mathrm{safe}}_{\ell}\}\big|h^-_{\ell}\big) {<} \delta$ implies that  $\forall b_{\ell}(h_{\ell})$ that are children of $h^-_{\ell}$ it will hold that $\mathrm{P}\big(\{x_{\ell} {\in} \mathcal{X}^{\mathrm{safe}}_{\ell}\}\big|h_{\ell}\big) {<} \delta$. 

Note that if  for every sampled observation  $\mathrm{P}\big(\{x_{\ell} {\in} \mathcal{X}^{\mathrm{safe}}_{\ell}\}\big|h^-_{\ell}, z^l_{\ell}\big){\geq} \delta$, it implies that
\begin{align}
	\textstyle\Big(\frac{1}{|C(h^-_{\ell})|} \sum_{l=1}^{|C(h^-_{\ell})|}\mathrm{P}\big(\{x_{\ell} {\in} \mathcal{X}^{\mathrm{safe}}_{\ell}\}\big|h^-_{\ell},z^l_{\ell}\big) \Big) {\geq} \delta. \label{eq:SampleMeanChildren}
\end{align}
To conclude, by constraining the propagated belief, we constrain the theoretical expectation of the posteriors given $h^-_{\ell}$, and by constraining each posterior we also constrain its sample approximation portrayed by Eq.~\eqref{eq:SampleMeanChildren}. Without constraining the propagated belief, if the number of children of $b^-_{\ell}(h^-_{\ell})$ is small, namely, $|C(h^-_{\ell})|$ is small, we anticipate poor robot's safety in execution of the best action found by  our planner (e.g. number of collisions). This is  because  constraining the propagated belief allows to account in expectation for all the observations in the observation space, and not only the sampled observations. This will happen if the number of MCTS tree queries is small.

It is possible that other definitions of safety of the beliefs can be utilized. 
While this is outside the scope of this paper,  we specified relevant operators $\phi$ in the Appendix, Section~\ref{sec:VaRCVaR}.
\\
\textbf{Remark:} To assure feasibility of our PC \eqref{eq:OuterConstr} at the limit of MCTS convergence,  the robot has to have a bounded support of the belief $b_0$ and bounded motion models. 
If we deal with a particle based representation of $b_0$ 
we perceive the particles as true robot positions, so it is left only to assure that the motion model is bounded. This is, however, natural since the robot cannot have limitless actuators. 
\section{PC-MCTS (anytime approach)}
\label{sec:Approach}
Our constraint depends on a stochastic policy. Similar to the objective \eqref{eq:ValueStochPolicy}  in our PC we land at the following result.
\begin{thm}[Representation of PC, recursive form] 
	\label{thm:Represent}
	The PC defined by \eqref{eq:PCInner} conforms to the following recursive form.  
	\begin{equation}
		\label{eq:MultInnerConstrReformRecursiveStochPolicy}
		\begin{gathered}
			\mathbf{1}_{A^{\delta}_{0}}(b_{0})\underset{b_1}{\mathbb{E}} \Big[ \mathbf{1}_{A^{\delta}_{1}} \underset{a_1}{\mathbb{E}} \Big[\underset{b_2}{\mathbb{E}}\Big[\mathbf{1}_{A^{\delta}_{2}}  \dots \\
			\mathbb{E}\big[\mathbf{1}_{A^{\delta}_{L}}| b_{L-1}, a_{L-1}\big] \dots \Big| b_1, a_1\Big]\Big| b_1, \pi_1, \Big| b_0, a_0\Big]{=}\!\!\!\!\!\!\!\!\\
			\mathbf{1}_{A^{\delta}_{0}}(b_{0})\underset{b_1}{\mathbb{E}}\Big[\underset{a_1{\sim}\mathbb{P}^{\pi}_{1}(a_1|b_1)}{\mathbb{E}}\Big[\\
			\!\!\!\!\!\! \prob\big(\big(\prod_{\ell=1}^{L} \mathbf{1}_{A^{\delta}_{\ell}}(b_{\ell})\big){=}1  \Big| b_1, a_1, \pi \Big)\Big|b_1, \pi_1\Big]  \Big|b_0, a_0\Big]. \!\!\!\!\!\!\!\!\!
		\end{gathered}
	\end{equation}
\end{thm}
\noindent We provide a detailed proof in the Appendix, Section~\ref{sec:Represent}.

%
In this section, we present our anytime safety approach.  
To invalidate the sample approximation of \eqref{eq:OuterConstr} it is sufficient that a single belief (propagated or posterior) in the belief tree fails to be safe and the corresponding indicator is zero. In our methodology, we leverage the classical iterative MCTS scheme of descending down the search tree of histories and ascending back to the root (Section~\ref{sec:mcts}). 
Once on the way down the tree an unsafe belief is encountered, we know that the PC enforced from each predecessor belief node is violated. We delete such an action from the search tree and fix the $\hat{Q}$ above. Let us delve into the details. 

Suppose the MCTS is configured to run without rollout.
Would we construct the estimated counterpart of \eqref{eq:MultInnerConstrReformRecursiveStochPolicy} from the belief tree constructed by MCTS our PC would be as such 
\begin{equation}
	\label{eq:MultInnerConstrSampleApproxStochPolicy}
	\begin{gathered}
		\!\!\!\!\!\!\!\!\! \textstyle \Big( \mathbf{1}_{A^{\delta}_{0}}(b_{0})  \sum_{i_{1} {\in} C(b_0a_0)} \!   \frac{\mathbf{1}_{A^{\delta}_{1}}(b^{i_{1}}_{1})}{|C(b_0a_0)|} \\  
		\textstyle\sum_{a_1 {\in} C(h^{i_{1}}_1)}\!\! \frac{n(h^{i_{1}}_{1}a_1)}{n(h^{i_{1}}_1)}\sum_{i_{2} {\in} C(h^{i_1}_1a_1)}  \frac{\mathbf{1}_{A^{\delta}_{2}}(b^{i_{2}}_{2})}{|C(h_1^{i_1}a_1)|}\cdots\!\!\!\!\!\!\!\!\!\!\!\!\!\!\!\!\!\!\!\!   \\ 
		\!\!\!\!\!\!\!\! {\cdots} 
		\textstyle\frac{\mathbf{1}_{A^{\delta}_{L-1}}(b^{i_{L-1}}_{L-1})}{|C(h^{i_{L-2}}_{L-2}a_{L-2})|}  
		\sum_{a_{L-1} {\in} C(h^{i_{L-1}}_{L-1})}\!\!\!\!\! \frac{n(h^{i_{L-1}}_{L-1}a_{L-1})}{n(h^{i_{L-1}}_{L-1})}\!\!\!\!\!\!\!\! \\ 
		\textstyle\sum_{i_{L} {\in} C(h^{i_{L-1}}_{L-1}a_{L-1})} \frac{\mathbf{1}_{A^{\delta}_{L}}(b^{i_{L}}_{L})}{|C(h^{i_{L-1}}_{L-1}a_{L-1})|}   \Big) {=} 1.
	\end{gathered}
\end{equation}
Since our constraint is defined using an indicator,  \eqref{eq:MultInnerConstrSampleApproxStochPolicy} translates to the fact that \emph{each} lace defined by the actions and the observations on the way down the tree shall consist of safe beliefs.

To emphasize that each belief in the search tree has a single parent and the  corresponding parent is attainable, let us introduce yet another notation    $b^{i_{\ell}|i_{\ell-1}}_{\ell}$. This means that the belief $b^{i_{\ell}|i_{\ell-1}}_{\ell}$ has global index $i_{\ell}$ and parent belief has global index $i_{\ell-1}$.
On the way down the tree we ensure that 
\begin{equation}
	\label{eq:MultInnerConstrSampleApproxISParentIndexLace}
	\begin{gathered}
		\!\!\!\!\!\!\!\!\Big(\mathbf{1}_{A^{\delta}_{0}}(b^1_{0}) \mathbf{1}_{A^{\delta}_{1}}( b^{i_{1}|1}_{1}) 
		\mathbf{1}_{A^{\delta}_{2}}(b^{i_{2}|i_1}_{2})
		\cdots  \\
		\mathbf{1}_{A^{\delta}_{L-1}}( b^{i_{L-1}|i_{L-2}}_{L-1})\mathbf{1}_{A^{\delta}_{L}}( b^{i_{L}|i_{L-1}}_{L}) \Big){=}1,	  \!\!\!\!\!\!\!\!\!\!\!\!\!\!
	\end{gathered}
\end{equation}
where the actions along the lace are 
$a_1 {\in} C(h^{i_1}_1), a_2 {\in} C(h^{i_2}_2), \dots, a_{L-1} {\in} C(h^{i_{L-1}}_{L-1}) $ and the beliefs are according to the observations indexed by $i_2 \in C(h^{i_1}_1a_1), i_3 {\in} C(h^{i_2}_2a_2), \dots, i_L {\in} C(h^{i_{L-1}}_{L-1}a_{L-1}) $.  
In other words we require that every propagated and posterior belief along the lace would be safe. \\
{\bf Remark} The equivalence of  \eqref{eq:MultInnerConstrSampleApproxStochPolicy} and the fact that every lace in search tree shall be safe is a  property of our PC formulation, e.g., this is no happening in case of popular Chance Constraint \cite{Moss24Arxiv}.

Note that in \eqref{eq:MultInnerConstrSampleApproxStochPolicy} we do not have distributional shift due to progressive widening of observations (and the fact that not in every tree query a new belief node is introduced) as opposed to the objective \eqref{eq:QEstApproxMCTSNORoll}. This is because we do not take into account the statistics dictated by the visitation counts of the observations.

As we see from \eqref{eq:MultInnerConstrSampleApproxStochPolicy}, the recursive form portrayed by \eqref{eq:MultInnerConstrReformRecursiveStochPolicy} transfers to MCTS estimator. For clarity of the exposition let us denote the product of the indicators in the inner constraint \eqref{eq:PCInner} by $c$ depending on the current and future beliefs. For example, at the root of the belief tree we have $c(b_{0:L}){=}\mathbf{1}_{A^{\delta}_{0}}(b_{0})\! \prod_{\ell=1}^{L} \!\! \mathbf{1}_{A^{\delta}_{\ell}}(b_{\ell})$. 
By design, \eqref{eq:MultInnerConstrReformRecursiveStochPolicy} and \eqref{eq:MultInnerConstrSampleApproxStochPolicy} equals one if and only if, the PC starting from each belief action node $ha$ in the tree is satisfied, namely $ \prob\big(c{=}1 |  b(h), a, \pi  \big){=}1$.
We now define the notion of dangerous action in belief tree.  

\noindent{\bf Remark}: We call an action dangerous if it is {\bf believed} to be dangerous. Meaning our notion of dangerous or safe actions based on beliefs and not the possible POMDP states as in Chance Constraint \cite{Santana16aaai}. 
\begin{defn}[Dangerous action] \label{def:DA}
	A dangerous action is action $a$ in a place $h$ in a search tree  that renders an estimator of \eqref{eq:MultInnerConstrReformRecursiveStochPolicy} smaller than one, namely $ \hat{\prob}\big(c{=}1 |  b_0, a_0, \pi  \big){\color{red}<}1$, where the estimator is as in \eqref{eq:MultInnerConstrSampleApproxStochPolicy}.
\end{defn}
Note that best stochastic future tree policy is dependent on the number of performed tree queries. 
%
\begin{cor} \label{cor:SafeAction}
	Each action in a search tree can be dangerous or safe. We define {\bf safe} action $a$ (or $a_0$) to be the action that is {\bf not} dangerous, namely $ \hat{\prob}\big(c{=}1 |  b_0, a_0, \pi  \big){\color{nicegreen}=}1$ and safe under the safe future tree policy, namely $ \hat{\prob}\big(c{=}1 |  b(h), a, \pi \big){\color{nicegreen}=}1$ for arbitrary future history $h$ as a result of mentioned safe policy. 
\end{cor}	
Let us reiterate that we build the search tree solely from the safe actions. 
Effectively, using our pruning  and fixing the values and statistics maintained by the search, to be explained shortly,  we assure preemptively that the sample approximation of \eqref{eq:MultInnerConstrReformRecursiveStochPolicy} defined by \eqref{eq:MultInnerConstrSampleApproxStochPolicy}  using the beliefs from the search tree built by MCTS equals to one. 
To assure that the \eqref{eq:MultInnerConstrSampleApproxStochPolicy} equals one it is required that every indicator function within is one. This is our mechanism to 
%
%
assure that in {\bf any} finite time the search tree contains only {\bf safe} actions as opposed to duality based methods where the constraint is satisfied only at the convergence limit, namely in infinite time (see Section \ref{sec:DualityBasedApproach}).   
When a newly sampled belief renders the corresponding indicator equal one, we add it to the belief tree. If the indicator is zero, we develop a mechanism to delete an action and fix the search tree upwards.  
\subsection{Pushing Forward in Time Only the Safe Trajectories}
\label{sec:IS}
%
Even if  \eqref{eq:MultInnerConstrSampleApproxStochPolicy} equals one, meaning every indicator inside equals one, when $\delta {<} 1$ and payoff operator as in \eqref{eq:ProbSafeGivenBelief}, it is possible that there  exist samples that are unsafe, e.g., falling inside an obstacle or a dangerous region. 
If the robot is operational it means the robot was safe before it commenced an action. Thus, we shall discard the unsafe portion of the belief before we update the belief with action and observation (barring the situation when $\delta{=}1$ and payoff operator as in \eqref{eq:ProbSafeGivenBelief} and \eqref{eq:ProbSafeGivenPropagatedBelief}).  
We define $\bar{b}^{\mathrm{safe}}$ as the belief constituted only by the safe particles, namely conditioned on the history and the events $\{x {\in} \mathcal{X}^{\mathrm{safe}}\}$. Such a belief is given by
\begin{align}
	\bar{b}^{\mathrm{safe}}_{\ell}(x_{\ell}) {=} \textstyle\probd(x_{\ell}| {\color{nicegreen}{b_{0}}}, a_{0:\ell-1}, z_{1:\ell}, \bigcap_{i=0}^{\ell} \{x_{i} {\in} \mathcal{X}^{\mathrm{safe}}_{i}\}). \label{eq:SafeBelief}
\end{align}
To convert $\bar{b}$ to $\bar{b}^{\mathrm{safe}}$ we remove not safe particles and resample with replacement the safe ones to the initial size.   
This means that the beliefs and observations in \eqref{eq:MultInnerConstrReformRecursiveStochPolicy} will be not as in objective \eqref{eq:ConstrObj} but as follows. 
We define $\bar{b}$ as the belief obtained by percolating forward in time belief that has been made safe sequentially, that is $\bar{b}'{=}\psi(\bar{b}^{\mathrm{safe}},a,z')$ where
\begin{align}
\bar{b}_{\ell}(x_{\ell}) {=} \textstyle\probd(x_{\ell}| {\color{nicegreen}{b_{0}}}, a_{0:\ell-1}, z_{1:\ell}, \bigcap_{i=0}^{{\color {red}{\ell-1}}} \{x_{i} {\in} \mathcal{X}^{\mathrm{safe}}_{i}\}), \label{eq:UpdatedFromSafeBelief}
\end{align}
and the belief propagated only with action and without an observation 
\begin{align}
\bar{b}_{\ell}^{{\color{red}{-}}}(x_{\ell}) {=} \textstyle \probd\big(x_{\ell}| {\color{nicegreen}{b_{0}}}, a_{0:\ell-1}, z_{1:{\color{red} \ell-1}}, \bigcap_{i=0}^{{\color{red}\ell-1}}\{x_{i}{\in} \mathcal{X}^{\mathrm{safe}}_{i}\}\big). \label{eq:PropagatedFromSafeBelief}
\end{align}
Both beliefs are generally unsafe. Our BMDP tuple is now augmented with another space of beliefs $\bar{\mathcal{B}}$ defined by \eqref{eq:UpdatedFromSafeBelief}. We have now  $$\langle\mathcal{B}, \underbrace{\bar{\mathcal{B}}}_{\substack{{\text{space}}\\{\text{of the beliefs}} \\ {\text{defined by \eqref{eq:UpdatedFromSafeBelief}}}}}, \mathcal{A}, \mathrm{T}_b, \underbrace{\rho}_{\text{reward}}, \underbrace{{\color{nicegreen}{\phi}}}_{\text{payoff}}, \gamma, b_0\rangle.$$ At this point we need to define another safe set $$\bar{A}^{\delta}_{\ell} {=} \{\bar{b}^-_{\ell},\bar{b}_{\ell}{:} \bar{b}^-_{\ell} {\in} \bar{\mathcal{B}}^-_{\ell}, \bar{b}_{\ell} {\in} \bar{\mathcal{B}}_{\ell}, \phi(\bar{b}^-_{\ell}) {\geq} \delta, \phi(\bar{b}_{\ell}) {\geq} \delta\}.$$ 
Here the $\bar{\mathcal{B}}^-_{\ell}$ and $\bar{\mathcal{B}}_{\ell}$ are reachable by changing observations in \eqref{eq:UpdatedFromSafeBelief} and \eqref{eq:PropagatedFromSafeBelief} spaces. Do note that only in time $0$ the set $A^{\delta}_0 {=} \bar{A}^{\delta}_0$. This is because in inference we know that robot is still operational. 
We always make safe the actual robot belief $b_0$.  
In planning, the belief is rendering the observation in the next time step (Fig.~\ref{fig:MakingBeliefSafe}). Thus in both CC and PC in time $\ell{+}1$, the observation PDF reads $\probd(z_{\ell+1}|\bar{b}^{\mathrm{safe}}_{\ell}, a_{\ell})$, whereas in objective we have $\probd(z_{\ell+1}|b_{\ell}, a_{\ell})$. We sample from the latter and the normalized ratios of these likelihoods 
\begin{align}
w^{a_{\ell}}_{\ell+1} \propto \nicefrac{\probd(z_{\ell+1}|\bar{b}^{\mathrm{safe}}_{\ell}, a_{\ell})}{\probd(z_{\ell+1}|b_{\ell}, a_{\ell})}
\end{align}
are the weights in the equation \eqref{eq:MultInnerConstrSampleApproxISStochPolicy}. Using Importance Sampling in such a way, we construct a single belief tree. However, for the constraint calculation we use the $\bar{b}$ corresponding to the belief $b$, please see Fig.~\ref{fig:MakingBeliefSafe}.   
The  Eq.~\eqref{eq:PCInner} transforms into 
\begin{equation}
	\prob\big(\textstyle\big(\mathbf{1}_{A^{\delta}_{0}}(b_{0})\! \prod_{\ell=1}^{L} \!\! \mathbf{1}_{\bar{A}^{\delta}_{\ell}}(\bar{b}_{\ell})\big){=}1|  b_0, a_0, \pi  \big).
\end{equation}
Let us reiterate, on the way down the tree, we ensure that every belief along the lace lightens up its indicator. Similar to  \eqref{eq:MultInnerConstrSampleApproxStochPolicy}, we ensure that under the stochastic policy approximated by the MCTS, the PC is satisfied.  We have that 
\begin{equation}
\label{eq:MultInnerConstrSampleApproxISStochPolicy}
\begin{gathered}
\Big(\mathbf{1}_{\bar{A}^{\delta}_{0}}(b_{0}) \textstyle  \sum_{i_{1} \in C(b_0a_0)} \! w^{a_0, i_1}_1 \mathbf{1}_{\bar{A}^{\delta}_{1}}(\bar{b}^{i_{1}}_{1}) \sum_{a_1 {\in} C(h^{i_{1}}_1)}\!\!\! \frac{n(h^{i_{1}}_{1}a_1)}{n(h^{i_{1}}_1)}\!\!\!\!\!\!\!\!\!\! \\
\textstyle\sum_{i_{2} {\in} C(h^{i_1}_1a_1)}   w_2^{a_1, i_2}\mathbf{1}_{\bar{A}^{\delta}_{2}}(\bar{b}^{i_{2}}_{2}){\cdots} \!\!\!\!\!\!\!\!\!\!\!\!\!\!\!\!\\
{\cdots} \textstyle\mathbf{1}_{\bar{A}^{\delta}_{L-1}}( \bar{b}^{i_{L-1}}_{L-1}) \sum_{a_{L-1} {\in} C(h^{i_{L-1}}_{L-1})}\!\!\!\!\! \frac{n(h^{i_{L-1}}_{L-1}a_{L-1})}{n(h^{i_{L-1}}_{L-1})} \!\!\!\!\!\!\!\!\!\!\!\!\!\!\!\!\\
\textstyle\sum_{i_{L} {\in} C(h^{i_{L-1}}_{L-1}a_{L-1})} \!\! w^{a_{L-1},i_L}_L\mathbf{1}_{\bar{A}^{\delta}_{L}}( \bar{b}^{i_{L}}_{L})\Big){=}1.	 \!\!\!\!\!\!\!\!\!\!\!\!\!\!\!\!
\end{gathered}
\end{equation}
Further in this paper we assume that the observation model $\mathrm{O}(z , x)$ has infinite support, to rephrase that  $\{z {\in} \mathcal{Z}, x {\in} \mathcal{X} {:}   \mathrm{O}(z , x) {>} 0\} {=} \mathcal{Z}{\times} \mathcal{X}$. This assumption ensures that there are no nullified weights in \eqref{eq:MultInnerConstrSampleApproxISStochPolicy}. Further, since the weights in  \eqref{eq:MultInnerConstrSampleApproxISStochPolicy} are normalized and all of them are nonzero, even a single weight missing  because the inner constraint is violated, renders \eqref{eq:MultInnerConstrSampleApproxISStochPolicy} smaller than one. This means that the constraint with respect to the root $b_0$ of the belief tree  is not satisfied. Since the weights are selfnormalized per action, to verify that \eqref{eq:MultInnerConstrSampleApproxISStochPolicy} equals to one we do not need to calculate weights at all. In fact, we never check the whole PC approximation. In contrast, as we already mentioned we only verify that each indicator equals to one on the way down the tree.

\subsection{Bounded Support  Motion Model}
\begin{figure}[t]
	\centering 
	\includegraphics[width=0.25\textwidth]{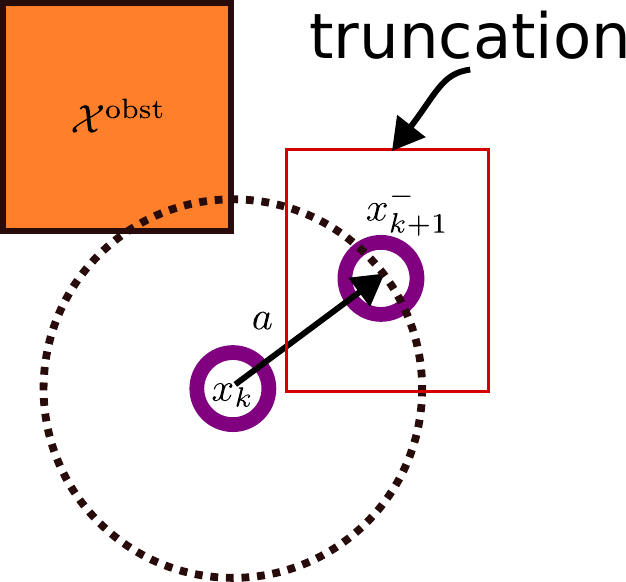}
	\caption{Illustration of the effect of truncation of motion model T.}
	\label{fig:TruncatingT}
\end{figure}
Suppose $\bar{b}^{\mathrm{safe}}$ is represented by the finite set of particles, belief update $\psi$ is a PF. If motion model has a support encapsulating the whole space $\mathbb{R}^{d}$, where $d$ is a dimension, for every possible action it will be eventually unsafe belief. This is because eventually, at the limit of MCTS, for every tried action it will be non safe belief. However, we know that robot cannot teleport and truncation of motion model is, therefore,  natural. In fact it is assumed often times to be Gaussian to bring infinite support to the table to alleviate the complexity of the solution. Observe Fig.~\ref{fig:TruncatingT}.  Without truncation for every action $a$ it is possible that the propagated sample will be unsafe and render next in time posterior belief also unsafe.

Using our further presented method we build a tree solely from safe actions.  Do note that all our algorithms can be run with various belief dependent operators. 
It is customary to maintain a pair of posterior beliefs $\bar{b}$ and $b$ as visualized in Fig.~\ref{fig:MakingBeliefSafe} or just maintain a single belief $b$. Further we stick to the former scheme as in Fig.~\ref{fig:MakingBeliefSafe}.

\subsection{Constraint Violation and Efficient Tree Cleaning}
\label{sec:pruning}
Before we begin this section we must clarify that from now on we slightly change the notations in text and the algorithms. To recap, we use following variables: 
$h$ represents a history $\{b, a_0, z_1, \dots a_{\ell-1}, z_\ell\}$, and  $haz'$ is the shorthand for history with $a$ and $z'$  appended to the end. In a similar manner, as mentioned earlier, $ha$ is the history $h$ with action $a$ appended to the end.    
$C(ha)$ is a set of the children of a belief-action node $ha$. Each such a child, now, is a triple of observation, reward, and posterior belief $\{z', r', b'\}$. 

\begin{figure*}[t]
	\centering
	\begin{minipage}[t]{0.33\textwidth}
		\centering 
		\includegraphics[width=\textwidth]{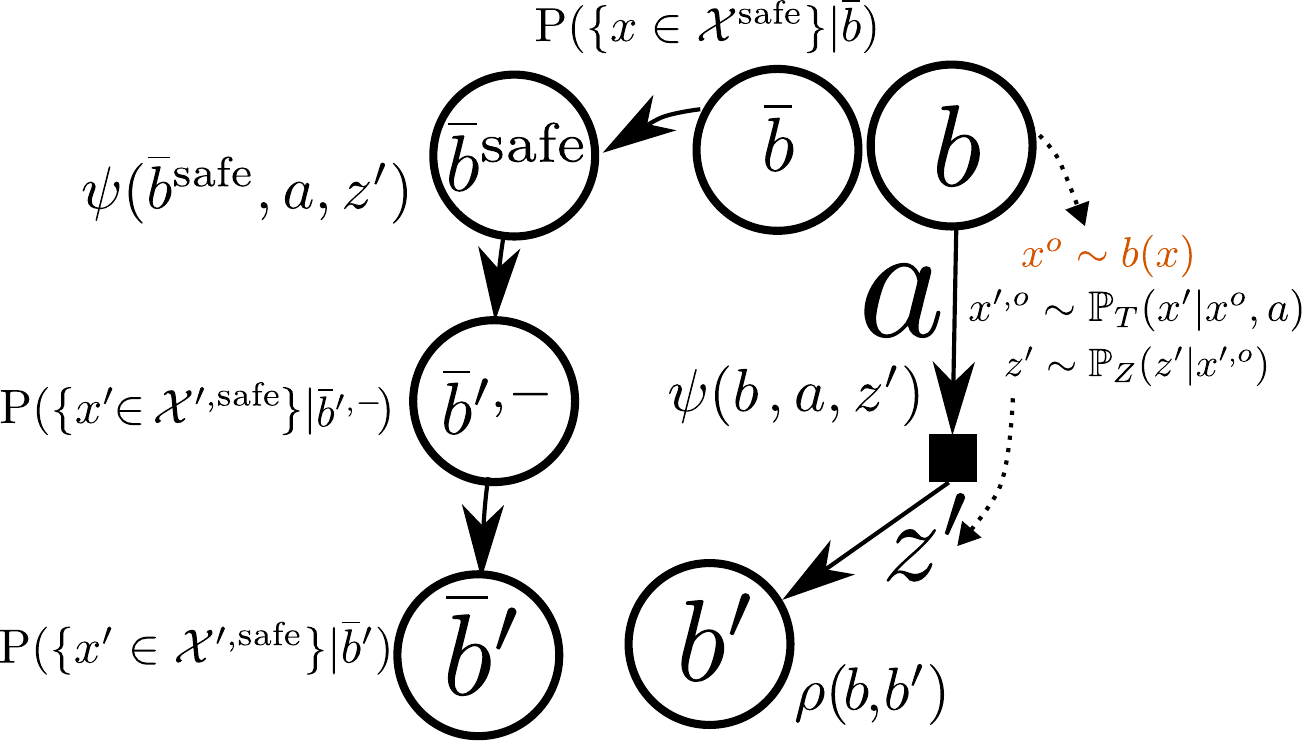}
		\subcaption{}
		\label{fig:MakingBeliefSafe}
	\end{minipage}%
	\hfill
	\begin{minipage}[t]{0.33\textwidth}
		\centering 
		\includegraphics[width=\textwidth]{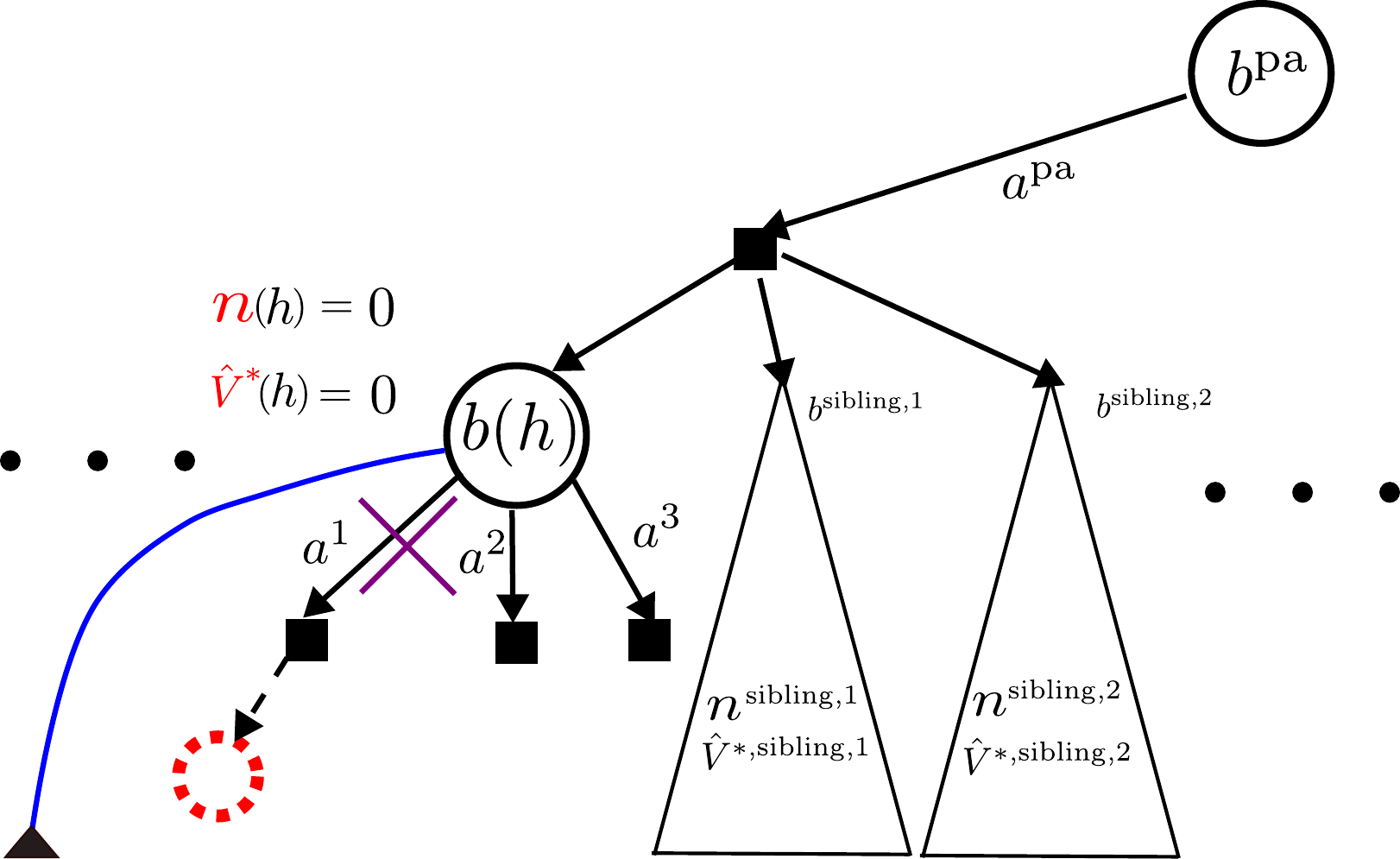}
		\subcaption{}
		\label{fig:CleanTreeEasy}
	\end{minipage}%
	\hfill
	\begin{minipage}[t]{0.33\textwidth}
		\centering 
		\includegraphics[width=\textwidth]{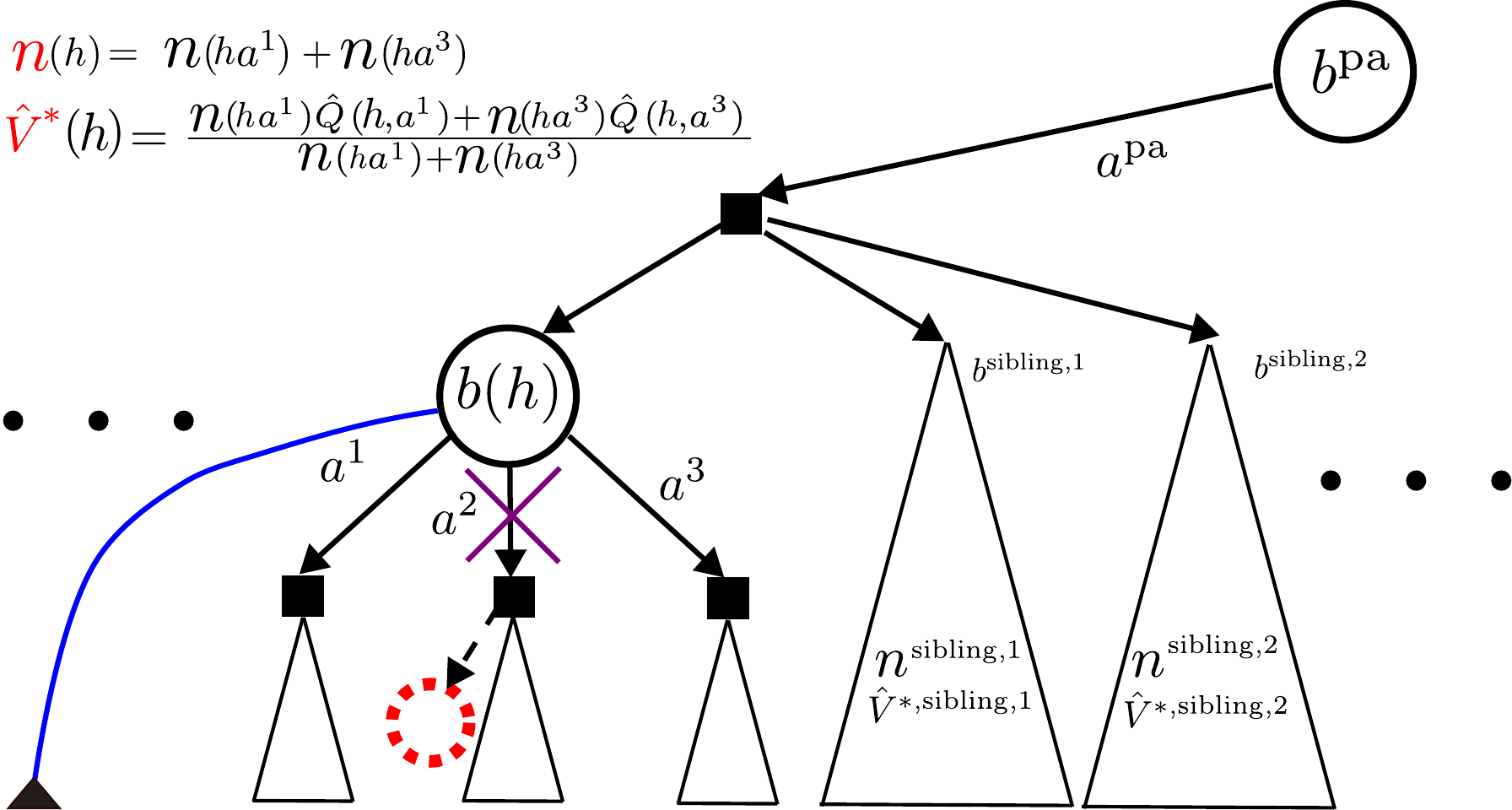}
		\subcaption{}
		\label{fig:CleanTreeHard}
	\end{minipage} 
	\caption{\textbf{(a)} Conceptual illustration of maintaining a pair of posterior beliefs, $b$ and $b'$ are used for the reward operator, while $\bar{b}$, $\bar{b}^{\prime,-}$ and $\bar{b}'$ for the safety operator. We create observation using belief $b$ highlighted by the {\color{nicebrown} brown} color. In the subfigures \textbf{(b)} and \textbf{(c)} we visualize the cleaning of the belief tree in case that new belief violated the inner constraint. By {\color{blue} blue } color we mark elements related to \emph{optional} rollout.  By the thick {\color{red} red} dashed circle we mark a newly expanded belief node. \textbf{(b)} In this scenario the belief node is a first child expanded from the first selected action $a^1$. The expanded belief violates the inner constraint. Thus we need to prune action $a^1$. Because we have not updated the visitation count of $b$ yet, we only need to delete action $a^1$. \iffalse The crossed values are still not present within $n(h^{\mathrm{pa}}a^{\mathrm{pa}})$ and $\hat{Q}(h^{\mathrm{pa}}a^{\mathrm{pa}})$ \fi \textbf{(c)}  Harder scenario for cleaning the tree.  Here we need to perform appropriate fixes to the action-value-estimates after we prune $a^2$. Note that UCB or PUCT, tries each action from each belief infinitely many times.}      
	\label{fig:CleanTree}	 
\end{figure*}
We now explain how we prune all dangerous actions (Def.~\ref{def:DA}) from the search tree and thereby our search tree always contains only the safe actions (Cor.~\ref{cor:SafeAction}). Actions at the root and tree future policy, which is stochastic due to exploration, are such that the PC is fulfilled starting from each belief node in the search tree.  
Suppose that our Probabilstically Constrained MCTS (Alg.~\ref{alg:PCMCTS} and ~\ref{alg:PCMCTSPUCT}) is currently at  a belief node $b(h)$ in the belief tree, with a corresponding history $h$ defining the unique place $h$ in the belief tree. 
The algorithm selects an action according to \eqref{eq:Expl} and suppose it creates a new belief. 
Every time we create a new belief node to be added to the search tree  we obtain $b'$ for the reward calculation and corresponding  $\bar{b}^{\prime,-}$ and $\bar{b}'$ for the constraint. We then check if $\phi(\bar{b}^{\prime,-}) {\geq} \delta$ and $\phi(\bar{b}') {\geq} \delta$ and if both inequalities are satisfied we add the newly created node to belief tree.    If $\phi(\bar{b}^{\prime,-}) {<} \delta$ or $\phi(\bar{b}') {<} \delta$, we shall prune an action leading to this belief node from the belief tree and fix the $\hat{Q}$ upwards since the laces emanating from the cleaned action participate in $\hat{Q}$ of every ancestor belief action node.  
Due to the fact that we assemble $\hat{Q}$ at each belief-action node from laces, at node $h$ it holds that
%
$\hat{V}^*(h) {=} \textstyle \sum_{a \in C(h)} \frac{n(ha) \hat{Q}(ha)}{n(h)}$ and $n(h) {=} \sum_{a \in C(h)} n(ha). $
%
The $\hat{Q}$ of the parent reads 
\begin{equation}
\label{eq:QestWithRollout}
\hat{Q}(h^{\mathrm{pa}}a^{\mathrm{pa}}) {=} \textstyle\frac{({\color{red}{n}}(h) {\color{blue}{+ 1}})\big(\rho(b^{\mathrm{pa}},a^{\mathrm{pa}},b)+ {\color{red}{\hat{V}^*}}(h)+{\color{blue} \mathrm{roll}}(h)\big)+\dots}{n(h^{\mathrm{pa}}a^{\mathrm{pa}})}.
\end{equation}
where the summation over all the sibling subtrees and the visitation count appears as 
%
$n(h^{\mathrm{pa}}a^{\mathrm{pa}}) {=} {\color{red}{n}}(h) {\color{blue}{+ 1}} + n^{\mathrm{sibling},1}{\color{blue}{+ 1}} \dots $,
%
where by the red {\color{red} {red}} color we denote values of the current belief node and by the {\color{blue} {blue}} color we denote optional values related to the activation of the rollout. 
We now turn to an explanation of how to clean the tree efficiently using subtraction and adding operations instead of assembling action-value estimates and visitation counts from scratch using updated values down the tree. Suppose the action leading to the newly added belief does not have sibling subtrees corresponding to another actions and this belief is the first child of such an action, as depicted in Fig.~\ref{fig:CleanTreeEasy}. In this case the visitation count $n(h)$ and $\hat{V}^*(h)$  are not present yet within $\hat{Q}(h^{\mathrm{pa}}a^{\mathrm{pa}})$. This is because the only rollout was commenced from $b(h)$.  We can just delete the action leading to the newly created belief node.    

We, now, focus on a more interesting setting depicted in  Fig.~\ref{fig:CleanTreeHard}. After we deleted the subtree defined by the belief node $b(h)$ and action $a$ ($a^2$ in Fig.~\ref{fig:CleanTreeHard}),  we need to update the visitation count $n(h)$ as such $n(h) {\leftarrow} n^{\mathrm{intree}}(h) {-} n^{\mathrm{intree}}(ha)$, where we denote values that are currently in the belief tree by $\square^{\mathrm{intree}}$. 
As shown in Fig.~\ref{fig:CleanTreeHard}, we have that 
$\hat{V}^{\ast,\mathrm{intree}}(h) {=} \textstyle\frac{\sum_{a \in C^{\mathrm{intree}}(h)} n^{\mathrm{intree}}(ha) \hat{Q}^{\mathrm{intree}}(ha)}{\sum_{a \in C^{\mathrm{intree}}(h)} n^{\mathrm{intree}}(ha)} \label{eq:ValueAssembly}.$
At this point we have everything to calculate the updated value function at belief $b(h)$ indexed by history $h$. To do that we use the relation:
\begin{equation}
	\begin{gathered}
		\label{eq:ValueUpdateDeletedSubtree}
		\textstyle \underbrace{\hat{V}^{\ast}(h){\cdot} n(h)}_{S(h)} {\gets} \underbrace{\hat{V}^{\ast,\mathrm{intree}}(h)n^{\mathrm{intree}}(h)}_{S^{\mathrm{intree}}(h)} {-}\\ \hat{Q}^{\mathrm{intree}} (ha) n^{\mathrm{intree}}(ha). 
	\end{gathered}
\end{equation}
For an efficient update by \eqref{eq:ValueUpdateDeletedSubtree} we need to cache the sum of the cumulative reward laces for each belief node. We define such a sum as $S(h) {\bydef}\hat{V}^{\ast}(h){\cdot} n(h)$.  In addition, we need to subtract the deleted action $a$ from the set of children of $b(h)$, namely $C(h) {\leftarrow}  C^{\mathrm{intree}}(h) {\setminus} \{a\}$. 
If $b(h)$ is a root node we just update its visitation count $n(h)$ to a new value. We do not need to store $S(h)$ for a root belief.  Else, we identify a parent action and node of $b(h)$ marked $a^{\mathrm{pa}}$ and $b^{\mathrm{pa}}$ respectively.    
We need to calculate $n(h^{\mathrm{pa}}a^{\mathrm{pa}})$ as such, $n(h^{\mathrm{pa}}a^{\mathrm{pa}}) {\leftarrow} n^{\mathrm{intree}}(h^{\mathrm{pa}}a^{\mathrm{pa}}) {-}  n^{\mathrm{intree}}(h) {+} n(h)$. We then shall update $\hat{Q}(h^{\mathrm{pa}}a^{\mathrm{pa}})$ as such 
\begin{equation}
\label{eq:QestUpdate}	
\begin{gathered}
\hat{Q}(h^{\mathrm{pa}}a^{\mathrm{pa}}){\cdot} n(h^{\mathrm{pa}}a^{\mathrm{pa}}) {\leftarrow}\hat{Q}^{\mathrm{intree}}(h^{\mathrm{pa}}a^{\mathrm{pa}}) {\cdot} n^{\mathrm{intree}}(h^{\mathrm{pa}}a^{\mathrm{pa}}) {\color{red} -} \\
\big(n^{\mathrm{intree}}(h){+}1 \big)\big(\rho(b^{\mathrm{pa}}a^{\mathrm{pa}}b){+}\hat{V}^{\ast,\mathrm{intree}}(h) {+} \mathrm{roll}^{\mathrm{intree}}(h)\big) {\color{red} +}\\
\big(n(b){+}1 \big)\big(\rho(b^{\mathrm{pa}}a^{\mathrm{pa}}b){+}\hat{V}^{\ast}(h) {+} \mathrm{roll}^{\mathrm{intree}}(h)\big).
\end{gathered}
\end{equation}
Note that \eqref{eq:QestUpdate} encompasses both cases. The case when the subtree is deleted \eqref{eq:ValueUpdateDeletedSubtree} and the case then only the visitation counts down the tree and the $\hat{Q}$ 
are updated (upper levels of the belief search tree). The value of $\hat{V}^*(h)$ subsumes both cases. In the latter case its update reads as such 
\begin{equation}
\begin{gathered}
\label{eq:ValueUpdateUpdateSubtree}
\underbrace{\hat{V}^{\ast}(h){\cdot} n(h)}_{S(h)} {\leftarrow} \underbrace{\hat{V}^{\ast,\mathrm{intree}}(h)n^{\mathrm{intree}}(h)}_{S^{\mathrm{intree}}(h)} {-} \\
\hat{Q}^{\mathrm{intree}} (ha) n^{\mathrm{intree}}(ha) + \hat{Q} (ha) n(ha).  
\end{gathered}  
\end{equation}
We now calculate a new visitation count of $b^{\mathrm{pa}}$ using
%
$n(h^{\mathrm{pa}}) {\leftarrow} n^{\mathrm{intree}}(h^{\mathrm{pa}}) - n^{\mathrm{intree}}(h^{\mathrm{pa}}a^{\mathrm{pa}}) {+} n(h^{\mathrm{pa}}a^{\mathrm{pa}}), \notag $
%
and $S(h^{\mathrm{pa}})$ using \eqref{eq:ValueUpdateUpdateSubtree} and renaming there the history from $h$ to $h^{\mathrm{pa}}$ and the action from $a$ to $a^{\mathrm{pa}}$. 
Now we can treat $b^{\mathrm{pa}}(h^{\mathrm{pa}})$ similarly as we treated $b(h)$.
We outlined the tree cleaning procedure in Alg.~\ref{alg:CleanTree}. To conclude, this way our search tree always consists of only safe actions (not dangerous with respect to Def.~\ref{def:DA}).

\section{The Algorithms and Guarantees}  
\begin{algorithm}[h!]
  \floatname{algorithm}{Listing}
\caption{Common procedures} \label{alg:common}
\begin{algorithmic}[1]
\Procedure{Plan}{belief: $b$,  horizon: $L$}
\For {$m$ iterations or timeout}
\State $h \leftarrow \emptyset$	
\State \Call{Simulate}{$b$, $b$, $h$, $L$} \textcolor{nicegreen}{\Comment{A single tree query}}
\EndFor 
\State \Return \Call{ActionSelection}{$b$, $h$, $0$}
\EndProcedure	
\end{algorithmic}
\end{algorithm}

\begin{algorithm}[p]
\caption{Probabilistically Constrained MCTS (PCMCTS)}
\begin{algorithmic}[1]
\Procedure{Simulate}{belief: $b$, belief: $\bar{b}$, history: $h$, depth: $d$}
\If { $d=0$ }
	\State \Return $0$
\EndIf
\State \textcolor{nicebrown}{ $\bar{b}^{\mathrm{safe}} \leftarrow $ \Call{MakeBeliefSafe}{$\bar{b}$}}  
\State SafeActionFlag $\gets$ false  
\While{ not(SafeActionFlag)}
\State $a \leftarrow \Call{ActionSelection}{h,c}$  
\State Calculate propagated belief $b'^{-}$ from $b$  \textcolor{nicebrown}{and $\bar{b}^{\prime,-}$ from $\bar{b}^{\mathrm{safe}}$ using $a$}
\If{$|C(ha)| \leq k_o n(ha)^{\alpha_{o}} $} \textcolor{nicegreen}{\Comment{observation Prog. Widening}}
\State $ z' \sim \probd_{\mathrm{O}}(z|x^o); x^o \sim b'^{-}$ 
\State $\color{nicebrown}{ \bar{b}^{\prime} \leftarrow \psi(\bar{b}^{\mathrm{safe}}, a, z')}$, Calculate $\mathbf{1}_{{\{\bar{b}^{\prime,-}, \bar{b}^{\prime}:\phi({\color{nicebrown}{\bar{b}^{\prime,-}}}) \geq \delta, \phi({\color{nicebrown}{\bar{b}^{\prime}}}) \geq \delta\}}}({\color{nicebrown}{\bar{b}^{\prime,-},\bar{b}^{\prime}}})$
\If{$\mathbf{1}_{\{\phi({\color{nicebrown}{\bar{b}^{\prime,-}}}) \geq \delta, \phi({\color{nicebrown}{\bar{b}^{\prime}}}) \geq \delta\}}({\color{nicebrown}{\bar{b}^{\prime,-}, \bar{b}^{\prime}}}) == 0$}
\State \Call{CleanTree}{$h, a$} \textcolor{nicegreen}{\Comment{Clean current belief tree to be safe using Alg.~\ref{alg:CleanTree}. }}    
\State {\bf Continue} \textcolor{nicegreen}{\Comment{Jump to line $14$}}    
\Else 
\State SafeActionFlag  $\leftarrow$ true
\EndIf
\State $b^{\prime} \leftarrow \psi(b, a, z')$,  $r' \leftarrow$ $\rho(b, a, z', b')$ \textcolor{nicegreen}{\Comment{Regular belief and the reward on top of it are obtained only for not pruned actions}}
\State $C(ha) \leftarrow  C(ha) \cup \{z', r', b'\}$
\State  $r^{\mathrm{lace}} \leftarrow r' + \gamma$  \Call{SafeRollout}{$b^{\prime}$, $d-1$}
\Else
\State SafeActionFlag  $\leftarrow$ true
\textcolor{blue}{\State $\{z', r', b'\} \leftarrow $ sample uniformly from  $C(ha)$}   \label{lin:PCMCTSSampleTriple}
\State  $r^{\mathrm{lace}} \leftarrow r' {+} \gamma$  \Call{Simulate}{$b^{\prime}$,$\bar{b}^{\prime}$, $haz'$, $d{-}1$}
\EndIf
\EndWhile
\State  $n(h) \leftarrow n(h) +1$ \textcolor{nicegreen}{\Comment{Initialized to zero}}
\State  $n(ha) \leftarrow n(ha) +1$ \textcolor{nicegreen}{\Comment{Initialized to zero}}
\State  $\hat{Q}(ha) \leftarrow  \hat{Q}(ha) + \frac{r^{\mathrm{lace}}-\hat{Q}(ha)}{n(ha)}$ \textcolor{nicegreen}{\Comment{Initialized to zero}}
\State $S(ha) \leftarrow S(ha) + r^{\mathrm{lace}}$ \label{lin:CacheSum} \textcolor{nicegreen}{\Comment{Initialized to zero}}
\State \Return $r^{\mathrm{lace}}$
\EndProcedure
\Procedure{ActionSelection}{$b$, $h$, $c$}
\If{$|C(h)| \leq k_a n(h)^{\alpha_{a}}$} \textcolor{nicegreen}{\Comment{action Prog. Widening}}
\State  $a \leftarrow$ \Call{NextAction}{$h$} \label{linPCMCTS:SamplingAction}
\State $C(h) \leftarrow  C(h) \cup \{a\}$
\EndIf 
\State return $\arg\max_{a \in C(h)}\hat{Q}(ha) +c \sqrt{\nicefrac{\log n(h)}{n(ha)}}$  				\textcolor{nicegreen}{\Comment{UCB}}
\EndProcedure
\end{algorithmic}
\label{alg:PCMCTS}	
\end{algorithm}		
\begin{algorithm}[h]
	\caption{Myopically Safe Rollout Action selection}
	\begin{algorithmic}[1]
		\Procedure{SafeRolloutPolicy}{$b$, $\mathcal{A}$}
		\State $\mathcal{A} \leftarrow \mathrm{shuffle}(\mathcal{A})$. $V^{\ast} \leftarrow -\infty$  \textcolor{nicegreen}{\Comment{by shuffle assure that action is selected randomly}}
		\For{ $a \in \mathcal{A}$}
			\For{$m$ iterations} 	
			\State Calculate propagated belief $b'^{-}$ from $b$ and $a$
			\State $ z' \sim \probd_{\mathrm{O}}(z|x^o); x^o \sim b'^{-}$ 
			\State $b^{\prime} \leftarrow \psi(b, a, z')$
			\State Calculate $\mathbf{1}_{\{\phi(b'^{-}) \geq \delta , \phi(b') \geq \delta \}}(b'^{-}, b') $ 
			\EndFor
			\State $\hat{V}^{(m)} \leftarrow \frac{1}{m}\sum_{i=1}^m \mathbf{1}_{\{\phi(b'^{-}) \geq \delta , \phi(b^{\prime}) \geq \delta \}}(b^{\prime,i,-}, b^{\prime,i})$
			\If{$\hat{V}^{(m)} \geq 1 - {\color{red} \epsilon}$} \textcolor{nicegreen}{\Comment{Note that we added $\epsilon$ here}}
			\State \Return $a$ \textcolor{nicegreen}{\Comment{First shuffled action satisfying myopic PC approx. is returned}}
			\ElsIf{$\hat{V}^{(m)} > \hat{V}^{(m)}_{\ast}$}
			\State $V^* \leftarrow V$, $a^* \leftarrow a$
			\EndIf
			\EndFor
			\State \Return $a^*$
			\EndProcedure
\end{algorithmic}
	\label{alg:SafeRollout}	
\end{algorithm}

This section describes our algorithms followed by convergence guarantees.  Alg.~\ref{alg:PCMCTS} summarizes our main result.  Similar to \cite{Sunberg18icaps}, we present a provable modified variant summarized by Alg.~\ref{alg:PCMCTSPUCT}.
\subsection{Detailed Algorithms Description}
\label{sec:Algorithms} 
The entry point of both these algorithms, listed in Listing \ref{alg:common},  is a  loop over trials of observation laces. The difference between the algorithms is in the \textproc{Simulate} function. We name a single call to \textproc{Simulate} a {\bf tree query}.  In each trial, we descend with the lace of observations and actions intermittently, calculate the beliefs and rewards along the way, and ascend back to the root of the belief tree.  Once, on the way down the tree, the unsafe belief is encountered we clean such action from the search tree and fix the action value estimates of all ancestor belief action nodes.  
Similar to the classical MCTS our approach can be run with rollout or without. In addition, if we do not want to use safe beliefs for the constraint we only need to remove parts marked by the {\color{nicebrown} brown} color in Alg.~\ref{alg:PCMCTS} and \ref{alg:PCMCTSPUCT} and use regular belief instead. We also present a Polynomial variant of our approach, Alg.~\ref{alg:PCMCTSPUCT} we named PC-MCTS with Polynomial Upper Confidence Tree (PC-MCTS-PUCT). In the next section, we prove the convergence with an exponential rate of Alg.~\ref{alg:PCMCTSPUCT} in probability.  Note that we cache the values of the summation of the cumulative reward for all belief nodes for both algorithms. This happens in line \ref{lin:CacheSum} of Alg.~\ref{alg:PCMCTS} and line \ref{lin:CacheSumPUCT} of Alg.~\ref{alg:PCMCTSPUCT}, where we denote  $S(h)=\hat{V}^{\ast}(h){\cdot} n(h)$. 
\begin{algorithm}[h]
	\caption{Cleaning Belief tree to be safe}
	\begin{algorithmic}[1]
		\Procedure{CleanTree}{$h$, $a$}
		\State Delete all children $b'$ of $ba$ belief-action node and delete $ba$ itself 	 $C(h) {\leftarrow}  C^{\mathrm{intree}}(h) \setminus \{a\}$
		\If{ $n^{\mathrm{intree}}(h) == 0$}
		\State \Return 
		\EndIf 
		\State $n(h) \leftarrow n^{\mathrm{intree}}(h) - n^{\mathrm{intree}}(ha)$
		\If{$b$ is root}
		\State $ C^{\mathrm{intree}}(h) \leftarrow C(h)$, $  n^{\mathrm{intree}}(h) \leftarrow n(h)$, 
		\State \Return
		\Else
		\State  Assemble $\hat{V}^{\ast}(h)$ 
		\EndIf
		\While{true} 
		\If{$b$ is root}
		\State $  n^{\mathrm{intree}}(h) \leftarrow n(h)$, 
		\State \Return
		\EndIf	
		\State Identify $a^{\mathrm{pa}}$ which is parent to $b$ 
		\State Identify $b^{\mathrm{pa}}$ such that $b^{\mathrm{pa}}a^{\mathrm{pa}}$ is a belief action node which is parent of $b$
		\State $n(h^{\mathrm{pa}}a^{\mathrm{pa}}) {\leftarrow} n^{\mathrm{intree}}(h^{\mathrm{pa}}a^{\mathrm{pa}}) {-}  n^{\mathrm{intree}}(h) {+} n(h)$  \textcolor{nicegreen}{\Comment{History $h^{\mathrm{pa}}$ corresponds to belief $b^{\mathrm{pa}}$}}
		\State Reconstruct $\hat{Q}(h^{\mathrm{pa}}a^{\mathrm{pa}})$ 
		and put $  n^{\mathrm{intree}}(h) {\leftarrow} n(h)$ and $  S^{\mathrm{intree}}(h) {\leftarrow} S(h)$
		\State  $n(h^{\mathrm{pa}}) {\leftarrow} n^{\mathrm{intree}}(h^{\mathrm{pa}}) {-} n^{\mathrm{intree}}(h^{\mathrm{pa}}a^{\mathrm{pa}}){+}n(h^{\mathrm{pa}}a^{\mathrm{pa}})$, 
		\State  Assemble $\hat{V}^{\ast}(h^{\mathrm{pa}})$ 
		and put $\hat{Q}^{\mathrm{intree}}(h^{\mathrm{pa}}a^{\mathrm{pa}}) {\leftarrow} \hat{Q}(h^{\mathrm{pa}}a^{\mathrm{pa}})$ \textcolor{nicegreen}{\Comment{We have $V^*(h^{\mathrm{pa}})$ and $n(h^{\mathrm{pa}}$) for the next iteration}}
		\State  $b {\leftarrow} b^{\mathrm{pa}}$ 
		\EndWhile
		\EndProcedure
	\end{algorithmic}
	\label{alg:CleanTree}	
\end{algorithm}	
\paragraph{Safe Rollout}
The rollout is not necessary for applying MCTS. Without rollout, upon opening a new belief node the MCTS would behave as it reached the leaf node. Nevertheless, the rollout helps to provide better results in finite time and apparently helps to accelerate convergence (We did not find any rigorous analysis for that). With this motivation in mind, we present the Safe Rollout routine for our approach summarized by Alg.~\ref{alg:SafeRollout}. Our safe rollout selects action randomly which myopically fulfills the sample approximation of myopic PC based on $m$ samples.  If no feasible action exists (which is not possible with our method since we always have an action ``do not do anything'') we select an action maximizing the sample approximation mentioned before.   

\subsection{Convergence Guarantees}
\label{sec:ConvergenceGuarantees}
Although we sample actions and observations and the number of samples of both is marching to infinity in discrete steps, MCTS converges, at each belief action node in probability, to the optimal value of the probailistically constrained problem defined by 
\begin{align}
	&Q^{\pi^*}(b(h), a;\boldsymbol{\rho}) \text{ \ \ subject to} \label{eq:ConstrObjEachHistory} \\
	&\prob\big(c{=}1 |  \bar{b}(h), a, \pi \big){=}1.   \label{eq:OuterConstrEachHistory} 
\end{align}
In \eqref{eq:ConstrObjEachHistory} we omitted the time indices.  MCTS approximates the stochastic policy $\pi^*$ by a discrete but infinite set of sampled continuous actions and statistics defined by the visitation counts.  
To give an intuition, the convergence in probability is a result of the fact that we are dealing with expectations (See Lemma~\ref{thm:RepresentValueStochPolicy} and Theorem~\ref{thm:Represent}) and the fact that every belief action node is visited infinite amount of times due to \eqref{eq:Expl} even in continuous domains where the new nodes are endlessly expanded. Moreover, the polynomial variant of DPW used in Alg.~\ref{alg:PCMCTSPUCT} allows to each belief-action node be visited a sufficient amount of times before the next new belief is introduced. Thus,  convergence in probability happens with an exponential rate.   
Although the tree policy is tree query dependent and improves over time, \cite{Auger13Sp} showed the convergence is from the leafs and upwards to the value under the optimal stochastic policy. 
Further we show that if the actions are continuous and have some natural distribution, MCTS will eventually sample an unsafe action (line \ref{linPCMCTS:SamplingAction} in Alg.~\ref{alg:PCMCTS} and line \ref{linPCMCTSPUCT:SamplingAction} in Alg.~\ref{alg:PCMCTSPUCT}). 
\begin{thm}
	Suppose, at the node $h$, we have an action $a$ sampler on top of the continuous probability space $(\mathcal{A}, \mathcal{F}, \prob)$ where $\mathcal{A}$ is the outcomes space, $\mathcal{F}$ events space and $\prob$ is the probability. At the limit of convergence in infinite time (after an infinite number of tree queries), it holds that $C(h)$ includes an action sampled from $\mathcal{A}$ that is arbitrary close to the optimal action with respect to the theoretical action-value function at each belief node. 
\end{thm}
 To prove that we take an arbitrary set $S {\in} \mathcal{A}$ such that $\mathrm{P}(S|h) {>} 0$. Note that the probability here depends on the history (belief) we focus on.   
Time marches to infinity in countable steps so as the samples are countable. Denote by $|C(h)|$ the number of $i.i.d.$ samples (line \ref{linPCMCTS:SamplingAction} in Alg.~\ref{alg:PCMCTS} and line \ref{linPCMCTSPUCT:SamplingAction} in Alg.~\ref{alg:PCMCTSPUCT} ). The probability of sampled action $a {\sim} \probd(a|h)$ not to be in $S$ is $\prob(\{a {\in} \mathcal{A}: a {\not\in} S\}|h){=}1 {-} \prob(S|h)$. When the number of samples tends to infinity we have that 
\begin{align}
	(1 {-} \prob(S|h))^{|C(h)|} \underbrace{\to}_{|C(h)| \to \infty} 0.
\end{align}
It holds that the probability not to sample an action in $S$ tends to zero with number of samples tending to infinity. 
Therefore, in the unconstrained MCTS approach, the action value function estimates $\hat{Q}$ will include unsafe actions if they are exist in the action space $\mathcal{A}$.  
On the contrary in our safe approach we remove the actions sampled from the unsafe sets of arbitrary small positive measure. 

We now show that the cleaning tree routine is necessary due to fact that we will sample an infinite amount of unsafe actions and this can shift the expectations with respect to actions in values maintained in search tree.

Without our pruning we will obtain infinitely many unsafe actions in each unsafe set $S$. It can be seen using the second Borel Cantelli Lemma. Towards this end let us define the event $E^i{\bydef}\{a^i {\sim} \mathbb{P}(a|h) {:} a^i {\in} S\}$, sampled action $i$ is a member of set $S$. The events $E^i$ are independent since we sample actions independently. The series $\sum_{i=1}^{\infty}\prob(E^i|h)$ are divergent since  $\prob(E^i|h){=} \prob(S|h) {>}0$ by definition. Thus, $\prob \big(\bigcap_{j=1}^{\infty} \bigcup_{i=j}^{\infty}E^i\big|h\big){=}1$, namely the event  sampled action is a member of set $S$ occurs infinitely often times. To rephrase that, without our pruning we would have sampled infinitely many dangerous actions and, therefore, the expectations can undergo a shift and even if at the root of the belief tree we select an optimal safe action, the influence of unsafe future actions can be substantial. 


In this section, we prove convergence in probability with an exponential rate of Polynomial Upper Confidence Tree (PUCT) version of our approach (Alg.~\ref{alg:PCMCTSPUCT}). 
We now list down the changes between Alg.~\ref{alg:PCMCTS} and Alg.~\ref{alg:PCMCTSPUCT}:
\begin{itemize}
	\item In Alg.~\ref{alg:PCMCTSPUCT} we have Polynomial Double Progressive Widening with depth dependent parameters defined in \cite{Auger13Sp};
	\item The rollout in Alg.~\ref{alg:PCMCTSPUCT} is missing;
	\item If Alg.~\ref{alg:PCMCTS} decided not to open a new branch it samples the triple $\{z', r', b'\}$ uniformly from $C(ha)$ (line~\ref{lin:PCMCTSSampleTriple} highlighted by the blue color). In contrast Alg.~\ref{alg:PCMCTSPUCT} selects the child with a minimal visitation count (line~\ref{lin:PCMCTSPUCTargminTriple} highlighted by the blue color).
\end{itemize}

 The following theorem provides its soundness. 
\begin{thm}[Convergence with Exponential Rate in Probability]
	\label{thm:Convergence}
	Every belief $h$ and belief action node $ha$ of  Alg.~\ref{alg:PCMCTSPUCT}, equipped with our pruning mechanism from Section \ref{sec:pruning} and summarized by Alg.~\ref{alg:CleanTree} converges in probability and with an exponential convergence rate to the optimal value function $V^*(b(h))$ and action-value function $Q(b(h)a)$, respectively, while satisfying the PC starting from the belief action node $ha$, namely $ \prob\big(c{=}1 |  \bar{b}(h), a, \pi^{\ast}  \big){=}1$. 
\end{thm} 

Next, we provide the proof under rather mild assumptions. To be specific we must assume that reward lies in a bounded interval and that sampling of actions covers the entire space with an arbitrary precision. For more precise definition see Def.~\ref{def:Regularity}. Our proof is valid for both approaches, namely with making belief safe before pushing forward in time with action and observation and without (in this case the constraint at each belief-action node is $\prob\big(c{=}1 |  b(h), a, \pi^{\ast} \big){=}1$).  
Similar to \cite{Sunberg18icaps} we leverage the proof by \cite{Auger13Sp}. 
\begin{algorithm}[h!]
\caption{Probabilistically Constrained MCTS PUCT}
\begin{algorithmic}[1]	
\Procedure{Simulate}{belief: $b$, belief: $\bar{b}$, history: $h$, depth: $d$}
\If { $d=0$ }
	\State \Return $0$
\EndIf
\State $\bar{b}^{\mathrm{safe}} \leftarrow $ \Call{MakeBeliefSafe}{$\bar{b}$}  
\State SafeActionFlag $\leftarrow$ false  
\While{ not(SafeActionFlag)}
\State $a \leftarrow \Call{ActionSelection}{h,c}$  
\State Calculate propagated belief $b'^{-}$ from $b$ and $\bar{b}^{\prime,-}$ from $\bar{b}^{\mathrm{safe}}$ using $a$
\If{$ \lfloor n(ha)^{\alpha_{o,d}} \rfloor  >  \lfloor (n(ha)-1)^{\alpha_{o,d}} \rfloor$} \textcolor{nicegreen}{\Comment{observation Prog. Widening}}
\State $ z' \sim \probd_{\mathrm{O}}(z|x^o); x^o \sim b'^{-}$ 
\State $\bar{b}^{\prime} {\leftarrow} \psi(\bar{b}^{\mathrm{safe}}, a, z')$, Calculate $\mathbf{1}_{\{\bar{b}^{\prime,-}, \bar{b}^{\prime}:\phi(\bar{b}^{\prime,-}) \geq \delta, \phi(\bar{b}^{\prime}) \geq \delta \}}(\bar{b}^{\prime,-},\bar{b}^{\prime}) $ 
\If{$\mathbf{1}_{\{\phi(\bar{b}^{\prime,-}) \geq \delta, \phi(\bar{b}^{\prime}) \geq \delta \}}(\bar{b}^{\prime,-},\bar{b}^{\prime}) == 0$}
\State \Call{CleanTree}{$h, a$} \textcolor{nicegreen}{\Comment{Clean current belief tree to be safe using Alg.~\ref{alg:CleanTree}.}}    
\State {\bf Continue} \textcolor{nicegreen}{\Comment{Jump to line $14$}}    
\Else 
\State SafeActionFlag  $\leftarrow$ true
\EndIf
\State $b^{\prime} \leftarrow \psi(b, a, z')$, $r' \leftarrow$ $\rho(b, a, z', b')$ \textcolor{nicegreen}{\Comment{Regular belief and reward on top of it are obtained only for not pruned actions}}
\State $C(ha) \leftarrow  C(ha) \cup \{z', r', b'\}$
\Else
\State SafeActionFlag  $\leftarrow$ true
\textcolor{blue}{\State $\{z', r', b'\} \leftarrow $ $\underset{\{z', r', b'\} \in C(ha)}{\arg\min} \frac{n(hao)}{n(h)} $}    \label{lin:PCMCTSPUCTargminTriple}
\EndIf
\EndWhile
\State  $r^{\mathrm{lace}} \leftarrow r' + \gamma$  \Call{Simulate}{$b^{\prime}$, $\bar{b}^{\prime}$, $haz'$,$d{-}1$}
\State  $n(h) \leftarrow n(h) +1$ \textcolor{nicegreen}{\Comment{Initialized to zero}}
\State  $n(ha) \leftarrow n(ha) +1$ \textcolor{nicegreen}{\Comment{Initialized to zero}}
\State  $\hat{Q}(ha) \leftarrow  \hat{Q}(ha) + \frac{r^{\mathrm{lace}}-\hat{Q}(ha)}{n(ha)}$ \textcolor{nicegreen}{\Comment{Initialized to zero}}
\State $S(ha) \leftarrow S(ha) + r^{\mathrm{lace}}$ \label{lin:CacheSumPUCT} \textcolor{nicegreen}{\Comment{Initialized to zero}}
\State \Return $r^{\mathrm{lace}}$
\EndProcedure
\Procedure{ActionSelection}{$b$, $h$, $c$}
\If{$\lfloor n(h)^{\alpha_{a,d}} \rfloor  >  \lfloor (n(h)-1)^{\alpha_{a,d}}\rfloor$} \textcolor{nicegreen}{\Comment{action Prog. Widening}}
\State  $a \leftarrow$ \Call{NextAction}{$h$} \label{linPCMCTSPUCT:SamplingAction}
\State $C(h) \leftarrow  C(h) \cup \{a\}$
\EndIf 
\State return $\arg\max_{a \in C(h)}\hat{Q}(ha){+} \sqrt{\nicefrac{ n(h)^{e_d}}{n(ha)}}$  				\textcolor{nicegreen}{\Comment{PUCT}}
\EndProcedure
\end{algorithmic}
\label{alg:PCMCTSPUCT}	
\end{algorithm}

Before we proceed let us mention that DPW of Alg.~\ref{alg:PCMCTS} with $k_z = k_a=1$ and depth dependent  $\alpha_d$ as described in \cite{Auger13Sp}  are identical to the one used in Alg.~\ref{alg:PCMCTSPUCT}. 
\begin{lem}
	Fix belief node $b(h)$ in belief tree and belief action node $ha$, $k_a = k_o=1$ in Alg.~\ref{alg:PCMCTS} and select in Alg.~\ref{alg:PCMCTS} and Alg.~\ref{alg:PCMCTSPUCT} same $\alpha_{o,d}$ and $\alpha_{a,d}$ $\in (0,1)$  in both algorithms (can be depth dependent). The condition $|C(ha)| \leq n(ha)^{\alpha_o,d}$ is equivalent to $ \lfloor n(ha)^{\alpha_{o,d}} \rfloor  >  \lfloor (n(ha)-1)^{\alpha_{o,d}} \rfloor$. In a similar manner   $|C(h)| \leq n(h)^{\alpha_a,d}$ is equivalent to $ \lfloor n(h)^{\alpha_{a,d}} \rfloor  >  \lfloor (n(h)-1)^{\alpha_{a,d}} \rfloor$. 
\end{lem}
We provide sketch of the proof. It is sufficient to prove that the first claim is identical since both have identical structure. Let us focus on $|C(ha)| \leq n(ha)^{\alpha_o,d}$. The new child is added if and only if  the visitation $n(ha)^{\alpha_o,d}$ passes the subsequent integer at some visitation of node $ha$. This is happening if and only if   $ \lfloor n(ha)^{\alpha_{o,d}} \rfloor  >  \lfloor (n(ha)-1)^{\alpha_{o,d}} \rfloor$.  
\qed

Now we would like to pay attention to the fact that since we clean the belief tree from the unsafe actions, we have $n(h) {\geq} \sum_{a \in C(h)} n(h,a)$. This is in contrast to the classical MCTS, where $n(h) {=} \sum_{a \in C(h)} n(h,a)$. Therefore, we shall fix the visitation count of each belief node and belief action node that have been affected by pruning. This is done by Alg.~\ref{alg:CleanTree}.

The following claim is required to understand  \cite{Auger13Sp} and we give now an informal proof missing in \cite{Auger13Sp}.  
\begin{lem}
	The $k^{\mathrm{th}}$ child of node $ha$ in Alg.~\ref{alg:PCMCTSPUCT} is added on visit $n(ha) =  \lceil k^{\frac{1}{\alpha}} \rceil  \bydef n_k(ha) $.
\end{lem}

Observe that the left hand side of $ \lfloor n(ha)^{\alpha_{o,d}} \rfloor  >  \lfloor (n(ha)-1)^{\alpha_{o,d}} \rfloor$ jumped $\lfloor n(ha)^{\alpha_{o,d}} \rfloor$  times and right hand side lagged exactly by single visitation. So the inequality is fulfilled exactly  $\lfloor n(ha)^{\alpha_{o,d}} \rfloor$ times. Moreover, the \emph{first} $n(ha)$ such that $n(ha)^{\alpha_{o,d}}$  passes a subsequent integer assures the jump. Meaning, we have two cases. The first case is $n(ha)^{\alpha_{o,d}} {=}\lfloor n(ha)^{\alpha_{o,d}} \rfloor {=}k$ and taking $k^{\frac{1}{\alpha_o,d}} = \lceil k^{\frac{1}{\alpha_o,d}} \rceil$  is returning us back to  $n(ha)$. The second case is $\lfloor n(ha)^{\alpha_{o,d}} \rfloor +1 > n(ha)^{\alpha_{o,d}} > \lfloor n(ha)^{\alpha_{o,d}} \rfloor   =k$. But we know that if $k^{\frac{1}{\alpha}}$ would be an integer, it would the previous case with a smaller $n(ha)$. The  $k^{\frac{1}{\alpha}}$ has to be slightly larger than integer so as ceil operator return the right natural $n(ha)$.
Similarly number of actions expanded from node $h$ is $\lfloor n(h)^{\alpha} \rfloor$.  
\qed

Our cleaning routine prunes only the actions and fixes the affected visitation counts so the proof by \cite{Auger13Sp} is not broken. Further we establish the definitions and the assumptions from \cite{Auger13Sp} in order to assure the validity of the proof. Some of them we take directly from \cite{Auger13Sp}, \cite{Sunberg18icaps}.  
\begin{defn}[Regularity Hypothesis] \label{def:Regularity} The Regularity
	hypothesis is the assumption that for any $\Delta {>} 0$, there
	is a non zero probability to sample an action that is optimal with precision $\Delta$. More precisely, there is a $\theta > 0$
	and a $p > 1$ (which remain the same during the whole simulation) such that for all $\Delta > 0$, 
	\begin{align}
	Q(ha) {\geq}  V^*(h) {-} \Delta
	\text{ with probability of at least } \min(1, \theta \Delta^P).
	\end{align}
\end{defn}

\begin{defn}[Exponentially sure in $n$] We say that
	some property depending on an integer $n$ is exponentially sure in $n$ if there exist positive constants $C, h$,
	and $\eta$ such that the probability that the property holds
	is at least 
	\begin{align}
	1 - C \exp(- hn^ \eta  ).
	\end{align}	
\end{defn}	
In addition we need to assume that the belief dependent reward is bounded from below and above, namely it lies in the closed interval $[\rho^{\mathrm{min}}, \rho^{\mathrm{max}}]$. Instead of $\rho: \mathcal{B}\times \mathcal{A} \times \mathcal{Z} \times \mathcal{B} \mapsto \mathbb{R}$ we require the mapping to be $\rho: \mathcal{B}\times \mathcal{A} \times \mathcal{Z} \times \mathcal{B} \mapsto [\rho^{\mathrm{min}}, \rho^{\mathrm{max}}]$  Under these assumptions the convergence result of Alg.~\ref{alg:PCMCTSPUCT} summarized by Theorem~\ref{thm:Convergence} holds.

\section{SOTA Continuous Constrained MCTS}
\label{sec:Baseline}
We now firm up the loose ends and turn to the description of the existing constrained POMDP considered in an anytime setting which will serve as our baseline. 
\subsection{Expectation Constrained Belief-dependent POMDPs} 
\label{sec:ExConstr}
The averaged constraint formulated with {\bf payoff} operator and including the propagated beliefs would be
\begin{align}
\mathbb{E}^{\mathrm{T},\mathrm{O}}\big[\textstyle \sum_{\ell=0}^{L}\phi(b^-_{\ell}, b_{\ell})\big| b_0, \pi \big] \geq \delta.  \label{eq:AveragedConstraintRoot} 
\end{align}
One possibility is to define $\phi(b^-_{\ell}, b_{\ell})  {=} \phi(b^-_{\ell}) {+}\phi(b_{\ell}) $. 
Clearly the cumulative averaged formulation   \eqref{eq:AveragedConstraintRoot}  is not suitable for safety since it permits deviations of the individual safety operators $\phi$. It can happen that with the low probability of future observation the resulting posterior belief will be {\bf extremely} unsafe. However, sometimes the operator $\phi$ is naturally bounded from above. 
It holds that $\prob(\{x_{\ell} {\in} \mathcal{X}^{\mathrm{safe}}_{\ell}\}|\square) {\leq} 1$. Thus, if we select an operator $\phi$ as in \eqref{eq:ProbSafeGivenBelief} and $\delta{=}2L$ it is sufficient to ensure safety. If $\delta {<} 2L$, it permits deviations of the individual belief dependent operators. Therefore, the averaged with respect to observations episodes and stochastic policy cumulative constraint is not sufficient to assure safety.   
The works \cite{Lee18nips}, \cite{Jamgochian23arxiv} impose the averaged cumulative constraint at the root of the belief tree as 
\begin{align}
V^{\pi}(b_0;\boldsymbol{\theta}_0) {\bydef} \mathbb{E}^{\mathrm{T},\mathrm{O}}\big[\textstyle\sum_{\ell=0}^{L}\theta({\color{turquoise}b^-_{\ell}}, b_{\ell})\big| b_0, \pi \big] \leq \delta^{\theta}. \label{eq:AveragedConstraintRootCost}
\end{align}
We introduced the optional dependence on $b^-$ of cost operator $\theta$ (emphasized by {\color{turquoise}{turquoise}} color), 
e.g 
\begin{align}
	\theta({\color{turquoise} b^-_{\ell}}, b_{\ell}) {=} {\color{turquoise}{\theta(b^-_{\ell})}}{+}\theta(b_{\ell})  \label{eq:CostProp}
\end{align}	 
where
\begin{equation}
	\label{eq:ProbNotSafeGivenBelief}
	\begin{gathered}
		\theta(\square) {=} 1{-} \overbrace{\prob\big(\{x_{\ell} \in \mathcal{X}^{\mathrm{safe}}_{\ell}\}\big|\square\big)}^{\phi(\square) \ \text{from \eqref{eq:ProbSafeGivenBelief}}}{=}\\
		\prob\big(\{x_{\ell} \notin \mathcal{X}^{\mathrm{safe}}_{\ell}\}\big|\square\big) {=} \mathbb{E}_{x_{\ell} \sim \square}[\mathbf{1}_{\{x_{\ell} \notin \mathcal{X}^{\mathrm{safe}}_{\ell}\}}]
	\end{gathered}
\end{equation}
with values in $\square$ are substituted by $b^-_{\ell}$ and $b_{\ell}$ respectively. 
Similar to the behavior of bounded payoff operator, here we can assure safety if $\delta^{\theta} {=} 0$. This will assure that \eqref{eq:AveragedConstraintRootCost} is satisfied if and only if {\bf all} $\theta(b^-_{\ell}, b_{\ell})$ inside are zero. This is because $\prob\big(\{x_{\ell} {\notin} \mathcal{X}^{\mathrm{safe}}_{\ell}\}\big|\square\big){\geq} 0$. 
In the light of the discussion about deviation of the cost values, further in this paper we assume that $\delta^{\theta}{=}0$. Now, if we set $\delta{=}1{-}\delta^{\theta}{=}1$ in our PC \eqref{eq:OuterConstr} and payoff as in \eqref{eq:ProbSafeGivenBelief} two formulations are equivalent. Yet, this will happen solely with payoff operator being as in \eqref{eq:ProbSafeGivenBelief}, cost as in \eqref{eq:ProbNotSafeGivenBelief} and $\delta{=}1$.  
Another possibility is to define cost in \eqref{eq:AveragedConstraintRootCost}  as 
\begin{align}
\theta_{\ell}(b^-_{\ell}, b_{\ell}) {\bydef} 1 {-} \mathbf{1}_{A^{\delta}_{\ell}}(b^-_{\ell}, b_{\ell}) \label{eq:CostasOur}
\end{align}
with $\delta^{\theta} {=} 0$ and $$		A^{\delta}_{\ell} {\bydef} 
\left\{b^-_{\ell}, b_{\ell}{:} b^-_{\ell} {\in} \mathcal{B}^-_{\ell}, b_{\ell} {\in} \mathcal{B}_{\ell}, \phi(b^-_{\ell}) {\geq} \delta, \phi(b_{\ell}) {\geq} \delta \right\}.$$ We obtain that the \eqref{eq:AveragedConstraintRootCost} is satisfied if and only if our PC \eqref{eq:OuterConstr} is satisfied and in both formulations we have the freedom to select operator $\phi$ and $\delta$ (we still need to assure that the $\delta$ is the same in both formulations). 
Importantly, unlike the cost from \eqref{eq:ProbNotSafeGivenBelief}, the  cost from \eqref{eq:CostasOur} can not be represented as expectation over the state dependent cost.  This cost is general belief dependent operator even if the payoff inside is as \eqref{eq:ProbSafeGivenBelief}.  
{\bf Remark}: In general the transition between cost constraint and payoff constraint is not trivial. To do that one must use the linearity of the expectation and the relation between the cost and payoff operators. 

\subsection{Duality Based Approach}
\label{sec:DualityBasedApproach}
We now turn to the discussion about duality based approach in continuous spaces suggested in \cite{Jamgochian23ICAPS}. Suppose that $\delta^{\theta}{=}0$ in \eqref{eq:AveragedConstraintRootCost};  The iterative scheme of duality based approach subsumes two steps iteratively solving the following objective  
\begin{align}
\max_{\pi} \min_{\lambda \geq 0} \big( V^{\pi}(b_0;\boldsymbol{\rho}_1){-} \lambda \underbrace{\overbrace{V^{\pi}(b_0;\boldsymbol{\theta}_0)}^{\text{as in \eqref{eq:AveragedConstraintRootCost}}} }_{\geq 0}\big), \label{eq:DualObj}
\end{align}
where one step minimizes for $\lambda$ and another maximizes for execution stochastic policy $\pi$. Here, $\boldsymbol{\theta}_0$ is a vector of  cost operators (starting from time $0$).  
The Dual ascend goes towards $V^{\pi}(b_0;\boldsymbol{\theta}_0) {=} 0$. The policy is feasible only in this case. In the $\lambda$ minimization step, since $V^{\pi}(b_0;\boldsymbol{\theta}_0) {\geq} 0$, the larger $\lambda$ will yield smaller objective. Thus, this part of the objective is becoming increasinlgy important with the iterations of the step of the minimization of $\lambda$.  
In practice, the \eqref{eq:DualObj} is approximated by the MCTS estimator. 
Many different suboptimal actions participate within every $\hat{Q}$ in the search tree. This is a direct result of the exploration exploitation tradeoff portrayed by the \eqref{eq:Expl} and the assembling each  $\hat{Q}$ from the laces (e.g. if the search tree rooted at $b_0$, the corresponding action value is  $\hat{Q}(b_0a_0) {=} \nicefrac{1}{n(b_0a_0)}\sum_j q(b^j_{0:L})$).
Another possibility would be on the way up the tree to take the maximum of the previously calculated $\hat{Q}(b(h), a)$ with respect to actions with visitation count $n(ha){>}0$. We need to exclude the actions with $n(ha){=}0$ in order not to take the initial values $q^{\mathrm{init}}$ (Fig.~\ref{fig:Exploration}). If we do that, on the way up the tree, instead of completing the lace with future cumulative reward we will complete it with the result of the maximum.  Still, the problem of many actions participating in the $\hat{Q}$ remains. 
Because the result of the maximum is changing as MCTS progresses, still suboptimal actions are participating in each $\hat{Q}$ besides the leaves. This aspect is detrimental to online planning under the safety constraint.  If the safety is formulated as in eq. \eqref{eq:AveragedConstraintRootCost} and \eqref{eq:ProbNotSafeGivenBelief} with $\delta^{\theta}{=}0$ and the objective is \eqref{eq:DualObj}, the robot will prefer to depart from unsafe regions as far as possible to ensure that the all expanded actions with at least single posterior are safe at as many beliefs as possible in the search tree.   
The importance of the all actions being safe increases closer to the root  because closer to the root actions participate within more laces. 
Our approach does not suffer from such a problem since we prune the dangerous actions in the first place.


\section{Simulations and Results}
\label{sec:SimResults}
\begin{figure}[t]
	\centering 
	\includegraphics[width=\columnwidth]{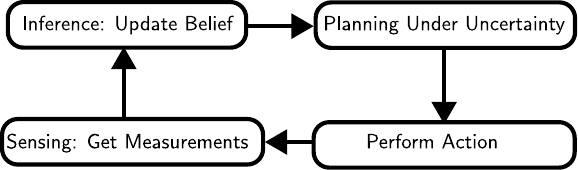}
	\caption{Autonomy loop.}
	\label{fig:AutonomyLoop}
\end{figure}
\begin{figure*}[t]
	\centering
	\begin{minipage}[t]{0.35\textwidth}
		\centering 
		\includegraphics[width=\textwidth]{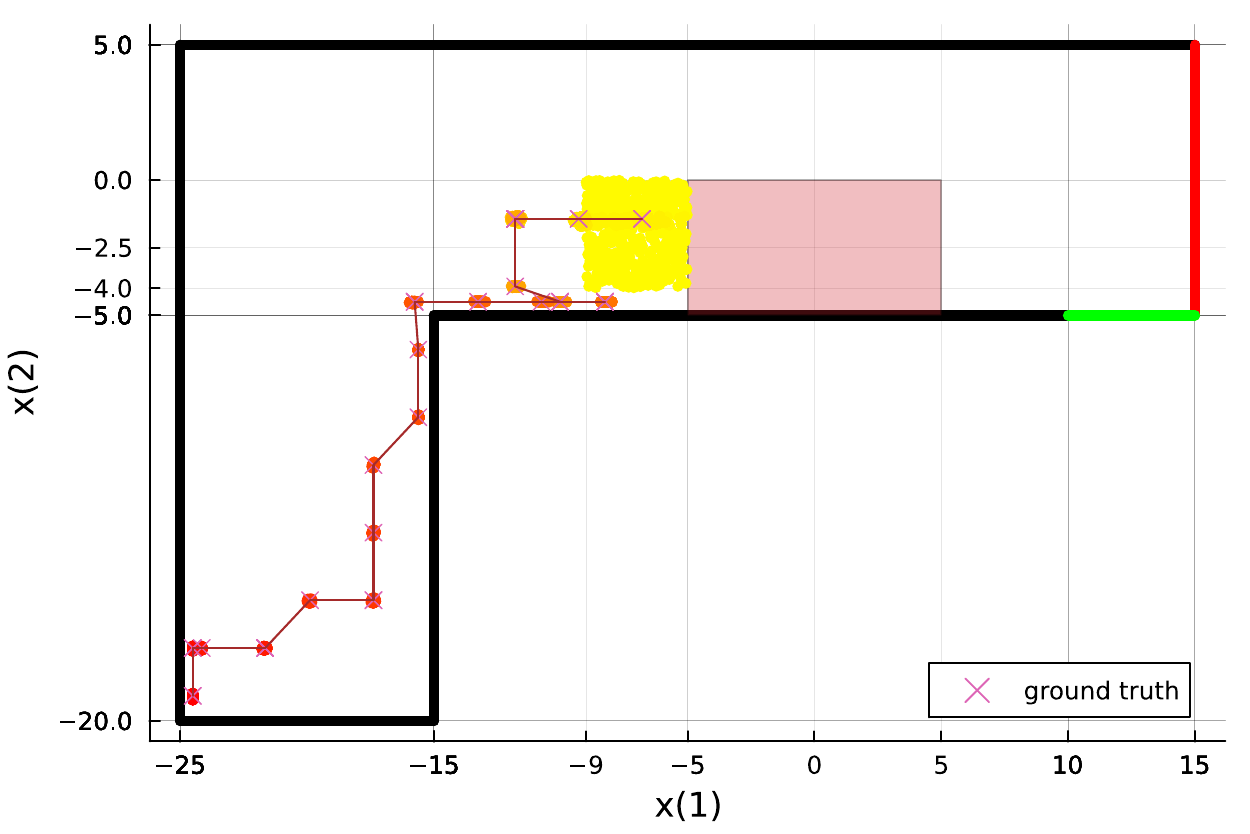}
		\subcaption{}
		\label{fig:RoombaCPFT}
	\end{minipage}%
	\hfill
	\begin{minipage}[t]{0.35\textwidth}
		\centering 
		\includegraphics[width=\textwidth]{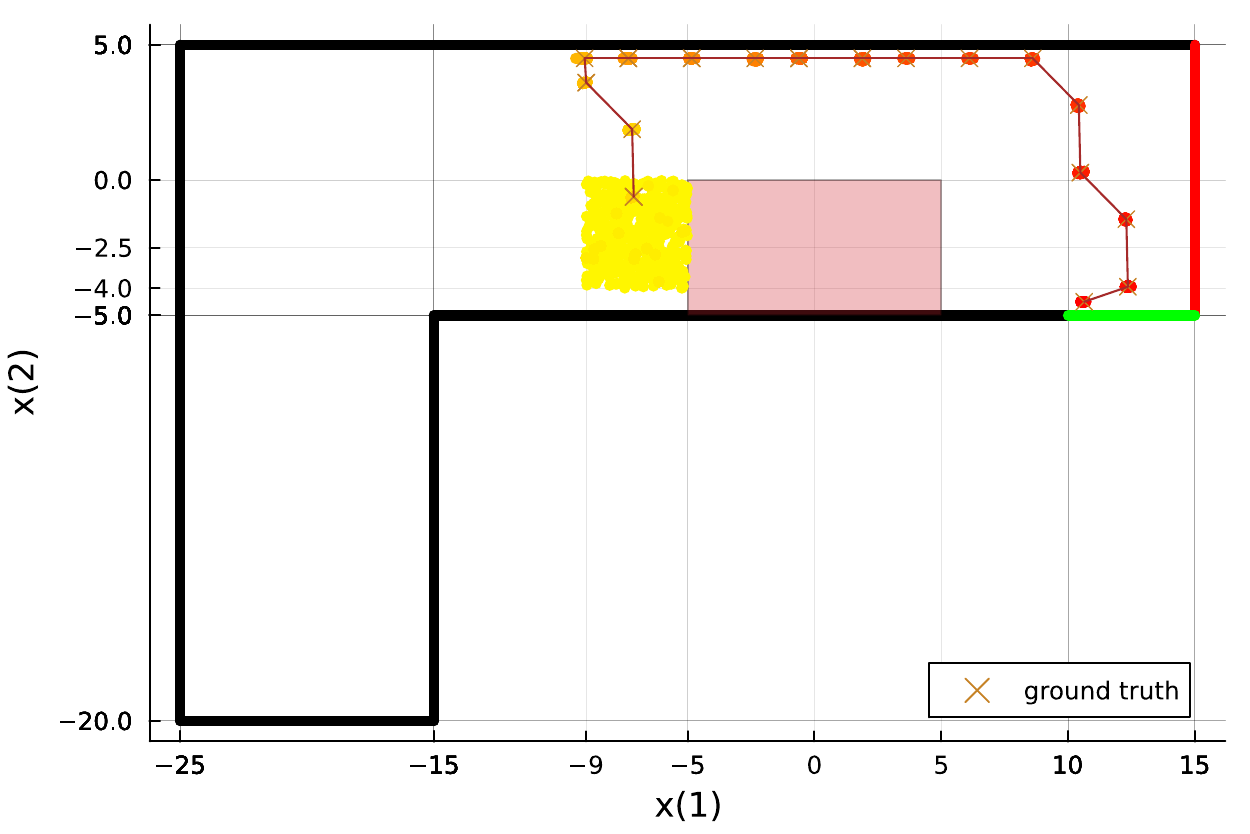}
		\subcaption{}
		\label{fig:RoombaPCPFT}
	\end{minipage} 
	\caption{Plot of one of the trials of the execution of the actions from planning in Lidar Roomba problem. Illustration of departing from the unsafe region problem in Lagrangian based methods. The yellow rectangle represents $500$ particles of $b_0$ sampled from a uniform distribution, the pink rectangle is the unsafe region to avoid. The green line is the exit area and the red line is the stairs. On both figures we plotted the ground truth robot positions and the beliefs that transit from yellow to red as time indexes progress; \textbf{(a)}  CPFT departs as far as possible from the unsafe region.  \textbf{(b)}  Our method behaves as expected: the agent goes to the green area while avoiding the unsafe region.}
	\label{fig:Roomba}	 
\end{figure*}
We are now eager to demonstrate our findings in simulations. 
We compare our approach (Alg.~\ref{alg:PCMCTS} with CPFT-DPW suggested in \cite{Jamgochian23ICAPS} with our modifications in terms of constraining propagated beliefs as described in Section~\ref{sec:ExConstr}.  For our approach named PC-PFT-DPW we select the payoff operator $\phi$ as in \eqref{eq:ProbSafeGivenBelief} and simulate for $\delta {=} 1$. Our baseline follows the averaged constraint formulation in the cost form as in \eqref{eq:AveragedConstraintRootCost} with cost operator as in \eqref{eq:CostasOur}, payoff operator $\phi$ and $\delta$ inside \eqref{eq:CostasOur} identical to one used in PC-PFT-DPW. As described in Section~\ref{sec:ExConstr}    
we set $\delta^{\theta}{=}0$. In PC-PFT-DPW we use our safe rollout (Alg.~\ref{alg:SafeRollout}) whereas in CPFT-DPW the rollout is set per problem.  

We always simulate the trials of a number of autonomy loop cycles.  A single cycle of autonomy loop is depicted at Fig.~\ref{fig:AutonomyLoop}. We now specify our problems under consideration.
\subsection{Problems Composition} 
We present the Safe Lidar Roomba problem. 
\paragraph{Safe Lidar Roomba} 
Roomba is a robotic vacuum cleaner that attempts to localize itself in a familiar room and reach the target region. The POMDP state is the position of the agent $x$, its orientation angle $\theta$, and the status. The status is a binary variable and it tells of whether the robot has reached goal state or stairs. The Roomba action space is defined as  
%
	$\mathcal{A}{=}\{a^1,a^2, a^3, a^4, a^5, a^6 \}. $
%
The action space $\mathcal{A}$ comprises the pairs $(v, \omega^{v})$. Each Roomba action is a pair $(v, \omega^{v})$. It comprises a velocity $v$ and a corresponding angular velocity $\omega^{v} {=}\nicefrac{\mathrm{d}\theta}{\mathrm{d}t}$. 
We discretized the velocities and the angular velocities and selected the following action space $a^1{=}(0, -\pi/2)$, $a^2{=}(0,0)$, $a^3{=}(0,\pi/2)$, $a^4{=}(5,-\pi/2)$, $a^5{=}(5,0)$, $a^6{=}(5, \pi/2)$.  We also have $v^{\mathrm{noise\_coeff}} {=} 0.2$ and  $\omega^{\mathrm{noise\_coeff}} {=} 0.05$ such that $v^{\mathrm{max}} {=} 5 {+} 0.5{\cdot} v^{\mathrm{noise\_coeff}}$ and $\omega^{\mathrm{max}} {=} \pi/2 +  0.5{\cdot} \omega^{\mathrm{noise\_coeff}}$. In our simulations we selected  $\mathrm{d}t {=} 0.5$ sec. We set $\sigma_{\mathrm{ray}} {=} 0.01$,  $\mathrm{rl}_{\mathrm{min}}{=} 0.001$, the stairs penalty is $-10000$, the goal reward is $10000$, 
the time penalty  is $-1000$.

The robot motion is deterministic with a predefined time step  $\mathrm{d}t$, but the action is noisy.  
When we apply PF each particle is propagated with a noisy action. The velocity noise is drawn from a uniform distribution over the interval $ (-0.5v^{\mathrm{noise}}, 0.5v^{\mathrm{noise}})$. In a similar manner, the angular velocity noise is uniform over the interval $ (-0.5\omega^{\mathrm{noise}}, 0.5\omega^{\mathrm{noise}})$. We draw the noise for each particle and add to the action $a {\in} \mathcal{A}$ before we apply the motion model.  To do so,
we first clamp velocity $v$ in the interval $[0, v^{\mathrm{max}}]$. We then clamp $\omega^{v}$ in the interval $[-\omega^{\mathrm{max}}, \omega^{\mathrm{max}}]$. The next $\theta^{\prime} {=} \theta {+} \omega^{v} {\cdot} \mathrm{d}t$ is wrapped to the interval $(-\pi,\pi]$. After the turn, next position of the agent is $x' {=} x {+} v{\cdot} \mathrm{d}t{\cdot} (\cos(\theta), \sin(\theta))^T$. If the robot hits the wall, it stops. The status  becomes $1$ if the robot hits the goal wall (green color in Fig.~\ref{fig:RoombaCPFT}) and $-1$ if the robot hits a stairs wall (red color in Fig.~\ref{fig:RoombaCPFT}).     
At the end of the motion step, the status is updated and the agent takes an observation. 
It first determines the ray length $\mathrm{rl}$ using the known workspace (room) and the position and heading direction $(\cos(\theta), \sin(\theta))^T$ of the robot.  The distribution of the observation conditioned on the robot pose is then Gaussian $\mathcal{N}(\mathrm{rl}, \sigma({\mathrm{rl}}))$ truncated from the left at zero, where $\sigma({\mathrm{rl}}) {=}  \sigma_{\mathrm{ray}}  \max(\mathrm{rl}, \mathrm{rl}_{\mathrm{min}})$. 
To introduce the safety aspect, similar to \cite{Jamgochian23arxiv}, we add a rectangular avoid region (Fig.~\ref{fig:Roomba}).  The reward is the expectation over the state reward that is a large reward for reaching the goal, large penalty for reaching the stairs and for each time instance.

\paragraph{Dangerous Light Dark}
We take inspiration from the one dimensional problem from \cite{Jamgochian23ICAPS}. The agent lives in a one dimensional space. We reach versatility of action space by the length of actions, such that
%
$	\mathcal{A} = \{0, \pm 0.5, \pm 1, \pm 1.5, \pm 2, \pm 2.5, \pm 6 \}. $
%
The agent's reward is the multi-objective and subsumes the expected state-dependent reward and the belief-dependent reward to localize itself
\begin{align} 
	\rho_{\ell+1}(b_{\ell}, a_{\ell}, b_{\ell+1}) {=} \mathbb{E}_{x \sim b_{\ell}}[r(x, a)] -\mathrm{tr}(\Sigma(b_{\ell+1})) \label{eq:LDreward}
\end{align}
where $\Sigma(b_{\ell+1})$ is the covariance matrix of $b_{\ell+1}$.
The agent’s state dependent goal is to get to location defined by interval $[-0.75, 0.75]$ as fast as possible and execute the action $0$ to stay there. Executing it within the interval $[-0.75, 0.75]$
will give the agent a reward of $100$, and executing it outside the radius will yield a negative reward of $-100$. For all other actions the state dependent  reward function is $-\mathrm{abs}(x)$.
The agent's motion model $\mathrm{T}$ is specified as  
\begin{align}
	x_{k+1} {=} x_k + a_k+w_k,  \label{eq:LDmotion}
\end{align}
where $w_{k}$ follows truncated Gaussian with $\sigma {=} 0.1$ and truncation with $\Delta {=} 0.5$ around nominal value $x_k{+}a_k$. The light region is located at $x{=}2$ and the observation model is
%
	$z_{k} = x_k{+}v_k$
%
where $v_k{\sim} \mathcal{N}(0, \sigma(x_k))$ and $\sigma(x) {=} \mathbf{1}_{\{x: |x {-} 2| \leq 1\}}(x)\cdot 10^{-10} {+} \mathbf{1}_{\{x: |x {-} 2| > 1\}}(x){\cdot} |x{-}2|$.  
At $x{=}{-}0.75$ there is a cliff such that if agent falls it crashes. In addition around the light source there is a pit. The safe space is 
%
	$\mathcal{X}^{\mathrm{safe}} {=} \{ -0.75 < x < 1\} \cap\{ x > 3\}. $	
%
The prior belief $b_k(x_k)$  is Gaussian  $\mathcal{N}(7,20)$ truncated such that its support is $[6, 8]$.

\paragraph{Simultaneous Localization and Mapping with Certain and Uncertain Obstacles (SLAM)}
\label{sec:SLAM}
Our action space comprises motion primitives and zero action,
%
	$\mathcal{A} = \{  \rightarrow, \nearrow, \uparrow, \nwarrow,\leftarrow,  \swarrow,   \downarrow, \searrow, \boldsymbol{0}\}. $	
%
If robot selected zero action $\boldsymbol{0}$, we do not apply motion model to each particles but do resampling to take into account received observation. This allows to robot not move if it is too dangerous.
In this problem the agent and the uncertain obstacles (landmarks) have circular form.  The motion model $\mathrm{T}$ for the agent is 
\begin{align}
	x_{k+1} = x_k + a_k + w_k. \label{eq:SLAMtransitionAgent}
\end{align}
Our goal is to epitomize the importance of safe state trajectories versus solely safe beliefs trajectory. Towards this end we draw randomly many tiny obstacles so as  one way or another the unsafe trajectory will be encountered by the robot if planning was done with pushing forward in time also the unsafe particles. Our observation model is bearing range with the noise inversely proportional to the distance to uncertain obstacle, the landmark $l$. The motion model for the landmark is 
\begin{align}
	l_{k+1} = l_k. \label{eq:SLAMtransitionLandmark}
\end{align}
We maintain belief over the last robot pose and the landmark. 
The observation model reads 
%
	$z_{k} = x_k - l_k + v_k $
%
where $v_k \sim \mathcal{N}(0, \Sigma_k(x_k, l_k))$. The $\Sigma_k(x_k, l_k)$ is a diagonal matrix  with main diagonal $\sigma^2_k(x_k, l_k)){=}\|x_k - l_k\|_2$. 

\subsubsection{Pushbox 2D Problem}
\label{sec:PushBox2Ddescription}
In this section we first describe our variation of PushBox2D problem with soft safety. We then transfer the soft safety to our formulation described in Section~\ref{sec:PF}. Clearly soft safety is not good enough. In cases there is no feasible solution exists we do not want robot to do any operations. Instead, it is desirable that robot decide that the goal is not achievable.  
\begin{figure*}[t] 
	\centering
	\begin{minipage}[t]{0.35\textwidth}
		\centering 
		\includegraphics[width=0.7\textwidth]{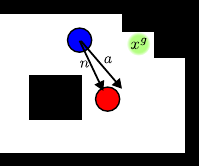}
		\subcaption{}
	\end{minipage}%
	\hfill
	\begin{minipage}[t]{0.35\textwidth}
		\centering 
		\includegraphics[width=0.7\textwidth]{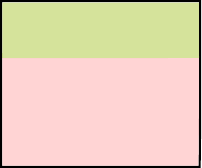}
		\subcaption{}
		\label{fig:PushBoxNoise}
	\end{minipage} 
	\caption{\textbf{(a)} Conceptual visualization of the PushBox2D problem. The agent is the blue circle. The puck is the red circle. The goal is the green circle. \textbf{(b)} The observation noise intensity map. Light green color denotes the lower noise intensity. } 
	\label{fig:PushBox}
\end{figure*}
The Pushbox2D problem is motivated by air hockey. A disk-shaped robot (blue disk) must push a disk-shaped puck (red disk) into a goal area (green circle) by bumping into it while avoiding any collision of itself and the puck with an edge area (black area).
The state space consists of the $xy$-locations of both the robot and the puck, i.e., $\mathcal{X} {=} \mathbb{R}^4,$  while the action space is defined by motion primitives of unit length . The action $\mathrm{Null}$ is terminal.  
If the robot is not in contact with the puck during a move, the state evolves according to 
\begin{equation}
\begin{gathered}
	x' {=} (f(x, a) + w, (x^p , y^p))^T,  \quad w {\sim} \mathcal{N}(0, W), \\
	 f (x, a) {=} (x^r + a^x , y^r + a^y )^T, 
\end{gathered}
\end{equation}
where $(x^r , y^r )$ and $(x^p , y^p )$ are the $xy$-coordinates the robot and the puck respectively, corresponding to state $x$, and $(a^x , a^y )$ is the displacement
vector corresponding to action $a$. 
If the robot bumps into the puck, the next position $(x'^p , y'^p)$ of the puck is
\begin{align}
	\textstyle\begin{pmatrix} x'^{p} \\ y'^p\end{pmatrix} = \begin{pmatrix} x^{p} \\ y^p\end{pmatrix} {+}  5r_s\begin{pmatrix} \begin{pmatrix} a^{x} \\ a^y\end{pmatrix} \cdot n\end{pmatrix} \begin{pmatrix} n {+} \begin{pmatrix}  r^x \\ r^y \end{pmatrix} \end{pmatrix},  
\end{align}
where $n$ is the unit directional vector from the center of the robot to the center of
the puck at the time of contact, and $r_s$ is a random variable drawn from a truncated Gaussian distribution $\mathcal{N}(\mu, \sigma^2 , l, u)$,
which is the Gaussian distribution $\mathcal{N}(\mu, \sigma^2)$ truncated to the interval $[l, u]$.  The variables
$r_x$ and $r_y$ are random variables drawn from a truncated Gaussian distribution $\mathcal{N} (0.0, 0.1 2 , -0.1, 0.1)$.  The prior belief $b_0(x_0)$ is a Gaussian over the robot position and deterministic over the puck position.
The robot has access to a noisy bearing sensor to localize itself observing the puck and a noise-free collision sensor which detects contacts between the robot and the puck. Specifically, given a state $x {\in} \mathcal{X}$, an observation $(o_c , o_b )$ consists of a binary component $o_c$ which indicates
whether or not a contact between the robot and the puck occurred, and a  bearing range component $o^{\mathrm{br}}$ calculated as
%
%
\begin{align}
	o^{\mathrm{br}} {=} h(x) {+} v, \ h(x) {=} (x^r {-}x^p , y^r {-}y^p )^T, 
\end{align}
where $x^r$ , $y^r$ and $x^p$ , $y^p$ are the $xy$-coordinates of the robot and the puck corresponding to the state $x$, and $r_o$ is a
random angle (expressed in radians) drawn from a truncated Gaussian distribution with magnitude of the variance dependent on the position on the map of the robot as in the Fig.~\ref{fig:PushBoxNoise}).
The reward for the MCTS baseline  is the distance to goal of the puck with boundary region and other obstacles
\begin{equation}
\begin{gathered}
	\rho(b) {=} -\mathbb{E}_{x \sim b}[\|x^p-x^g\|_2] {-}1000 \cdot \mathbf{1}_{\{ o_c == 1\}}(o_c){+}\\
	{\color{red}{\prob(\{x {\in} \mathcal{X}^{\mathrm{safe}}\}|b)}}{-}\mathcal{H}(b).
\end{gathered}
\end{equation}
Here we have a soft chance constraints since it is not clear how to enforce chance constraints to MCTS.  In case of our approach we shift the $\prob(\{x {\in} \mathcal{X}^{\mathrm{safe}}\}|b)$ component to our probabilistic constraint.

\subsection{Experiments}
\begin{figure*}[t]
	\centering
	\begin{minipage}[t]{0.35\textwidth}
		\centering 
		\includegraphics[width=\textwidth]{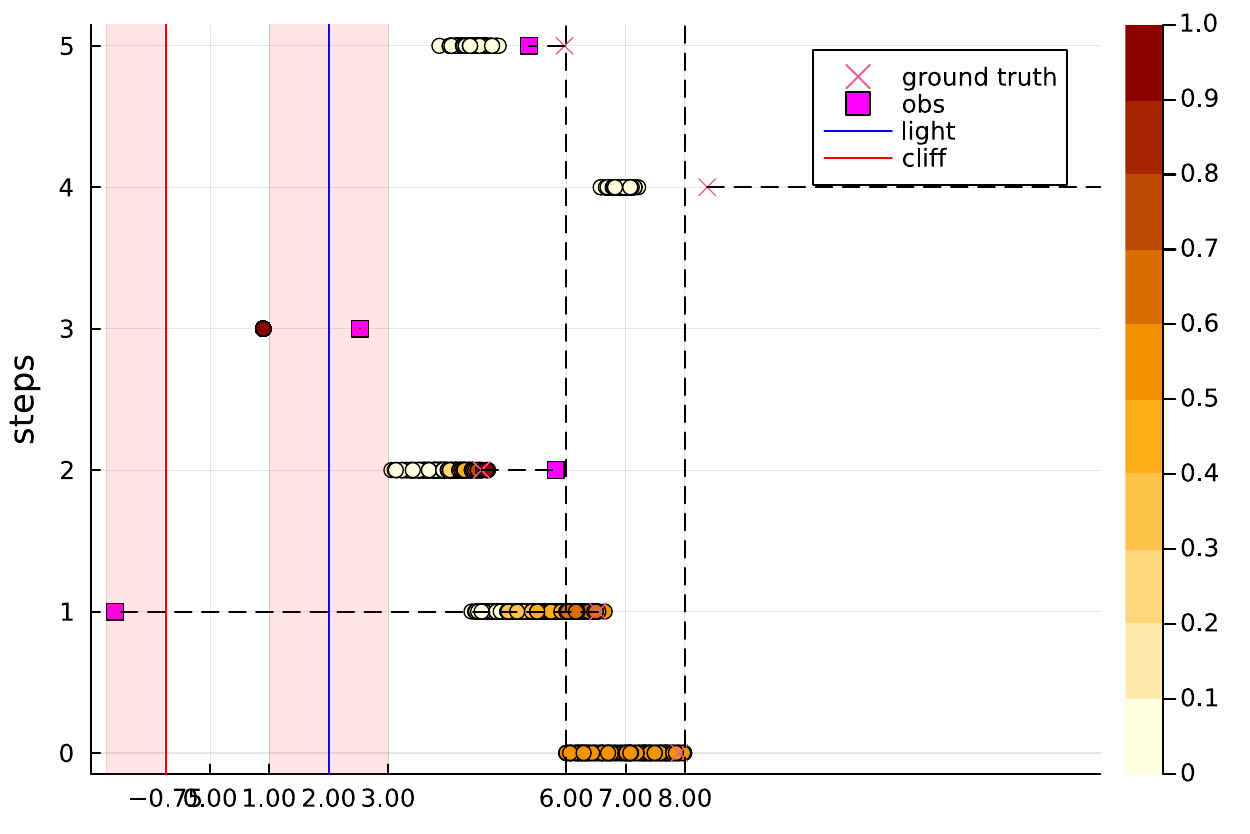}
		\subcaption{}
		\label{fig:lightdarkCPFT}
	\end{minipage}%
	\hfill
	\begin{minipage}[t]{0.35\textwidth}
		\centering 
		\includegraphics[width=\textwidth]{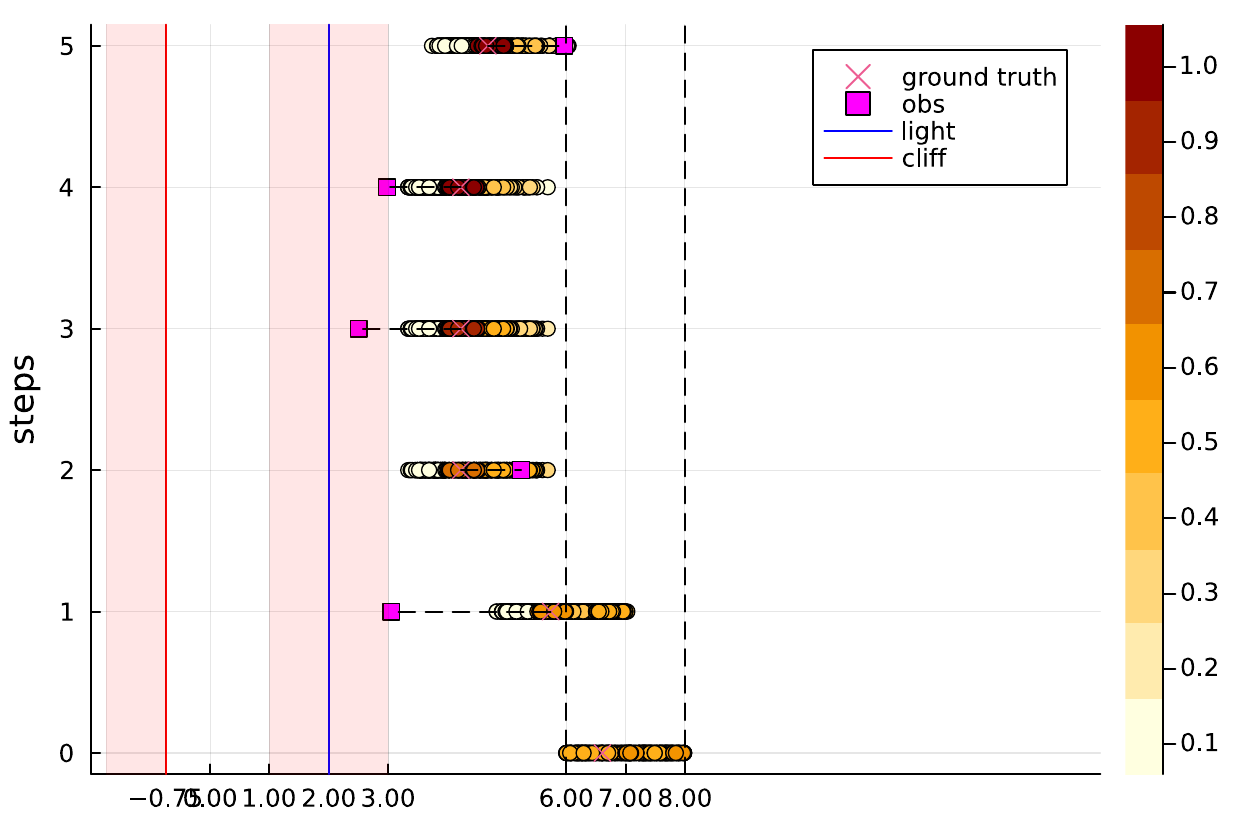}
		\subcaption{}
		\label{fig:lightdarkCCPCPFT}
	\end{minipage} 
	\caption{Illustration of faulty scenario of CPFT as opposed to our approach \textbf{(a)} CPFT \textbf{(b)} Our Alg.~\ref{alg:PCMCTS}.}
	\label{fig:lightdark}	 
\end{figure*}
We benchmarked our approach using the Lidar Roomba problem, the famous Light Dark problem, active SLAM problem and PushBox2D problem. We have shown the issue described in Section~\ref{sec:DualityBasedApproach} on Lidar Roomba and the satisfiability of the constraint solely at the limit of MCTS convergence situation on a Light Dark problem. Considering an active SLAM problem, we visualized the importance of making belief safe in planning, as described in Section~\ref{sec:IS}.  With  PushBox2D problem we verified the importance of making the propagated belief safe and simulated for several values of $\delta$.  

In the Lidar Roomba problem the robot performs at most $50$ cycles of autonomy loop.  In the Light Dark problem, the robot performs $5$ cycles of autonomy loop. 
We do $70$ trials of each such a scenario and approximate the $\mathrm{P}(S|b_0) {\approx} \hat{\mathrm{P}}(S|b_0) {=} \sum_{i=1}^{70} \nicefrac{\mathbf{1}_S(\tau^i_0)}{70}$ using the simulated trajectories. The event $S{=}\{\tau_0 {\in} \times_{\ell=0}^{L}\mathcal{X}^{\mathrm{safe}}_{\ell}\}$, where $\tau_0 {=} x_{0:L}$ means each state in the actual robot trajectory starting at time $0$ was safe. $\hat{V}^*(b_0;\boldsymbol{\rho}_1) {=} \frac{1}{70}\sum_{i=1}^{70} \sum_{\ell=0}^{L(i)} \rho_{\ell+1}(b^i_{\ell}, a^i_{\ell}, b^i_{\ell+1})$, where in Roomba $L(i){\leq} 50$ since we have terminal state and in Light Dark $L(i){\equiv} 5$. 

In SLAM problem the robot makes $50$ trials of at most $20$ cycles of autonomy loop. In PushBox2D problem the robot performs $20$ trials of at most $20$ cycles of autonomy loop.

In all four problems we take $500$ belief particles. 
%
%
%
\begin{table*}[t]
	\begin{minipage}[t]{.48\textwidth}
		\vspace{0pt}
		\centering
		\resizebox{\textwidth}{!}{
			\begin{tabular}{|c|c|c|c|c|c|c|c|c|c|}
				\toprule
				\textbf{Model}  & $\substack{ {\text{tree }} \\ {\text{queries}}}$ & $\hat{V}^*(b_0; \boldsymbol{\rho}_1)$  & $ \hat{\mathbb{E}}[\|x^{t} - x^{g}\|^2_2]{\pm} \mathrm{std} $ &$ \hat{\mathrm{P}}(S|b_0)$  & num. coll\\
				\midrule
				\texttt{CPFT-DPW}  &$1000$   &$-46500.0 {\pm} 136.31$ &  ${\color{red}{829.78 {\pm} 525.88}}$     & $1$ &$0/70$\\
				\midrule
				{\bf PCPFT-DPW}  &$1000$     &$ -28086 {\pm} 14399 $ & $56.86 {\pm} 215.12$    & $1$ & $0/70$\\
				\bottomrule
		\end{tabular}}
		\caption{The Lidar Roomba problem. }
		\label{Tbl:depart}
		\vspace{0pt}
	\end{minipage}%
	\hfill
	\begin{minipage}[t]{.35\textwidth}
		\vspace{0pt}
		\centering
		\resizebox{\textwidth}{!}{
			\begin{tabular}{|c|c|c|c|c|c|c|c|c|}
				\toprule
				\textbf{Model}  &  $\substack{ {\text{tree }} \\ {\text{queries}}}$   &$\hat{V}^*(b_0; \boldsymbol{\rho}_1)$  & $\hat{\mathrm{P}}(S|b_0)$ &  num. coll\\
				\midrule
				\texttt{CPFT-DPW}  & $15$  & $-75.67 \pm 57.66$ &    ${\color{red}{0.77}}$    & ${\color{red}{16/70}}$\\
				\midrule
				{\bf PCPFT-DPW}&   $15$   &     $ -115.27 \pm 94.28 $ &          $1$             & $0/70$ \\ 
				\bottomrule
		\end{tabular}}
		\caption{The Light Dark problem. }
		\label{tbl:LDprop}
		\vspace{0pt}
	\end{minipage} 
\vspace{0pt}
\end{table*}

\subsection{Discussion and Results Interpretation}

Before we proceed it shall be noted that the number of collisions and approximated probability that the trajectory is safe in relevant tables are connected as follows
\begin{align}
	\textstyle\hat{\mathrm{P}}(\{\tau_0 \in \times_{\ell=0}^{L}\mathcal{X}^{\mathrm{safe}}_{\ell}|b_0) = \hat{\mathrm{P}}(S|b_0) = 1 {-} \frac{\text{num. coll}}{\text{num. trials}}.  \label{eq:ProbSafeTraj}
\end{align}
Let us interpret the results.
\paragraph{Roomba}
Table~\ref{Tbl:depart} corresponds to Roomba problem.
From Table~\ref{Tbl:depart} we behold that the cumulative reward yielded by CPFT is much lower than our method. In particular, the Roomba never reaches the goal and not stairs. We also calculate an empirical mean of the distance between the terminal Roomba position and the middle of the goal region. As we see, CPFT makes Roomba to depart from the obstacle as far as possible.   

\paragraph{Light Dark}
Table~\ref{tbl:LDprop} corresponds to Light Dark problem. In Table~\ref{tbl:LDprop} we see that with a small number of MCTS iterations, CPFT makes $16$ collisions from $70$ trials in contrast to  $0$ collisions with our technique.
We illustrate the scenario in Fig.~\ref{fig:lightdark}. In this problem it is dangerous to the agent to jump to desired interval. This is because the width of the belief $b_0$ is larger than the desired area and robot can fall off the cliff or to the pit (assuming the motion model as in \eqref{eq:LDmotion} and without the stochastic noise $w_k$). Our approach prevent the robot to jump to desired area since any belief particle can be the ground truth.    

\paragraph{SLAM}
\begin{figure*}[t]
	\centering
	\begin{minipage}[t]{0.24\textwidth}
		\centering 
		\includegraphics[width=\textwidth]{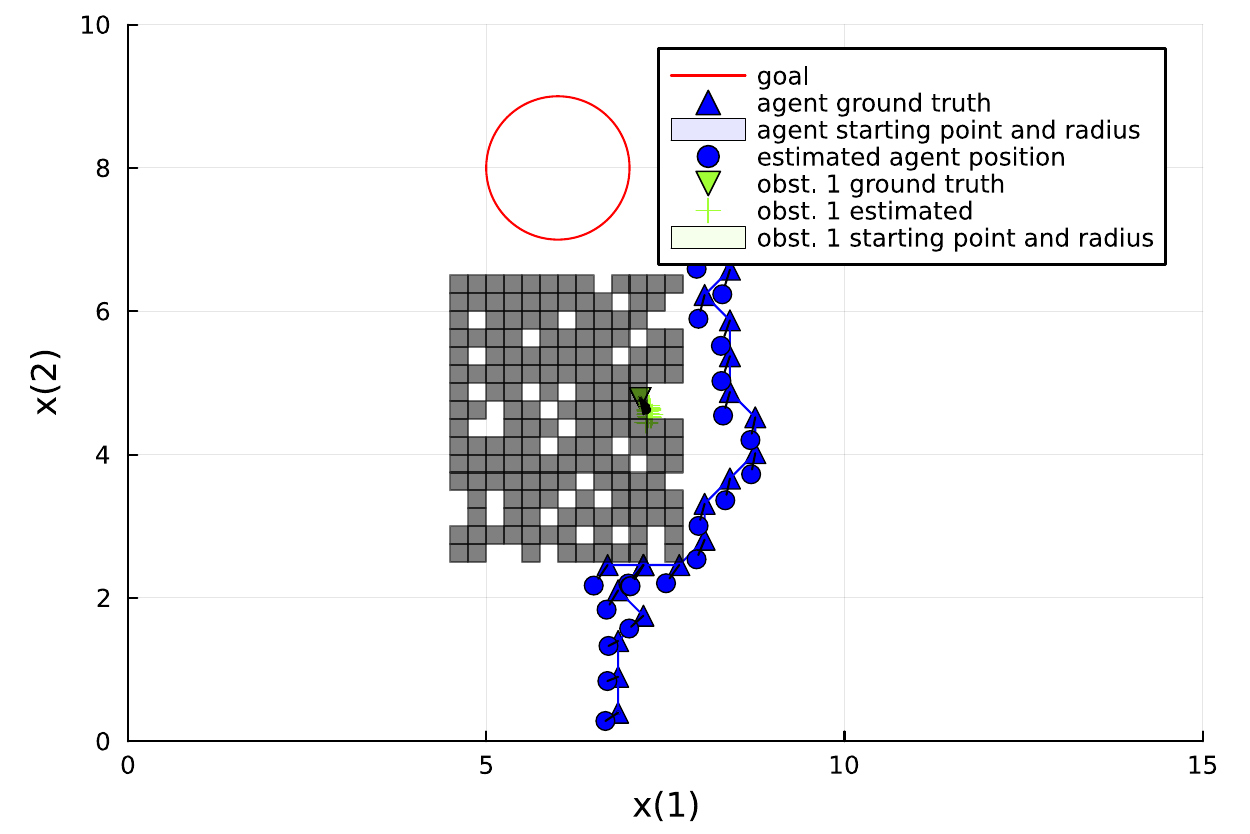}
		\subcaption{}
		\label{fig:RandomTinyAgentGTandEST}
	\end{minipage}%
	\hfill
	\begin{minipage}[t]{0.24\textwidth}
		\centering 
		\includegraphics[width=\textwidth]{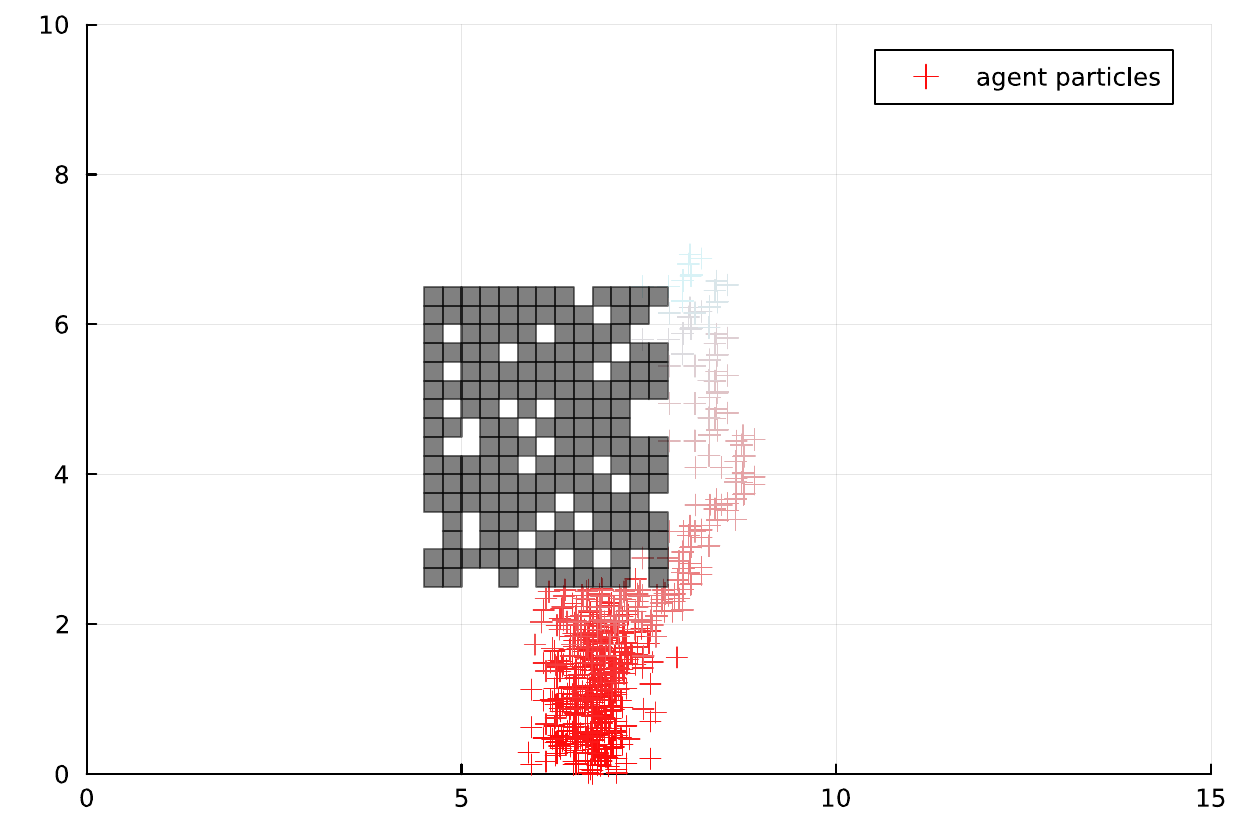}
		\subcaption{}
		\label{fig:RandomTinyParticles}
	\end{minipage} 
	\begin{minipage}[t]{0.24\textwidth}
		\centering 
		\includegraphics[width=\textwidth]{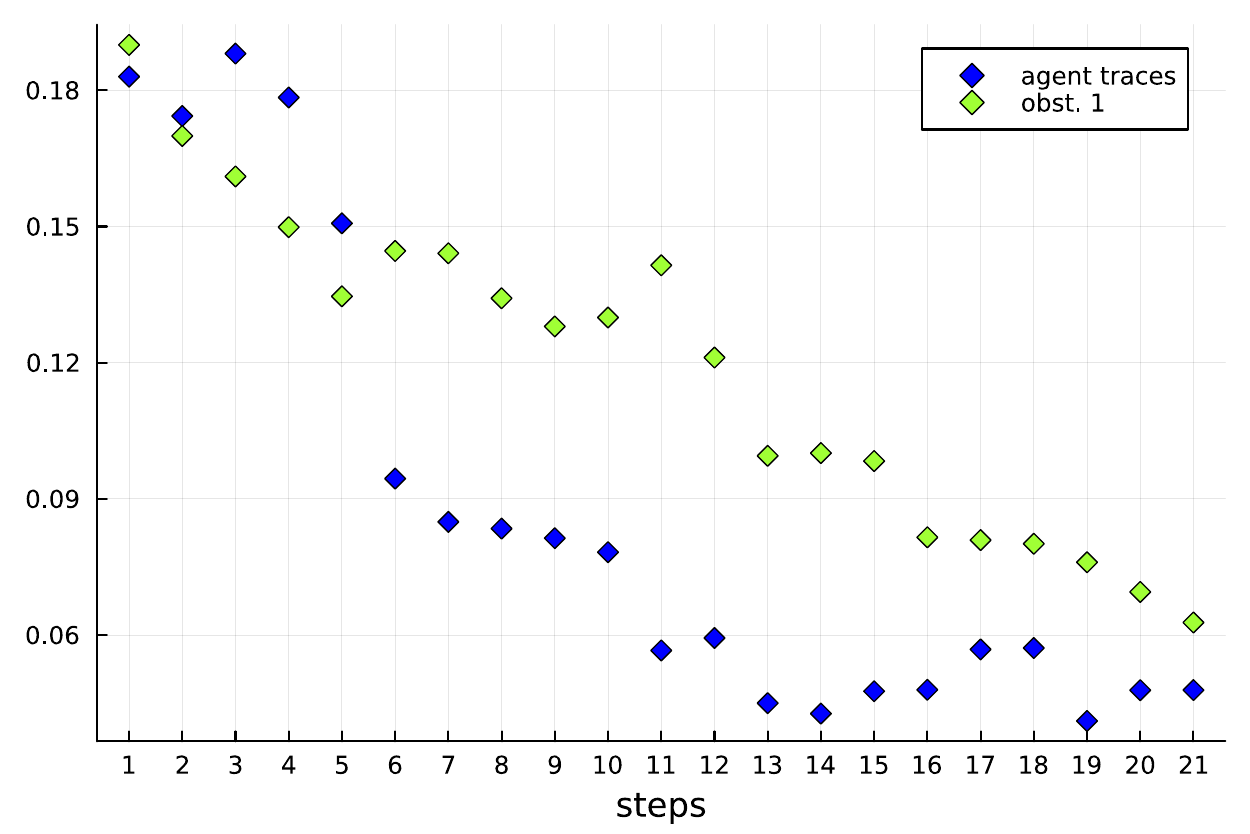}
		\subcaption{}
		\label{fig:RandomTinyTraces}
	\end{minipage} 
	\hfill
	\begin{minipage}[t]{0.24\textwidth}
		\centering 
		\includegraphics[width=\textwidth]{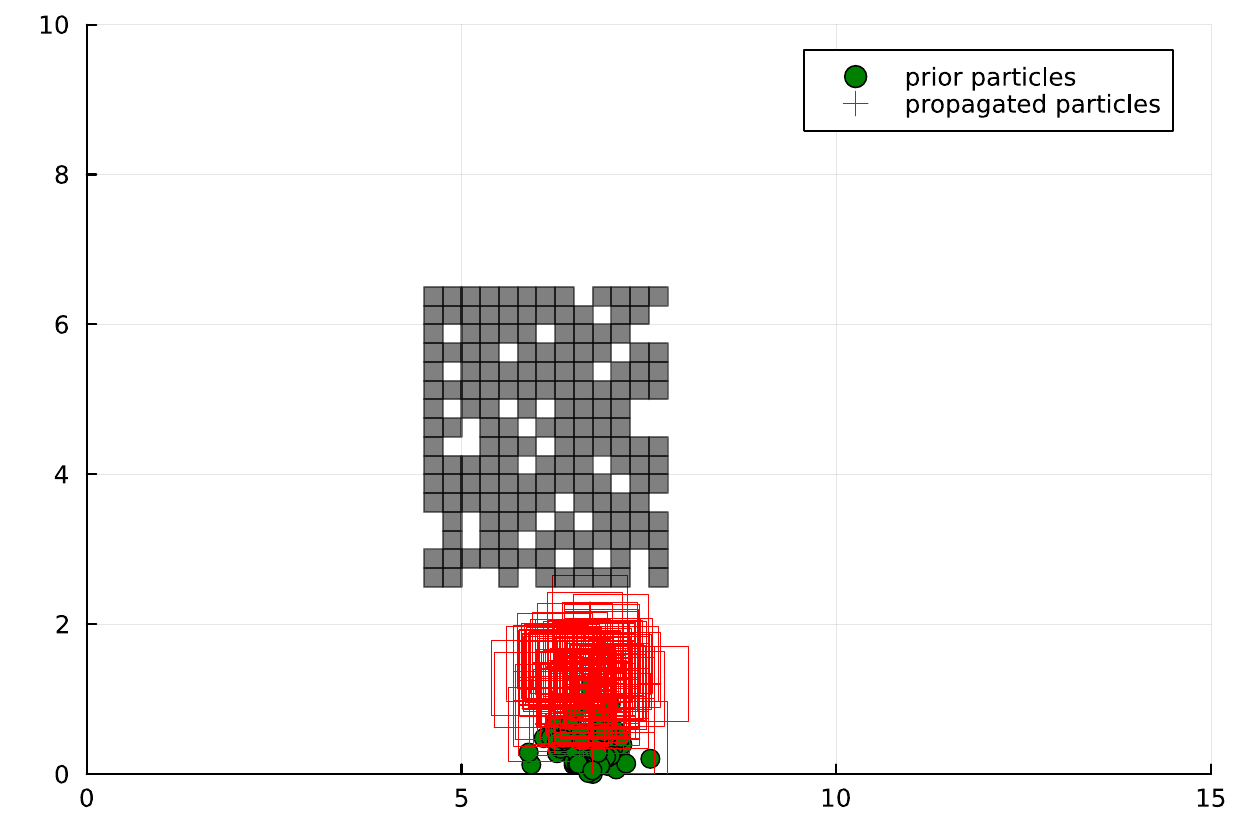}
		\subcaption{}
		\label{fig:RandomTiny}
		\hfill
	\end{minipage} 
	\caption{This simulation setup is associated with Table~\ref{tbl:CollsRewardsSLAMRandomTiny}. In this figure we plot one of the trials shown in Table~\ref{tbl:CollsRewardsSLAMRandomTiny}. Here we nullify unsafe part of the belief in planning and run CCPC-PFT. \textbf{(a)} Here, we plot the goal, agent ground truth, estimated agent positions and the obstacles; \textbf{(b)} Belief particles, where the colors symbolize the time instance;  \textbf{(c)} Traces of the agent and the landmark (obstacle); \textbf{(c)} Visualization of the truncation. Here we move each particle of $b_0$ with action selected by the agent and plot the truncation region of the stochastic motion model.}      
	\label{fig:CCPCPFTSLAMTiny}	 
\end{figure*}
\begin{figure*}[t]
	\centering
	\begin{minipage}[t]{0.24\textwidth}
		\centering 
		\includegraphics[width=\textwidth]{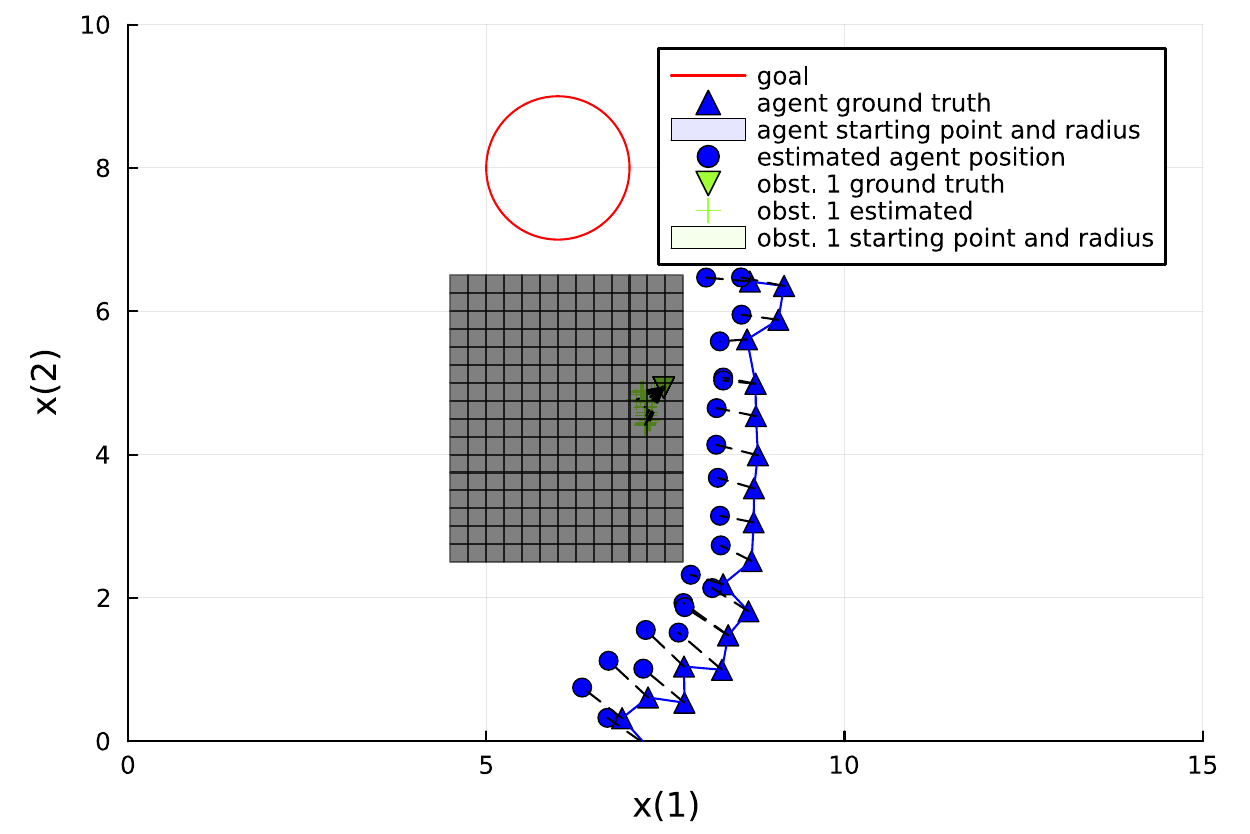}
		\subcaption{}
		\label{fig:}
	\end{minipage}%
	\hfill
	\begin{minipage}[t]{0.24\textwidth}
		\centering 
		\includegraphics[width=\textwidth]{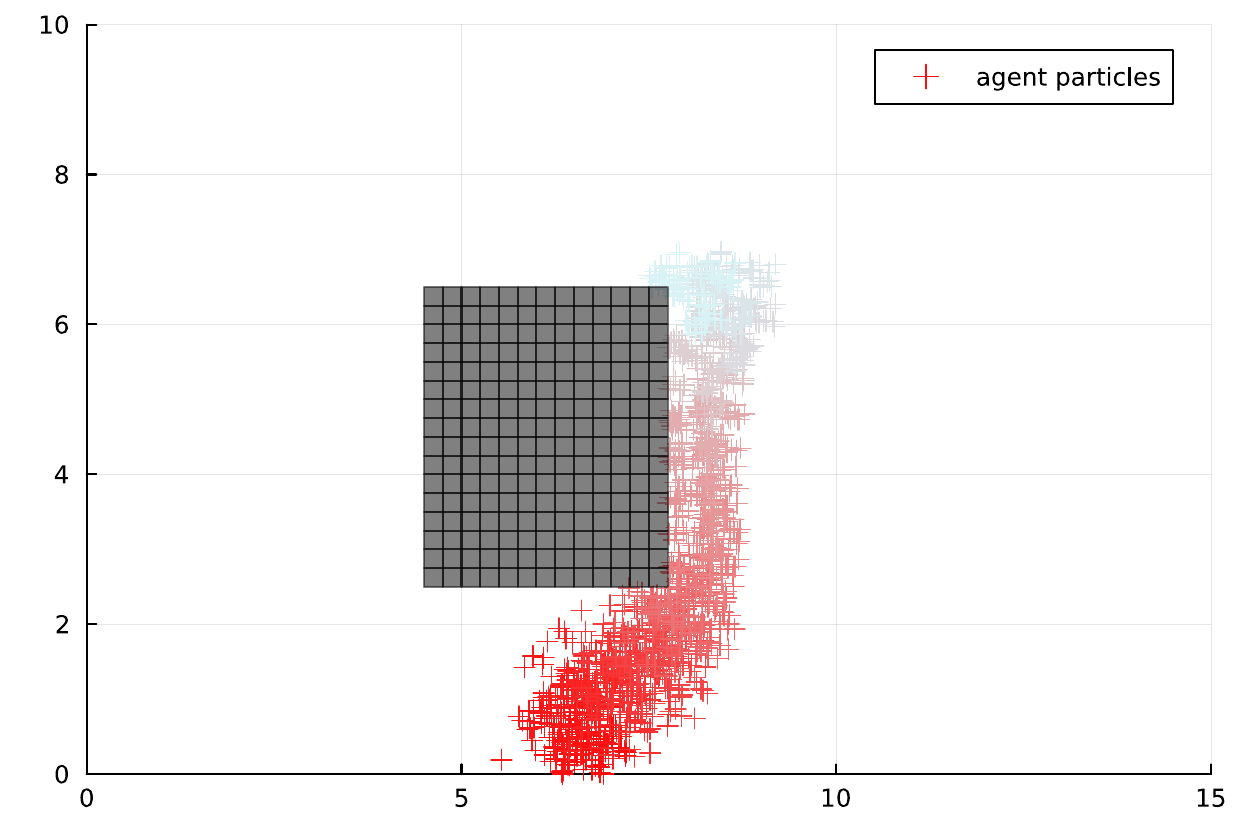}
		\subcaption{}
		\label{fig:}
	\end{minipage} 
	\begin{minipage}[t]{0.24\textwidth}
		\centering 
		\includegraphics[width=\textwidth]{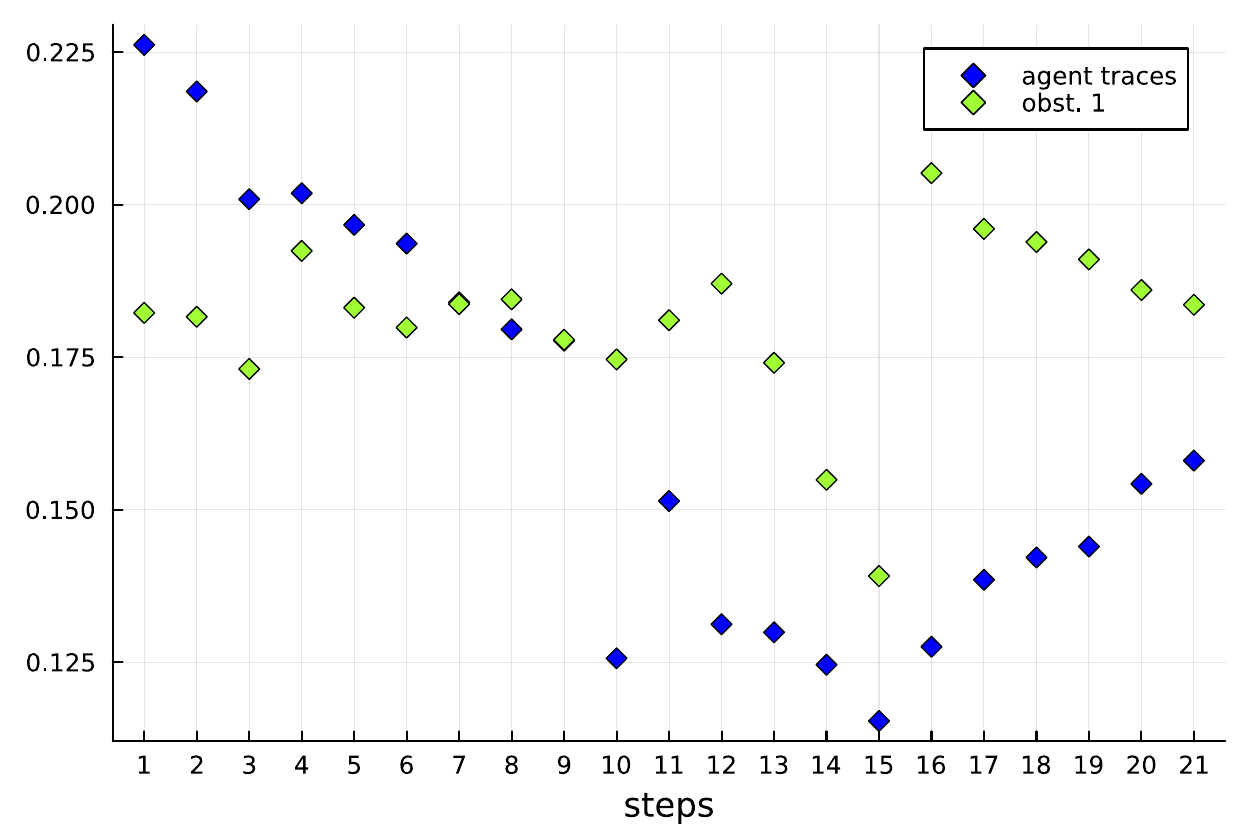}
		\subcaption{}
		\label{fig:}
	\end{minipage} 
	\hfill
	\begin{minipage}[t]{0.24\textwidth}
		\centering 
		\includegraphics[width=\textwidth]{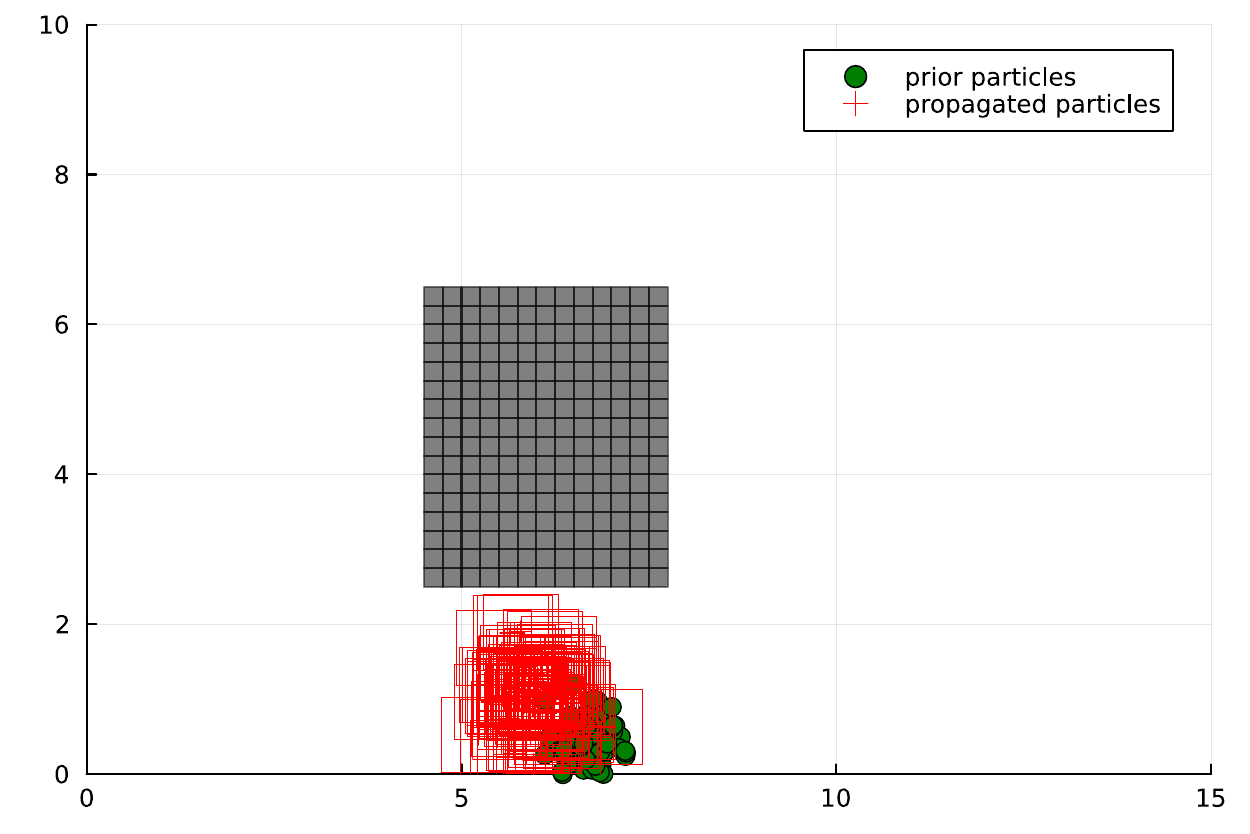}
		\subcaption{}
		\label{fig:}
		\hfill
	\end{minipage} 
	\caption{This simulation setup is associated with Table~\ref{tbl:CollsRewardsSLAMRandomTinyFull} columns related to CCPC-PFT-DPW and here we show one of the trials. In this figure we nullify unsafe part of the belief in planning. \textbf{(a)} Here, we plot the goal, agent ground truth, estimated agent positions and the obstacles; \textbf{(b)} Belief particles, where the colors symbolize the time instance;  \textbf{(c)} Traces of the agent and the landmark (obstacle); \textbf{(c)} Visualization of the truncation. Here we move each particle of $b_0$ with action selected by the agent and plot the truncation region of the stochastic motion model.}      
	\label{fig:CCPCPFTSLAMTinyFull}	 
\end{figure*}
\begin{figure*}[t]
	\centering
	\begin{minipage}[t]{0.24\textwidth}
		\centering 
		\includegraphics[width=\textwidth]{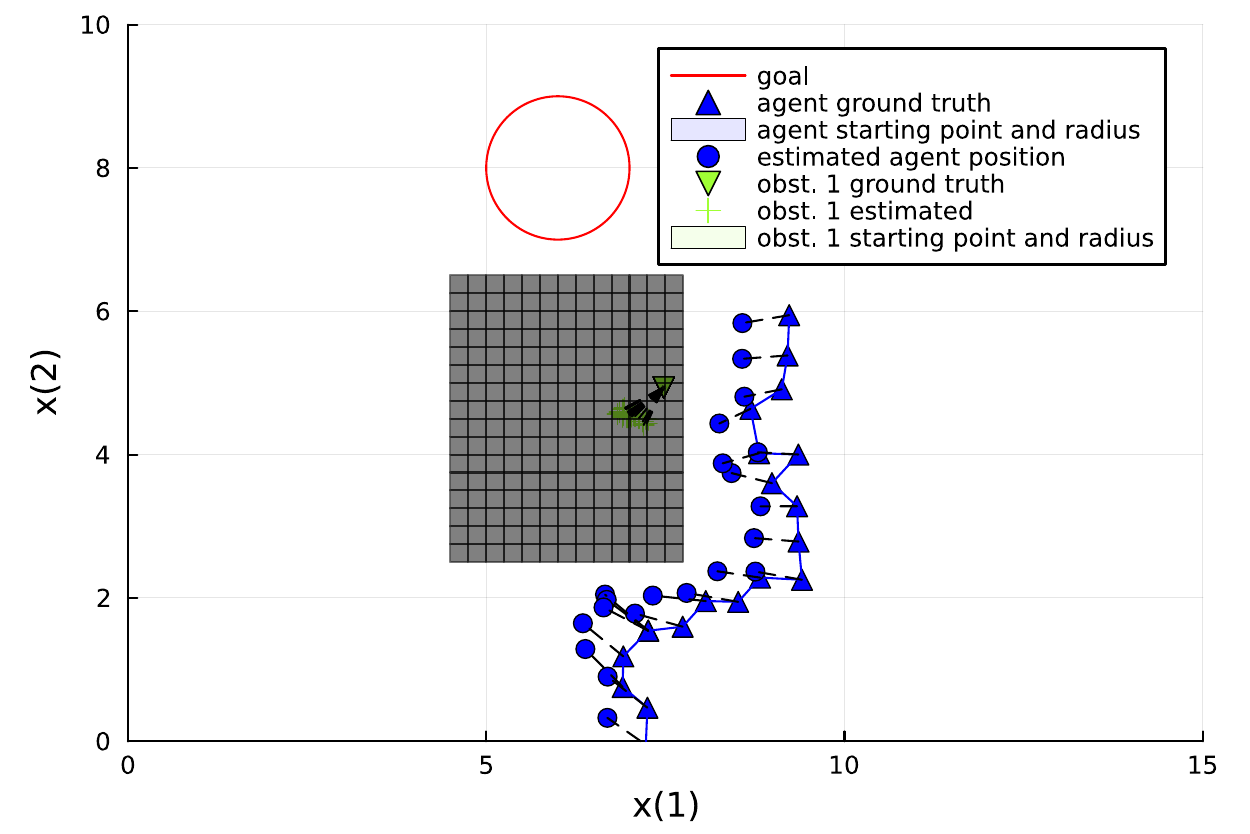}
		\subcaption{}
		\label{fig:}
	\end{minipage}%
	\hfill
	\begin{minipage}[t]{0.24\textwidth}
		\centering 
		\includegraphics[width=\textwidth]{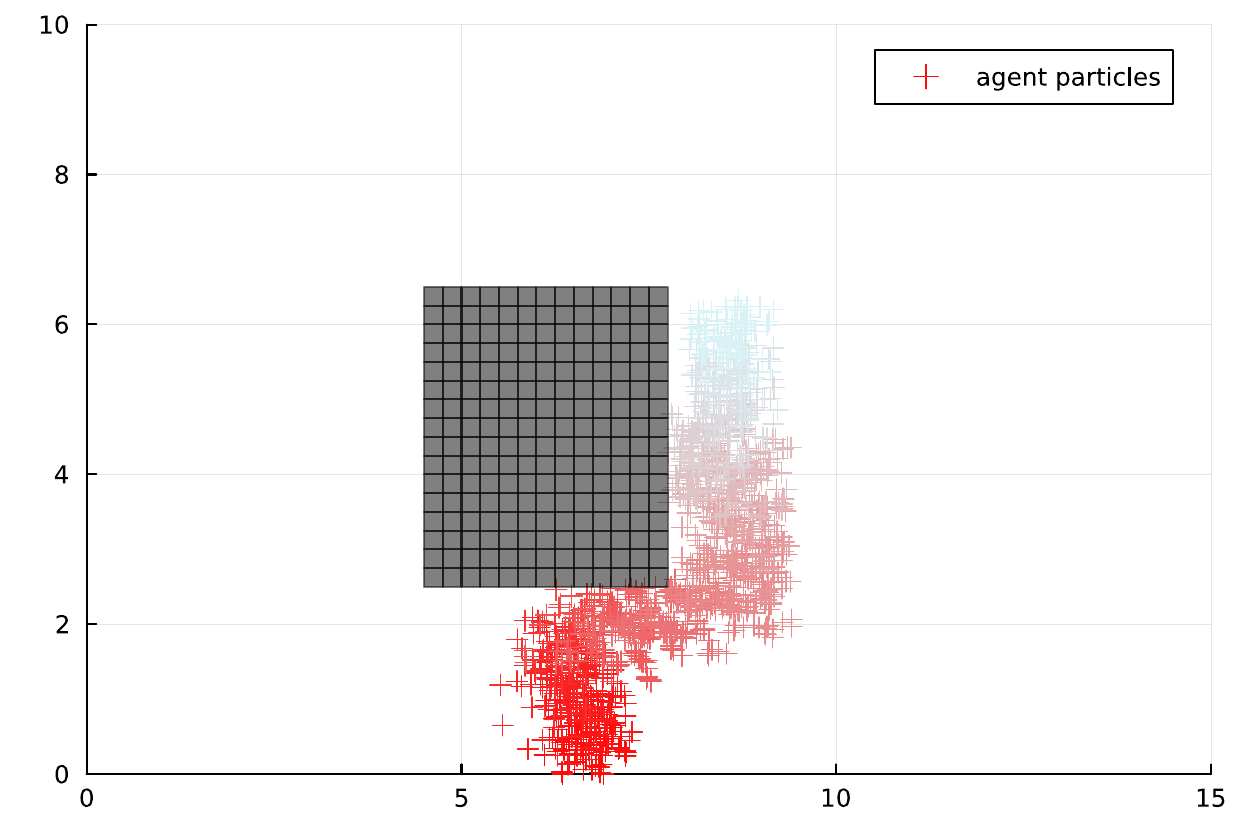}
		\subcaption{}
		\label{fig:}
	\end{minipage} 
	\begin{minipage}[t]{0.24\textwidth}
		\centering 
		\includegraphics[width=\textwidth]{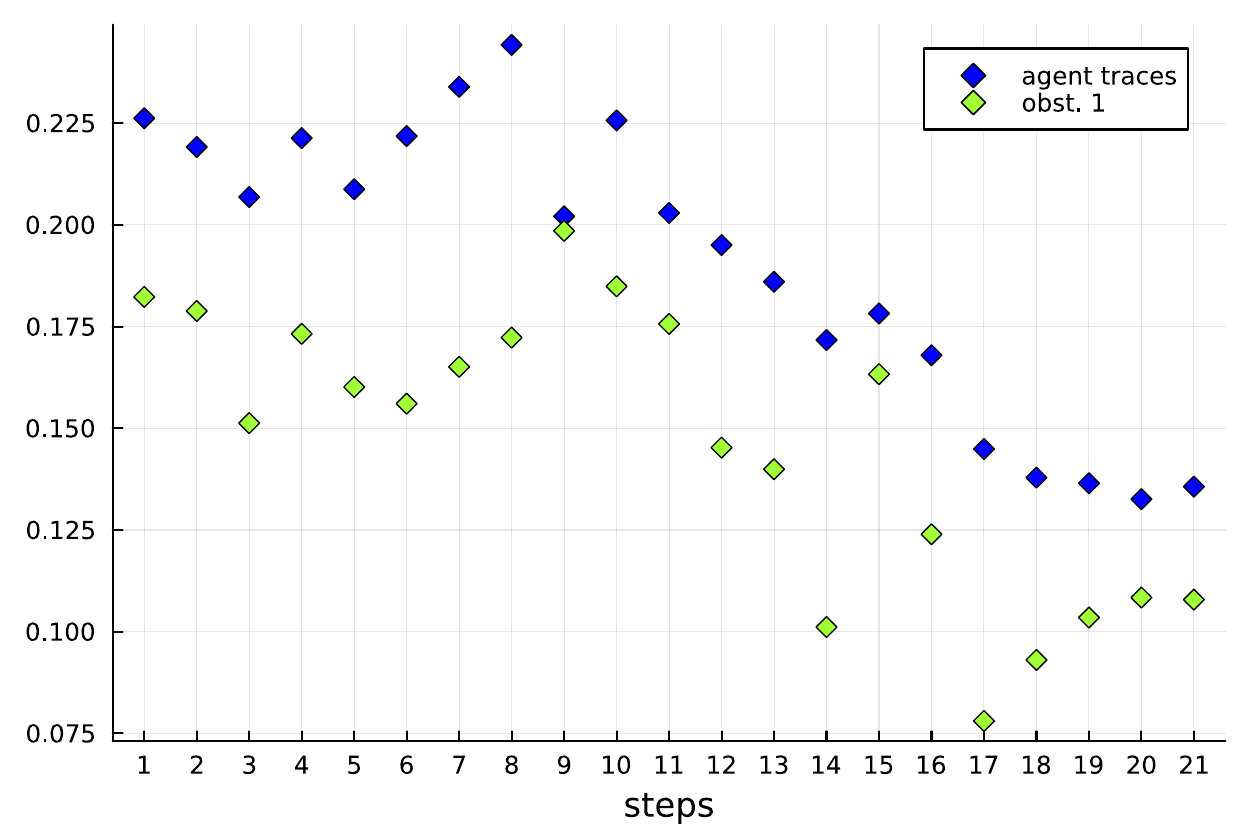}
		\subcaption{}
		\label{fig:}
	\end{minipage} 
	\hfill
	\begin{minipage}[t]{0.24\textwidth}
		\centering 
		\includegraphics[width=\textwidth]{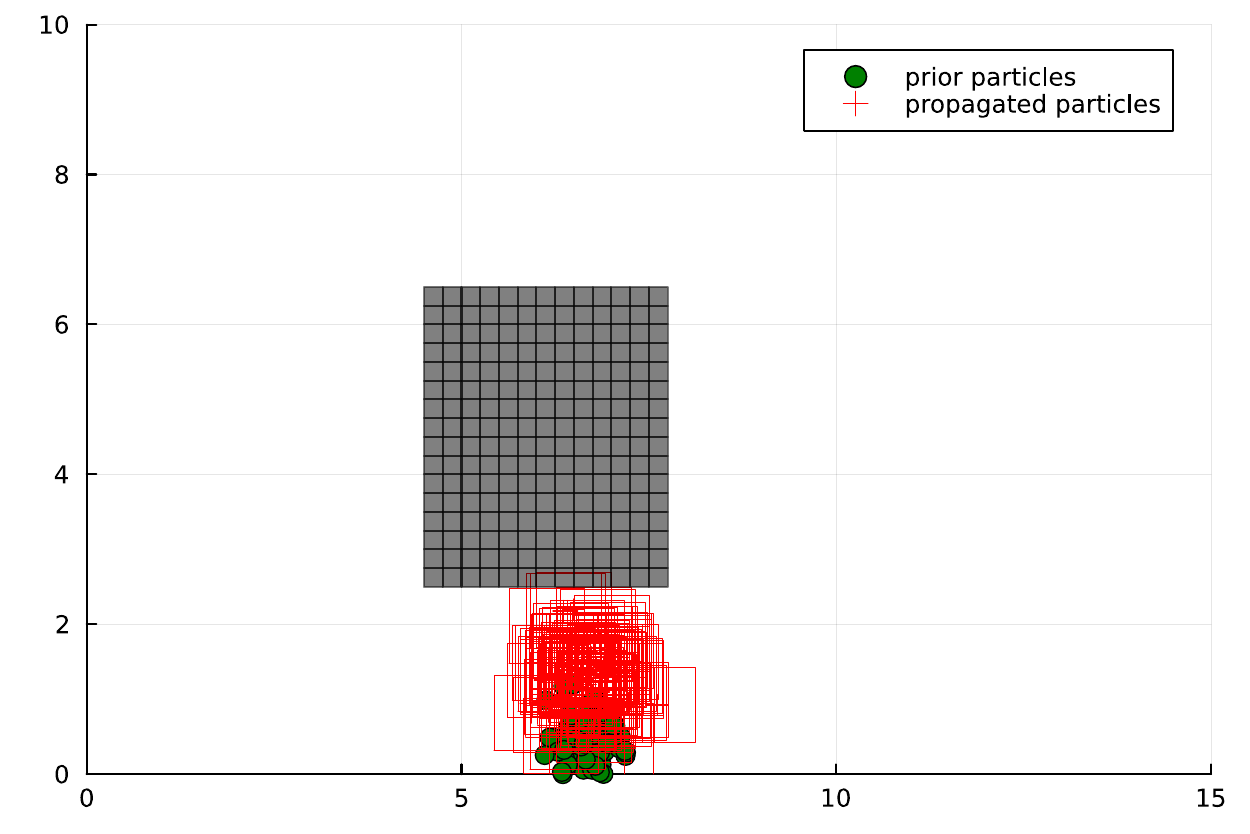}
		\subcaption{}
		\label{fig:}
		\hfill
	\end{minipage} 
	\caption{This simulation setup is associated with Table~\ref{tbl:CollsRewardsSLAMRandomTinyFull} columns related to PC-PFT-DPW and here we show one of the trials. In this figure we {\bf do not} nullify unsafe part of the belief in planning. \textbf{(a)} Here, we plot the goal, agent ground truth, estimated agent positions and the obstacles; \textbf{(b)} Belief particles, where the colors symbolize the time instance;  \textbf{(c)} Traces of the agent and the landmark (obstacle); \textbf{(c)} Visualization of the truncation. Here we move each particle of $b_0$ with action selected by the agent and plot the truncation region of the stochastic motion model.}      
	\label{fig:PCPFTSLAMTinyFull}	 
\end{figure*}
\begin{table*}[h]
	\caption{ $50$ Trials of at most $20$ cycles of autonomy loop Fig.~\ref{fig:AutonomyLoop} where planning sessions implemented by  Algorithm CCPC-PFT-DPW versus PC-PFT-DPW. Same seed in both  algorithms. This problem is the {\bf SLAM} described in Section \ref{sec:SLAM} in { \bf our first scenario} shown at Fig.~\ref{fig:CCPCPFTSLAMTiny}. Here we study the number of collisions and the reward value.}
	\centering
	\resizebox{\textwidth}{!}{
	\begin{tabular}{|c|c|c|c|c|c|c|c|}
		\hline
		\multicolumn{2}{|c|}{ Parameters } & \multicolumn{2}{|c|}{$\hat{\mathrm{P}}(S|b_0)$} &\multicolumn{2}{|c|}{ num coll. } &  \multicolumn{2}{|c|}{ mean cum. rew. $\pm$ std}  \\
		\hline
		Operator $\phi$  & $\delta$  &  CCPC-PFT-DPW   & PC-PFT-DPW  &  CCPC-PFT-DPW & PC-PFT-DPW &  CCPC-PFT-DPW & PC-PFT-DPW  \\
		\hline  
		\eqref{eq:ProbSafeGivenBelief} & $0.8$ & $0.64$ & - &$18/50$ &  -  & $-106.37 \pm 12.37$  &   -  \\
		\hline
	\end{tabular}
}
	\label{tbl:CollsRewardsSLAMRandomTiny} 
\end{table*}

\begin{table*}[h]
	\caption{ $50$ Trials of at most $20$ cycles of autonomy loop Fig.~\ref{fig:AutonomyLoop} where planning sessions implemented by Algorithm CCPC-PFT-DPW versus PC-PFT-DPW. Same seed in both  algorithms. This problem is the {\bf SLAM} described in Section \ref{sec:SLAM} in { \bf our second scenario} shown at Fig.~\ref{fig:CCPCPFTSLAMTinyFull} and Fig.~\ref{fig:PCPFTSLAMTinyFull}. Here we study the number of collisions and the reward value.}
	\centering
	\resizebox{\textwidth}{!}{
		\begin{tabular}{|c|c|c|c|c|c|c|c|}
			\hline
			\multicolumn{2}{|c|}{ Parameters } & \multicolumn{2}{|c|}{$\hat{\mathrm{P}}(S|b_0)$} &\multicolumn{2}{|c|}{ num coll. } &  \multicolumn{2}{|c|}{ mean cum. rew. $\pm$ std}  \\
			\hline
			Operator $\phi$  & $\delta$  &  CCPC-PFT-DPW   & PC-PFT-DPW  &  CCPC-PFT-DPW & PC-PFT-DPW &  CCPC-PFT-DPW & PC-PFT-DPW  \\
			\hline  
			\eqref{eq:ProbSafeGivenBelief} & $0.8$ & $0.6$ &  $0.6$ &$28/70$ &  $28/70$  & $-109.92 \pm 11.55$  &   $-106.68  \pm 12.77$  \\
			\hline
		\end{tabular}
	}
	\label{tbl:CollsRewardsSLAMRandomTinyFull} 
\end{table*}

For the SLAM 
problem we study the influence of making posterior belief safe before pushing forward in time. Since this aspect stemmed from Chance Constraints, to differentiate between two approaches, in this study we change the name of our approach with nullifying the unsafe part of belief to CC-PC-PFT (Alg.~\ref{alg:PCMCTS}).  We call our method, with pushing forward in time with action and the observation generally unsafe belief,  PC-PFT. 
Let us remind that the difference in the inner constraint as such. Instead of payoff operator as \eqref{eq:ProbSafeGivenBelief} we set 
\begin{equation}
	\begin{gathered}
	\phi(\bar{b}_{\ell}) {=} \prob\big(\{x_{\ell} {\in} \mathcal{X}^{\mathrm{safe}}_{\ell}\}\big|\bar{b}_{\ell}\big) {=} \\
	\textstyle\prob\big(\{x_{\ell} {\in} \mathcal{X}^{\mathrm{safe}}_{\ell}\}\big|b_{0}, a_{0:\ell-1}, z_{1:\ell}, \bigcap_{i=0}^{\ell-1} \{x_{i} {\in} \mathcal{X}^{\mathrm{safe}}_{i}\}\big)
	\end{gathered}
\end{equation}	
\begin{equation}	
\begin{gathered}
	\!\! \phi(\bar{b}^-_{\ell}) {=} \prob\big(\!\{x_{\ell} {\in} \mathcal{X}^{\mathrm{safe}}_{\ell}\}\big|\bar{b}^-_{\ell}\big) {=} \\
	\textstyle\prob\big(\!\{x_{\ell} {\in} \mathcal{X}^{\mathrm{safe}}_{\ell}\}\big|b_{0}, a_{0:\ell-1}, z_{1:\ell-1},\! \bigcap_{i=0}^{\ell-1} \{x_{i} {\in} \mathcal{X}^{\mathrm{safe}}_{i}\}\!\big). \!\!\!\!\!\!\!\!
	\end{gathered}
\end{equation}

Table~\ref{tbl:CollsRewardsSLAMRandomTiny} presents the results for SLAM in our first setup with tiny obstacles. 
Here our setup is as follows. We have a rectangular area where we randomly sow rectangular tiny obstacles without replacement. It means if we randomly sow the number of tiny obstacles equal to the number of cells within the large rectangle we will obtain a complete large rectangle. We randomly sow the tiny obstacles in each trial. We make transition model of the agent \eqref{eq:SLAMtransitionAgent} deterministic by nullifying the noise. With drawing $80\%$ of tiny obstacles (Fig.~\ref{fig:CCPCPFTSLAMTiny}) the PC-PFT-DPW, without making belief safe and maintaining a pair of the beliefs, the scenario in Fig.~\ref{fig:CCPCPFTSLAMTiny}  reached the belief node where {\bf all the actions were claimed unsafe and pruned}, even the $\boldsymbol{0}$ action. As we have seen in the simulation $\boldsymbol{0}$ action was pruned the last and this is a direct result of the fact that unsafe belief particles were propagated with $\boldsymbol{0}$ action and updated with received observation.  When we do the same operation previously making belief safe, we obtain again the safe belief since particles were propagated with $\boldsymbol{0}$ action and, therefore, stay at the same places. 

In our second setup we fill the complete rectangle with tiny obstacles in a random manner as previously debated (Fig.~\ref{fig:PCPFTSLAMTinyFull}).   We show our results in Table~\ref{tbl:CollsRewardsSLAMRandomTinyFull}. We did not obtained a significant difference in two approaches. Interestingly, as we see the safety is much challenging in this problem due to challenging robot localization with simultaneous mapping of uncertain single landmark. Our prior $b_0$ in SLAM problem is Gaussian with diagonal variances of $0.1$.

\paragraph{PushBox2D}

\begin{figure*}[t]
	\centering
	\begin{minipage}[t]{0.24\textwidth}
		\centering 
		\includegraphics[width=\textwidth]{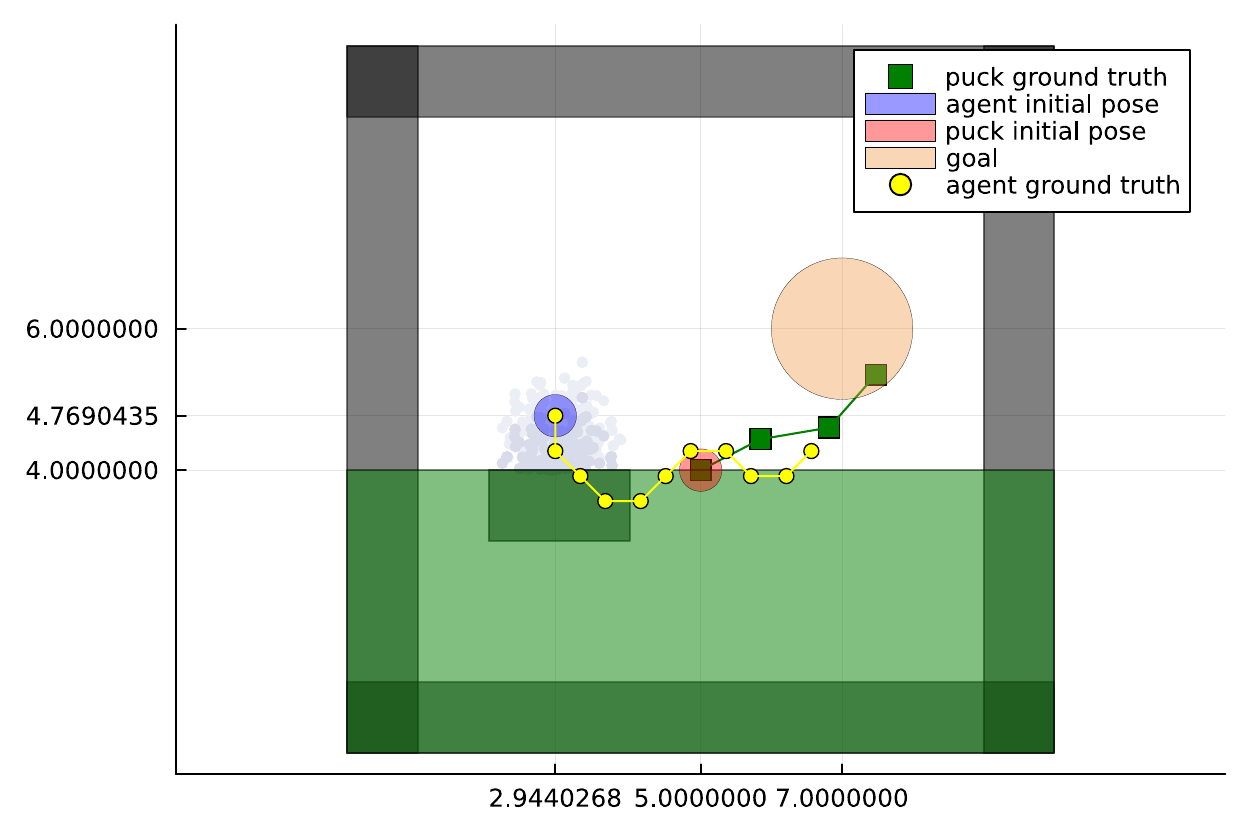}
		\subcaption{$\delta=0.0$}
		\label{fig:}
	\end{minipage}%
	\hfill
	\begin{minipage}[t]{0.24\textwidth}
		\centering 
		\includegraphics[width=\textwidth]{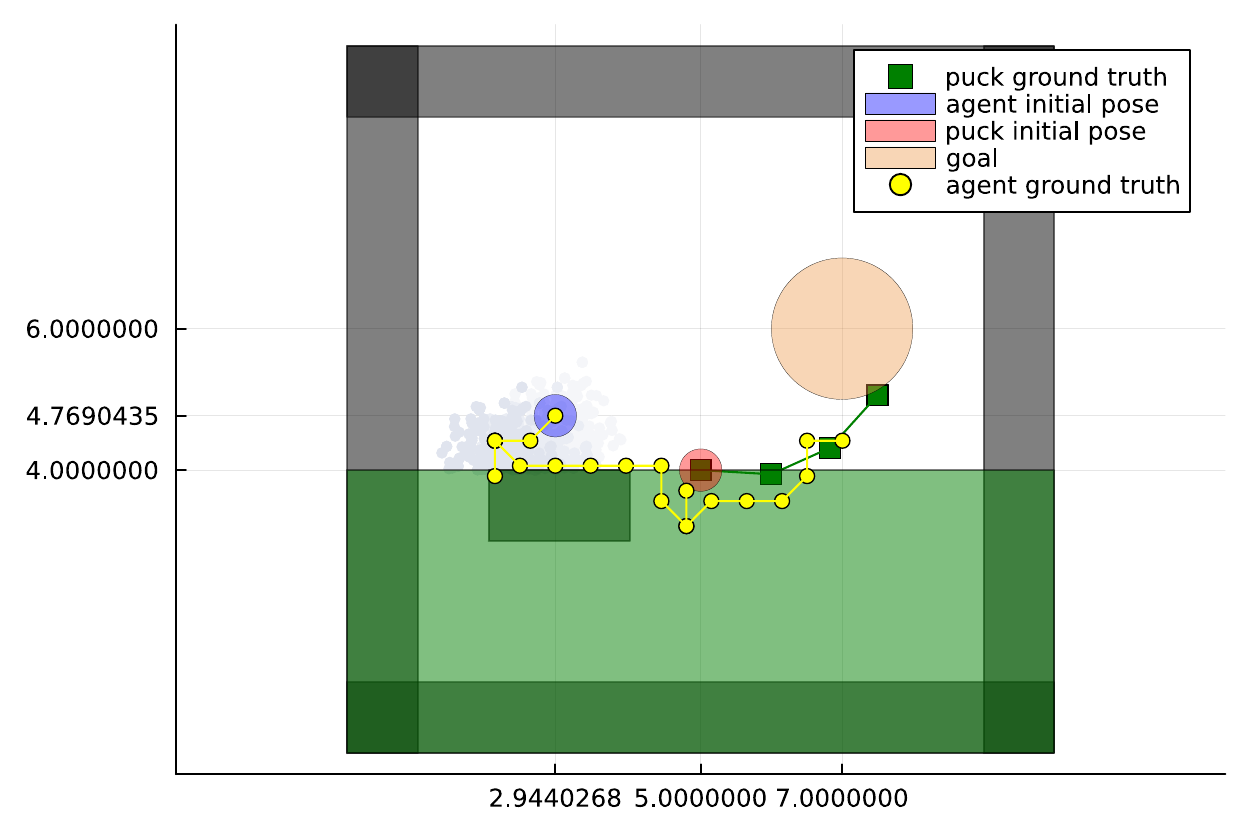}
		\subcaption{$\delta = 0.3$}
		\label{fig:}
	\end{minipage} 
	\begin{minipage}[t]{0.24\textwidth}
		\centering 
		\includegraphics[width=\textwidth]{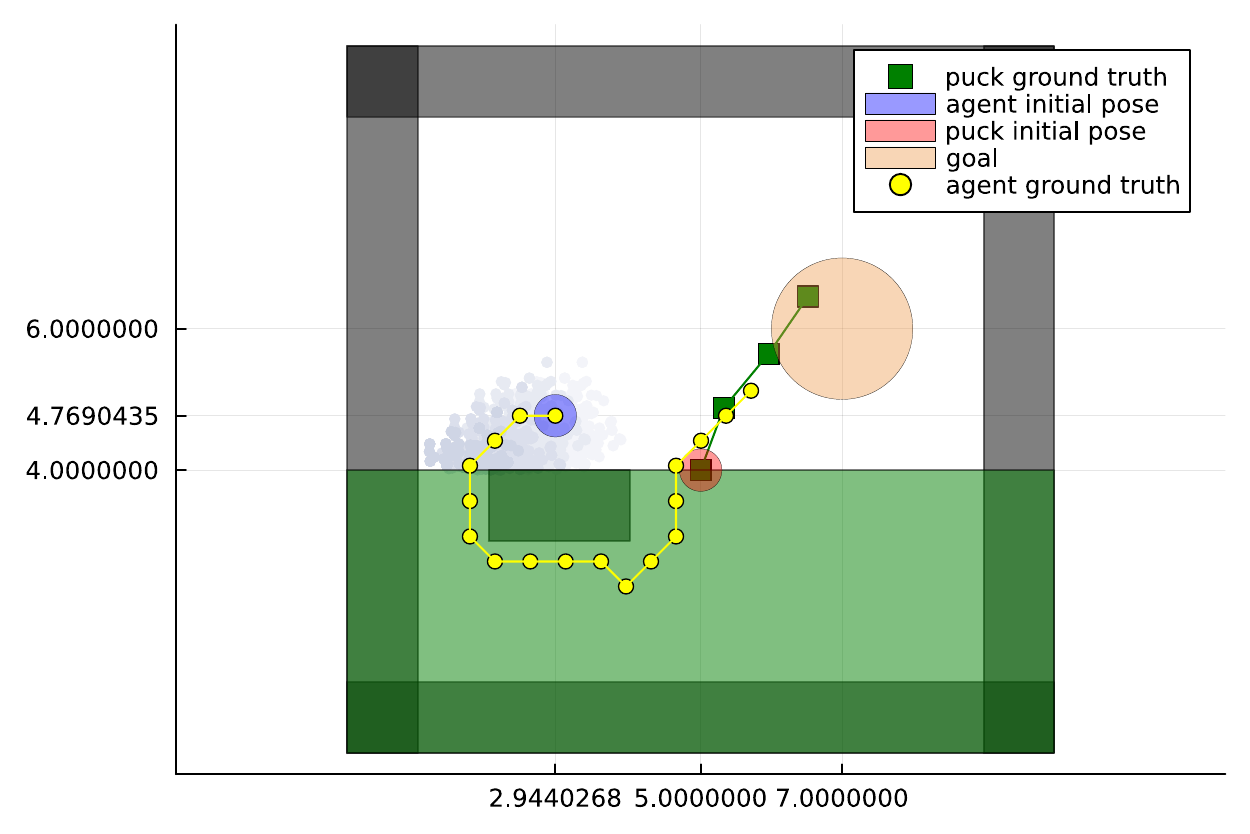}
		\subcaption{$\delta = 0.7$}
		\label{fig:}
	\end{minipage} 
	\hfill
	\begin{minipage}[t]{0.24\textwidth}
		\centering 
		\includegraphics[width=\textwidth]{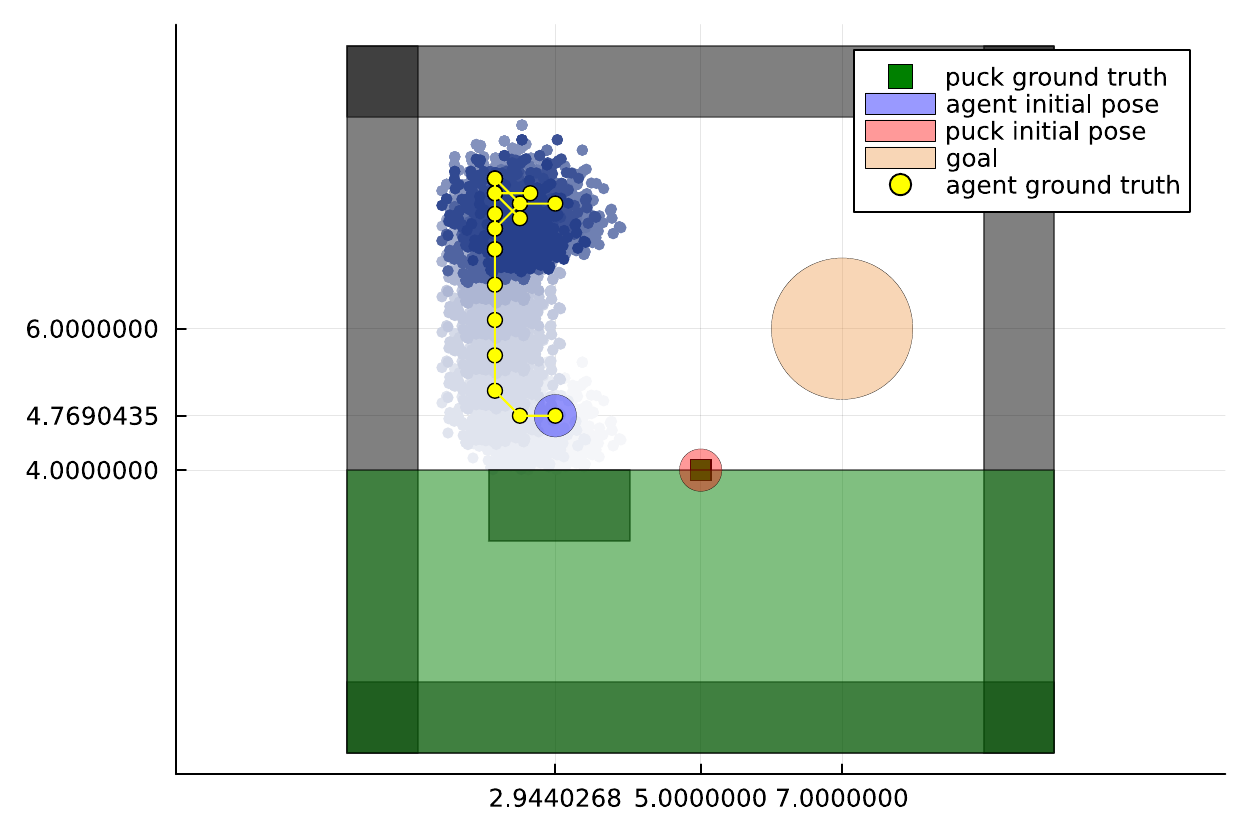}
		\subcaption{$\delta = 1.0$}
		\label{fig:}
		\hfill
	\end{minipage} 
	\caption{Visualization of actual PushBox2D simulation with several values of $\delta$. }      
	\label{fig:Pushbox2Dscenario}	 
\end{figure*}
\begin{table*}[h]
	\caption{ $20$ Trials of at most $20$ autonomy loop cycles of Algorithms CCPC-PFT-DPW versus PC-PFT-DPW. Same seed in both  algorithms. This problem is the {\bf  2DPushBox} described in Section \ref{sec:PushBox2Ddescription}. The operator $\phi$ conforms to \eqref{eq:ProbSafeGivenBelief} and \eqref{eq:ProbSafeGivenPropagatedBelief}. }
	\centering
	\resizebox{\textwidth}{!}{
	\begin{tabular}{|c|c|c|c|c|c|c|c|c|c|c|}
		\hline
		\multicolumn{5}{|c|}{ Parameters } & \multicolumn{2}{|c|}{$\hat{\mathrm{P}}(S|b_0)$ in accord to \eqref{eq:ProbSafeTraj}} &\multicolumn{2}{|c|}{ Est. prob. of  puck reaching the goal} &  \multicolumn{2}{|c|}{ mean cum. rew. $\pm$ std}  \\
		\hline
		tree queries  & $\delta$  & $L$ & propagated constr. &  rollout &CCPC-PFT-DPW   & PC-PFT-DPW  &  CCPC-PFT-DPW & PC-PFT-DPW &  CCPC-PFT-DPW & PC-PFT-DPW  \\
		\hline  
		$1500$ & $0.0$  & $20$&Yes & Yes & $0.1$& $0.0$ & $1.0$  & $1.0$ &$-20.39 \pm 6.12$&$ -20.73\pm 4.32$   \\
		\hline
		$1500$ & $0.3$  & $20$&Yes & Yes & $0.45$& $0.3$ & $0.85$ & $0.85$ &$ -37.33\pm 11.93$  & $-32.31 \pm 11.58$\\
		\hline
		$1500$ & $0.7$  & $20$&Yes & Yes & $0.65$& $0.7$ & $0.65$  & $0.8$ &$-43.04 \pm 6.82$&$  -41.56 \pm 9.31 $   \\
		\hline
		$1500$ & $0.7$  & $20$&No & Yes & $0.2$& $0.3$ & $0.9$  & $0.9$ &$-38.38\pm 9.69$&$   -34.07\pm 9.20 $   \\
		\hline
		$1500$ & $0.7$  & $20$&Yes & No & $0.55$& $0.65$ & $0.0$  & $0.0$ &$-59.82\pm 0.25$&$ -59.85 \pm  0.26$   \\
		\hline
		$1500$ & $1.0$  & $20$&Yes & Yes &- & $1.0$ & -  & $0.0$ &-&$ -60.18  \pm  0.18$   \\
		\hline
		$1500$ & $1.0$  & $20$&No & Yes &- & $0.2$ & -  & $0.9$ &-&$  -35.81 \pm 7.59$   \\
		\hline
	\end{tabular}
}
	\label{tbl:PushBox2D} 
\end{table*}

We constructed a challenging scenario where evading the obstacle significantly complicates putting the puck into the hole (the goal). 
Let us contemplate the results presented in Table.~\ref{tbl:PushBox2D}.  We selected $m=10$ and $\epsilon=0$ in rollout summarized by Alg.~\ref{alg:SafeRollout}.  As we see, constraining the propagated belief significantly improves safety while preserving reaching the goal by the puck.  
\section{Conclusions}
\label{sec:Concl}
In this work, we introduced an anytime online approach to perform Safe and Risk Aware Belief Space Planning in continuous domains in terms of states, actions, and observations. We rigorously analyzed our approach in terms of convergence. Our prominent novelty is assuring safety with respect to the belief tree expanded so far. As opposed to SOTA in continuous domains, we are not mixing safe and dangerous actions in the search tree. Our belief tree is safe with respect to our PC and consist solely of the safe actions. Moreover, when our PC is satisfied, it is satisfied starting from each belief action node, ensuring a match in a current planning session and future planning sessions.    
We corroborated our theoretical development by simulating {\bf four} different problems in continuous domains. 
Each problem exhibited a different phenomenon caught by our methodology. 

\section{Acknowledgment}
This work was supported by the Israel Science Foundation (ISF).


\bibliographystyle{IEEEtran}




\begin{thebibliography}{10}
\providecommand{\url}[1]{#1}
\csname url@samestyle\endcsname
\providecommand{\newblock}{\relax}
\providecommand{\bibinfo}[2]{#2}
\providecommand{\BIBentrySTDinterwordspacing}{\spaceskip=0pt\relax}
\providecommand{\BIBentryALTinterwordstretchfactor}{4}
\providecommand{\BIBentryALTinterwordspacing}{\spaceskip=\fontdimen2\font plus
\BIBentryALTinterwordstretchfactor\fontdimen3\font minus
  \fontdimen4\font\relax}
\providecommand{\BIBforeignlanguage}[2]{{%
\expandafter\ifx\csname l@#1\endcsname\relax
\typeout{** WARNING: IEEEtran.bst: No hyphenation pattern has been}%
\typeout{** loaded for the language `#1'. Using the pattern for}%
\typeout{** the default language instead.}%
\else
\language=\csname l@#1\endcsname
\fi
#2}}
\providecommand{\BIBdecl}{\relax}
\BIBdecl

\bibitem{Madani03AI}
O.~Madani, S.~Hanks, and A.~Condon, ``On the undecidability of probabilistic
  planning and related stochastic optimization problems,'' \emph{Artificial
  Intelligence}, vol. 147, no. 1-2, pp. 5--34, 2003.

\bibitem{Santana16aaai}
P.~Santana, S.~Thi{\'e}baux, and B.~Williams, ``Rao*: An algorithm for
  chance-constrained pomdp's,'' in \emph{Proceedings of the AAAI Conference on
  Artificial Intelligence}, vol.~30, no.~1, 2016.

\bibitem{Zhitnikov22arxiv}
A.~Zhitnikov and V.~Indelman, ``Risk aware adaptive belief-dependent
  probabilistically constrained continuous pomdp planning,'' \emph{arXiv
  preprint arXiv:2209.02679}, 2022.

\bibitem{Zhitnikov24TRO}
------, ``Simplified continuous high dimensional belief space planning with
  adaptive probabilistic belief-dependent constraints,'' \emph{{IEEE} Trans.
  Robotics}, 2024.

\bibitem{Jamgochian23ICAPS}
A.~Jamgochian, A.~Corso, and M.~J. Kochenderfer, ``Online planning for
  constrained pomdps with continuous spaces through dual ascent,'' in
  \emph{Proceedings of the International Conference on Automated Planning and
  Scheduling}, vol.~33, no.~1, 2023, pp. 198--202.

\bibitem{Jamgochian23arxiv}
A.~Jamgochian, H.~Buurmeijer, K.~H. Wray, A.~Corso, and M.~J. Kochenderfer,
  ``Constrained hierarchical monte carlo belief-state planning,'' \emph{arXiv
  preprint arXiv:2310.20054}, 2023.

\bibitem{Lee18nips}
J.~Lee, G.-H. Kim, P.~Poupart, and K.-E. Kim, ``Monte-carlo tree search for
  constrained pomdps,'' \emph{Advances in Neural Information Processing
  Systems}, vol.~31, 2018.

\bibitem{Altman99book}
E.~Altman, \emph{Constrained Markov decision processes}.\hskip 1em plus 0.5em
  minus 0.4em\relax CRC Press, 1999.

\bibitem{Beck17first}
A.~Beck, \emph{First-order methods in optimization}.\hskip 1em plus 0.5em minus
  0.4em\relax SIAM, 2017.

\bibitem{Ho2023arxiv}
Q.~H. Ho, T.~Becker, B.~Kraske, Z.~Laouar, M.~Feather, F.~Rossi, M.~Lahijanian,
  and Z.~N. Sunberg, ``Recursively-constrained partially observable markov
  decision processes,'' \emph{arXiv preprint arXiv:2310.09688}, 2023.

\bibitem{Munos2014book}
R.~Munos, \emph{From Bandits to Monte-Carlo Tree Search: The Optimistic
  Principle Applied to Optimization and Planning}, 2014.

\bibitem{Ajdarow23AAAI}
M.~Ajdar{\'o}w, {\v{S}}.~Brlej, and P.~Novotn{\`y}, ``Shielding in
  resource-constrained goal pomdps,'' in \emph{AAAI Conf. on Artificial
  Intelligence}, vol.~37, no.~12, 2023, pp. 14\,674--14\,682.

\bibitem{Mazzi23ai}
G.~Mazzi, A.~Castellini, and A.~Farinelli, ``Risk-aware shielding of partially
  observable monte carlo planning policies,'' \emph{Artificial Intelligence},
  vol. 324, p. 103987, 2023.

\bibitem{Silver10nips}
D.~Silver and J.~Veness, ``Monte-carlo planning in large pomdps,'' in
  \emph{Advances in Neural Information Processing Systems (NeurIPS)}, 2010, pp.
  2164--2172.

\bibitem{Moss24Arxiv}
R.~J. Moss, A.~Jamgochian, J.~Fischer, A.~Corso, and M.~J. Kochenderfer,
  ``Constrainedzero: Chance-constrained pomdp planning using learned
  probabilistic failure surrogates and adaptive safety constraints,''
  \emph{arXiv preprint arXiv:2405.00644}, 2024.

\bibitem{Chou22WAFR}
G.~Chou, N.~Ozay, and D.~Berenson, ``Safe output feedback motion planning from
  images via learned perception modules and contraction theory,'' in
  \emph{International Workshop on the Algorithmic Foundations of
  Robotics}.\hskip 1em plus 0.5em minus 0.4em\relax Springer, 2022, pp.
  349--367.

\bibitem{Dean21CORL}
S.~Dean, A.~Taylor, R.~Cosner, B.~Recht, and A.~Ames, ``Guaranteeing safety of
  learned perception modules via measurement-robust control barrier
  functions,'' in \emph{Conference on Robot Learning}.\hskip 1em plus 0.5em
  minus 0.4em\relax PMLR, 2021, pp. 654--670.

\bibitem{Sunberg18icaps}
Z.~Sunberg and M.~Kochenderfer, ``Online algorithms for pomdps with continuous
  state, action, and observation spaces,'' in \emph{Proceedings of the
  International Conference on Automated Planning and Scheduling}, vol.~28,
  no.~1, 2018.

\bibitem{Kocsis06ecml}
L.~Kocsis and C.~Szepesv{\'a}ri, ``Bandit based monte-carlo planning,'' in
  \emph{European conference on machine learning}.\hskip 1em plus 0.5em minus
  0.4em\relax Springer, 2006, pp. 282--293.

\bibitem{Auger13Sp}
D.~Auger, A.~Couetoux, and O.~Teytaud, ``Continuous upper confidence trees with
  polynomial exploration--consistency,'' in \emph{Machine Learning and
  Knowledge Discovery in Databases: European Conference, ECML PKDD 2013,
  Prague, Czech Republic, September 23-27, 2013, Proceedings, Part I 13}.\hskip
  1em plus 0.5em minus 0.4em\relax Springer, 2013, pp. 194--209.

\bibitem{Zhitnikov22ai}
A.~Zhitnikov and V.~Indelman, ``Simplified risk aware decision making with
  belief dependent rewards in partially observable domains,'' \emph{Artificial
  Intelligence, Special Issue on ``Risk-Aware Autonomous Systems: Theory and
  Practice"}, 2022.

\bibitem{Lim23jair}
M.~H. Lim, T.~J. Becker, M.~J. Kochenderfer, C.~J. Tomlin, and Z.~N. Sunberg,
  ``Optimality guarantees for particle belief approximation of pomdps,''
  \emph{Journal of Artificial Intelligence Research}, vol.~77, pp. 1591--1636,
  2023.

\end{thebibliography}

\begin{appendices}
	
\section{Proof of Lemma~\ref{thm:RepresentValueStochPolicy}(Representation of the Value function).}  \label{proof:RepresentValueStochPolicy}
Before we begin, let us clarify that when we write  $\{\probd^{\pi}_{\ell}(a_{\ell}|b_{\ell}) \}_{\ell=1}^{L-1}$, the $a_{\ell}$ and $b_{\ell}$ inside the $\{\probd_{\ell}(a_{\ell}|b_{\ell}) \}_{\ell=1}^{L-1}$ can be a random variables for all relevant $\ell$ or corresponding realizations. However, $\{\probd^{\pi}_{\ell} \}_{\ell=1}^{L-1}$ is the series of distributions of length $L{-}1$ and corresponding actions and beliefs are unknown.  In addition, we remind to the reader that $\pi_{\ell}(a_{\ell}, b_{\ell}) = \probd^{\pi}_{\ell}(a_{\ell}|b_{\ell}) \quad \forall \ell \in 1:L{-}1$ and $\pi {=} \{\probd^{\pi}_{\ell} \}_{\ell=1}^{L-1}$.  
\begin{equation}
\begin{gathered}
\mathbb{E}^{\mathrm{T},\mathrm{O}}\big[\sum_{\ell=0}^{L-1}\rho_{\ell+1}(b_{\ell}, a_{\ell}, b_{\ell+1})\big| b_0, \pi \big] {=} \\
\sum_{\ell=0}^{L-1} \mathbb{E}^{\mathrm{T},\mathrm{O}}\big[\rho_{\ell+1}(b_{\ell}, a_{\ell}, b_{\ell+1})\big| b_0, \pi \big]{=}\sum_{\ell=0}^{L-1} \mathbb{E}^{\mathrm{T},\mathrm{O}}\big[\rho_{\ell+1}\big| b_0, \pi \big] \!\!\!\!\!\!\!\!\!
\end{gathered}
\end{equation}
\begin{equation}
\begin{gathered}
\mathbb{E}^{\mathrm{T},\mathrm{O}}\big[\rho_{\ell+1}\big| b_0, \pi \big] {=} \int_{\rho_{\ell+1}}\!\!\!\!\!\!\!\! \rho_{\ell+1} \probd(\rho_{\ell+1}| b_0, \{\probd^{\pi}_i \}_{i=0}^{L-1})  \mathrm{d} \rho_{\ell+1}=\\
\int_{\rho_{\ell+1}}\!\!\!\!\! \rho_{\ell+1}\!\!\!\!\!\!\!\!\!\! \int\limits_{\substack{b_{1:\ell} \\ a_{0:\ell} \in \times_{i=1}^{\ell}\mathcal{A}}}\!\!\!\!\!\!\!\!\!\! \probd(\rho_{\ell+1}, b_{1:\ell}, a_{1:\ell}| b_0, \{\probd^{\pi}_i \}_{i=0}^{L-1})\mathrm{d} b_{1:\ell} \mathrm{d} a_{0:\ell}  \mathrm{d} \rho_{\ell+1} {=} \\
\int_{\rho_{\ell+1}} \!\!\!\!\rho_{\ell+1} \!\!\!\!\!\!\!\!\!\!\int\limits_{\substack{b_{1:\ell} \\ a_{0:\ell} \in \times_{i=1}^{\ell}\mathcal{A}}}\!\!\!\!\!\!\!\!\!\! \probd(\rho_{\ell+1}| b_{0:\ell}, a_{0:\ell}) \\
\probd(b_{1:\ell}, a_{0:\ell}| b_0, \{\probd^{\pi}_i \}_{i=0}^{L-1})\mathrm{d} b_{1:\ell} \mathrm{d} a_{0:\ell}  \mathrm{d} \rho_{\ell+1} = \\
\int\limits_{\substack{b_{1:\ell} \\ a_{0:\ell} \in \times_{i=1}^{\ell}\mathcal{A}}} \Bigg( \int_{\rho_{\ell+1}} \rho_{\ell+1} \probd(\rho_{\ell+1}| b_{0:\ell}, a_{0:\ell})\mathrm{d} \rho_{\ell+1} \Bigg)\\
\probd(b_{1:\ell}, a_{0:\ell}| b_0, \{\probd^{\pi}_i \}_{i=0}^{L-1})\mathrm{d} b_{1:\ell} \mathrm{d} a_{0:\ell}
\end{gathered}
\end{equation} 
We now use a chain rule from the future time back on $\probd(b_{1:\ell}, a_{0:\ell}| b_0, \{\probd^{\pi}_i \}_{i=0}^{L-1})$  an got  
\begin{equation}
\begin{gathered} 
\mathbb{E}^{\mathrm{T},\mathrm{O}}\big[\rho_{\ell+1}\big| b_0, \pi \big] = \underset{a_0}{\mathbb{E}}\Bigg[\underset{b_1}{\mathbb{E}} \Bigg[  \underset{a_1}{\mathbb{E}} \Bigg[\underset{b_2}{\mathbb{E}}\Big[  \dots \\
\underset{a_{\ell}}{\mathbb{E}}\big[\mathbb{E}\big[\rho_{\ell+1}| b_{\ell}, a_{\ell}\big]\big| b_{\ell},\pi_{\ell}\Big] \dots \\
\Big| b_1, a_1\Big]\Bigg|b_1, \pi_1\Bigg] \Bigg| b_0, a_0\Bigg]\Bigg| b_0, \pi_0\Bigg].
\end{gathered}
\end{equation}
\qed
\section{Proof of Theorem~\ref{thm:Represent}(Representation of Our Outer Constraint).}  
\label{sec:Represent}
Before we begin, let us clarify that when we write $$\textstyle\prob\left(\Big(\mathbf{1}_{A^{\delta}_{0}}(b_{0})\prod_{\ell=1}^{L} \mathbf{1}_{A^{\delta}_{\ell}}(b_{\ell})\Big) {=}1 |b_0, a_0, \{\mathbb{P}^{\pi}_{\ell}\}_{\ell=1}^{L-1} \right),$$ the actions $a_{\ell}$ and the beliefs $b_{\ell}$  inside $\{\probd^{\pi}_\ell\}_{\ell=1}^{L-1}$ are unknown random quantities. In addition, we remind to the reader that  $\pi_{\ell}(a_{\ell}, b_{\ell}){=}\mathbb{P}^{\pi}_{\ell}(a_{\ell}|b_{\ell}) \quad \forall \ell {\in} 1{:}L{-}1$ and $\pi{=}\{\mathbb{P}^{\pi}_{\ell}\}_{\ell=1}^{L-1}$. Moreover, in this paper each posterior belief is associated with corresponding propagated belief. Therefore we can rescind the explicit dependence of the indicator on propagated belief.  
\begin{equation}
	\label{eq:PCIntegral}
	\begin{gathered}
		\!\!\!\!\textstyle\mathbb{E}\Big[\mathbf{1}_{A^{\delta}_{0}}(b_{0})\prod_{\ell=1}^{L} \mathbf{1}_{A^{\delta}_{\ell}}(b_{\ell}) |b_0, a_0, \{\probd^{\pi}_\ell \}_{\ell=1}^{L-1}  \Big] {=}\\
		\int\limits_{\substack{b_{1:L} \\ a_{1:L-1} {\in} \times_{\ell=1}^{L-1}\mathcal{A}}}\!\!\!\!\!\!\!\!\!\!\!\!\! \mathbf{1}_{A^{\delta}_{0}}(b_{0})\prod_{\ell=1}^{L} \mathbf{1}_{A^{\delta}_{\ell}}(b_{\ell})\\
		\!\!\!\!\!\probd(b_{1:L}, a_{1:L-1} |b_0, a_0, \{\probd^{\pi}_\ell \}_{\ell=1}^{L-1})\mathrm{d}b_{1:L}\mathrm{d}a_{1:L-1}.
	\end{gathered}
\end{equation}
Now, we need to handle $\probd(b_{1:L}, a_{0:L-1} |b_0, a_0, \{\probd^{\pi}_\ell \}_{\ell=1}^{L-1})$. It holds that $$\probd(b_{1:L}, a_{1:L-1} |b_0, a_0, \{\probd^{\pi}_\ell \}_{\ell=1}^{L-1})$$ equals to
\begin{equation}
	\label{eq:PDFBeliefActions}
	\begin{gathered}
		\probd(b_{2:L}, a_{2:L-1} |b_0, a_0, b_1, a_1, \probd^{\pi}_1(a_1|b_1), \{\probd^{\pi}_\ell \}_{\ell=2}^{L-1})\\
		\probd(b_1, a_1 |b_0, a_0, \{\probd_\ell \}_{\ell=1}^{L-1}) = \\
		\!\!\probd(b_{2:L}, a_{2:L-1} |b_1, a_1, \{\probd^{\pi}_\ell \}_{\ell=2}^{L{-}1})\probd^{\pi}_1(a_1 | b_1 ) \probd(b_1| b_0, a_0){=}\!\!\!\!\!\! \\
		\textstyle\probd(b_{L}| b_{L-1}, a_{L-1})\! \prod_{\ell=1}^{L-1} \probd^{\pi}_{\ell}(a_{\ell} | b_{\ell} ) \probd(b_{\ell}| b_{\ell-1}, a_{\ell-1}). 
	\end{gathered}
\end{equation}
We now merge \eqref{eq:PCIntegral} and \eqref{eq:PDFBeliefActions}, and land at the desired result 
\begin{equation}
	\begin{gathered}	
		\mathbf{1}_{A^{\delta}_{0}}(b_{0})\!\!\!\!\!\!\!\!\!\!\!\!\textstyle\int\limits_{\substack{b_{1:L} \\ a_{1:L-1}\in \times_{i=1}^{L-1}\mathcal{A}}}\!\!\!\!\!\!\!\!\!\!\!\!\mathbf{1}_{A^{\delta}_{L}}( b_{L})\probd(b_{L}| b_{L-1}, a_{L-1}) \notag\\
		\textstyle\prod_{\ell=1}^{L-1}\Big( \probd^{\pi}_{\ell}(a_{\ell} | b_{\ell} ) \probd(b_{\ell}| b_{\ell-1}, a_{\ell-1})\mathbf{1}_{A^{\delta}_{\ell}}(b_{\ell})\Big)
		\mathrm{d}b_{1:L}\mathrm{d}a_{1:L-1} {=}\\
		\mathbf{1}_{A^{\delta}_{0}}(b_{0}) \int_{b_1}\!\!\! \probd(b_{1}| b_{0}, a_{0})\mathbf{1}_{A^{\delta}_{1}}(b_{1}) \int_{a_1} \probd^{\pi}_1(a_{1} | b_{1} ) \Big(   \dots  \\
		\int_{b_L}\mathbf{1}_{A^{\delta}_{L}}(b_{L})\probd(b_{L}| b_{L-1}, a_{L-1})\mathrm{d}b_L \dots \Big)\mathrm{d}a_1\mathrm{d}b_1  = \nonumber\\
		\mathbf{1}_{A^{\delta}_{0}}(b_{0})\underset{b_1}{\mathbb{E}} \Bigg[ \mathbf{1}_{A^{\delta}_{1}}(b_{1}) \underset{a_1}{\mathbb{E}} \Bigg[\underset{b_2}{\mathbb{E}}\Big[\mathbf{1}_{A^{\delta}_{2}}(b_{2})  \dots \\
		\mathbb{E}\big[\mathbf{1}_{A^{\delta}_{L}}(b_{L})| b_{L-1}, a_{L-1}\big] \dots \Big| b_1, a_1\Big]\Bigg|b_1, \pi_1\Bigg] \Bigg|b_0, a_0\Bigg]=\\
		\mathbf{1}_{A^{\delta}_{0}}(b_{0})\underset{b_1}{\mathbb{E}} \Bigg[ \underset{a_1 {\sim}\probd^{\pi}_1(a_1|b_1) }{\mathbb{E}} \Bigg[ \\
		\prob\big(\big(\prod_{\ell=1}^{L} \mathbf{1}_{A^{\delta}_{\ell}}(b_{\ell})\big){=}1  \Big| b_1, a_1, \pi\Big)\Bigg|b_1, \pi_1\Bigg] \Bigg|b_0, a_0\Bigg].
	\end{gathered}	
\end{equation}
\qed

\section{Proof of Theorem~\ref{thm:Cond} (Necessary condition for theoretical posteriors to be safe)}
\label{proof:Cond}
For the necessary condition we prove the inverse implication.  Suppose that $\forall z_{\ell} {\in} \mathcal{Z}$ it holds that $\mathrm{P}\big(\{x_{\ell} {\in} \mathcal{X}^{\mathrm{safe}}_{\ell}\}\big|h^-_{\ell}, z_{\ell}\big) 
{\geq} \delta$. We arrive at 
\begin{equation}
		\Big(\int_{z_{\ell} \in \mathcal{Z}}\mathrm{P}\big(\{x_{\ell} {\in} \mathcal{X}^{\mathrm{safe}}_{\ell}\}\big|h^-_{\ell}, z_{\ell}\big)\probd(z_{\ell}|h^-_{\ell}) \mathrm{d}z_{\ell} \Big){\geq}\delta.
\end{equation}  
\qed
\section{VaR and CVaR as safety cost operators.}
\label{sec:VaRCVaR}
\begin{figure}
	\centering 
	\includegraphics[width=0.2\textwidth]{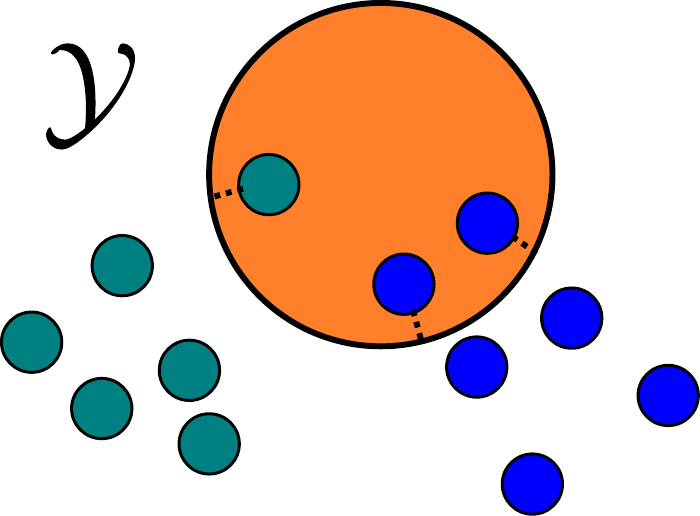}
	\caption{Illustration of complex safety operators in multirobot setting. }
	\label{fig:TwoRobotsCircObstacle}
\end{figure}
Suppose we have particle represented belief and the obstacle of the circular form Fig.~\ref{fig:TwoRobotsCircObstacle}. In addition we have two robots {\color{teal} teal} and {\color{blue} blue}. 
Each particle of the belief is a concatenated position of each robot such that if $x$ is a particle, the $x[1{:}2]$ corresponds to the first robot and $x[3:4]$ corresponds to the second robot. We shall check such a constraint for each robot separately. For clarity let $x$ denote the position of the one of the robots. Suppose the map $\mathcal{M}$ is given. We first define a distance from the safe space $\mathcal{Y} \subseteq \mathcal{M}$ as $\mathrm{dist}(x, \mathcal{Y}) = \mathrm{min}_{y\in \mathcal{Y}} \| x -y \|_2$. We then define  Value at Risk (VaR) as 
\begin{equation}
\begin{gathered}
	\theta(b) {\bydef} \mathrm{VaR}^b_{\alpha}[\mathrm{dist}(x, \mathcal{Y})] {=}\\
	 \min\{\xi | \prob(\mathrm{dist}(x, \mathcal{Y}) {\leq} \xi ) {\geq} 1{-}\alpha \}. 
\end{gathered}
\end{equation}
The Conditional Value at Risk (CVaR) is specified as 
\begin{equation}
	\begin{gathered}
	\theta(b) {\bydef} \mathrm{CVaR}^b_{\alpha}[\mathrm{dist}(x, \mathcal{Y})] {=} \\
	\mathbb{E}[\mathrm{dist}(x, \mathcal{Y}) |\{x: \mathrm{dist}(x, \mathcal{Y}) \geq  \mathrm{VaR}^b_{\alpha}[\mathrm{dist}(x, \mathcal{Y})]\}].
\end{gathered}
\end{equation}
Both of these operators are {\bf cost} operators. 
\end{appendices}
\end{document}